\documentclass{article}

\usepackage{arxiv}

\usepackage[utf8]{inputenc} % allow utf-8 input
\usepackage[T1]{fontenc}    % use 8-bit T1 fonts
\usepackage{hyperref}       % hyperlinks
\usepackage{url}            % simple URL typesetting
\usepackage{booktabs}       % professional-quality tables
\usepackage{amsfonts}       % blackboard math symbols
\usepackage{nicefrac}       % compact symbols for 1/2, etc.
\usepackage{microtype}      % microtypography
\usepackage{algorithm}
\usepackage{algorithmic}
\usepackage{graphicx}
\usepackage{subfig}

\usepackage{xcolor}         % colors
\usepackage{amsmath}
\usepackage{fourier}
\usepackage{float}

\DeclareMathAlphabet{\mathcal}{OMS}{cmsy}{m}{n}
\SetMathAlphabet{\mathcal}{bold}{OMS}{cmsy}{b}{n}
\title{Importance of Empirical Sample Complexity Analysis\\
for Offline Reinforcement Learning}

\author{
  Samin Yeasar Arnob  \\
  Mila, McGill University, \\
  \texttt{samin.arnob@mail.mcgill.ca} 
  \And
  Riashat Islam \\
  Mila, McGill University, \\
  \AND
  Doina Precup \\
 Mila, McGill University, \\
 DeepMind\\
}

\begin{document}
\maketitle

\begin{abstract}
We hypothesize that empirically studying the sample complexity of offline reinforcement learning (RL) is crucial for the practical applications of RL in the real world. Several recent works have demonstrated the ability to learn policies directly from offline data. In this work, we ask the question of the dependency on the number of samples for learning from offline data. Our objective is to emphasize that studying sample complexity for offline RL is important, and is an indicator of the usefulness of existing offline algorithms. We propose an evaluation approach for sample complexity analysis of offline RL.
\end{abstract}

% keywords can be removed

\section{Introduction}

Reinforcement Learning (RL) is a powerful framework in solving complex problems. However, applying RL to real-world application is tricky as it needs to actively interact with the environment. In many applications (i.e self-driving car, power-system automation, financial trading, medical trials etc.) it can get very expensive or risky to collect samples in-between training.
Similar to supervised learning, Offline-RL \cite{fittedQ, batchRL} offers a data-driven alternative approach. Offline-RL leverages previously logged data or expert samples and are trained offline without the need to interact with the environment.

Often it is hard to guarantee the quality of the training dataset. Thus in Offline-RL, it is important to benchmark performance \cite{D4RL} with different types of datasets such as; expert, medium-expert, random etc.; to guarantee a reliable performance despite the quality of training samples. But we do not benchmark offline-RL algorithms under \emph{sample complexity}. In Offline-RL we assume not having any constraint in collecting training dataset. For example, in continuous control tasks \cite{D4RL, BCQ, CQL,fisher_RBC, TD3-BC} RL agents are trained with 1 million training samples. But in real-world applications collecting so many samples may not be possible. For more complex tasks, there is no way to quantify how many training sample it may require for the agent to get trained like an expert. In such scenario, there is no-way to provide performance guarantee without letting the agent to perform in the real-world, which again can be very expensive/risky. Thus we need to construct offline-RL algorithms such that it tries it's best to retain performance even with smaller samples.

Several works have proposed offline RL algorithms on standard benchmark tasks, where the assumption is that certain amount of data is always available for learning policies from offline dataset. However, to our surprise, none of the existing works study training and validation performance for offline RL, given its close approximity to a supervised learning setting.

It is shown in \cite{REM, workflow_offlineRL} the offline RL agents exhibit overfitting, i.e., after certain number of gradient updates, their performance starts to deteriorates. \cite{REM, workflow_offlineRL} restores to online performance evaluation to identify the performance drop and early stopping to avoid overfitting. But “true” offline RL requires offline policy evaluation. In this work we show offline RL agent overfits over the expert dataset very early on when trained with smaller number of training samples, i.e., improvement in minimizing the policy training objective gives a false notion of improvement, whereas policy evaluation on validation dataset, which can be done completely offline, indicate agent's actual performance.

% We hypothesize that studying the basic questions of overfitting and underfitting in offline RL, allows us to evaluate the significance of sample complexity when learning from offline data. 

% Our objective is to grab attention of the community to emphasis on sample-complexity so that offline RL methods are more reliable in real-world application

Our key contributions are as follows : 

\begin{enumerate}
    \item We emphasize the importance of sample complexity analysis for offline RL, and compare performance of existing offline RL algorithms by varying the size of training dataset. 
    
    \item We propose that existing works should study overfitting and validation performance of offline RL algorithms that can be computed completely offline. Our comparison of the offline evaluation on the validation set replicate the policy performance trend of the online policy evaluation in MuJoCo continuous control tasks. Thus this provides insights on the offline RL algorithms performance, especially important when applied in the real-world applications.
    
    \item Our empirical findings show that while existing offline algorithms can work really well under the standard benchmark size of training samples, the performance of these algorithms is quite different when studies under a low data regime. This indicates that certain algorithms are more likely to overfit than others. Along with data-diversity, sample-complexity analysis further validates agents reliability and robustness.

    % \item Sample complexity analysis in Offline-RL algorithms is important to guarantee performance in real-world application.
    % \item We also show, with smaller expert-samples, the algorithms overfits over given samples. Thus while training loss can give a false sense of improvement, we propose using validation dataset that indicates policy's actual performance.
\end{enumerate}

In this work, we emphasize the importance of sample complexity analysis for offline RL algorithms, which has perhaps been overlooked in existing studies. By ranging from a large data regime to a small data regime, we show that the performance of different offline RL algorithms is not always consistent across benchmark tasks. To further clarify our studies, we propose a training and validation split for offline RL, akin to the basic supervised learning problem, and find that different algorithms have different overfitting properties given the same algorithm complexity in terms of the policy and value functions. This suggests the importance of sample complexity analysis for offline RL, clearly showing that the existing performance metric may not always be a good indicator of the usefulness of an offline RL algorithm, especially when the goal is to take offline RL to real world applications.

% In this work we compare offline-RL algorithm under sample complexity in continuous control tasks to show why this is an important metric for benchmarking Offline-RL algorithm. Furthermore, we show how a validation dataset can provide better guarantee/indication of algorithms true performance without performing evaluation in the real-world. 

% % set the landscape why this is important
% Offline Reinforcement learning (Offline-RL) leverages expert-data and gets trained offline without interacting with the environment. But the assumption is we have large repository of expert samples, which may not always be the case.
% % objective of the paper & how we plan to reach to that objective
% We do empirical study of the recent offline-RL algorithms with smaller expert sample and study how reliable they are under such scenario. Our objective is to grab attention of the community to emphasis on sample-complexity so that offline RL methods are more reliable in real-world application
% % finding 
% Our analysis suggests, with smaller expert samples, performance drops dramatically despite decaying loss functions. And the best performing offline agent does not guarantee to the best under sample complexity. Thus we propose sample complexity to be a metric to guarantee a more reliable offline RL algorithms for all applications.
% % and what does the finding suggests
% For better understanding of the offline-RL performance we propose using validation dataset since the loss curve gives a false sense of improvement. 

\section{Preliminaries}
We consider learning in a Markov decision process (MDP) described by the tuple ($S, A, P, R$). The MDP tuple consists of states $s \in S$, actions $a\in A$, transition dynamics $P(s'|s,a)$, and reward function $r=R(s,a)$. We use $s_t$, $a_t$ and $r_t=R(s_t,a_t)$ to denote the state, action and reward at timestep t, respectively. A trajectory is made up of sequence of states, action and rewards $\tau=(s_0, a_0, r_0, s_1, a_1, r_1, ..., s_T,a_T,r_T)$. For continuous control task we consider an infinite horizon, where $T=\infty$ and the goal in reinforcement learning is to learn a policy which maximizes the discounted expected return $\mathbb{E}[\sum_{t=t'}^T \gamma^t r_t]$ in an MDP. In offline reinforcement learning, instead of obtaining data through environment interactions, we only have access to some fixed limited dataset consisting of trajectory rollouts of arbitary policies.  
%This setting is harder for agent to explore the environment and collect additional feedback.
%\Riashat{backgrund from BCQ and CQL papers}
\section{Sample Complexity in Offline RL}

\textbf{Sample Complexity : }An important concept for our analysis is to define \emph{sample complexity}. In general, by finding the sample complexity of any algorithm we refer to the number of training samples required to learn a good approximation of the target. But for complex task, especially in infinite state-action space it's not trivial to define this quantity, for our analysis we refer \emph{sample complexity} as to sensitivity of the algorithms to training sample size. %\Riashat{need to expand this further}

% introduce experiment setup: 1. env 2. algorithms 3. expriment setup
\textbf{Experiment Setting : } 
In this section, we describe our framework and experimental pipeline for evaluating the sample complexity for different offline RL algorithms. We investigate sample complexity in continous control benchmark tasks, based on the D4RL dateset \cite{D4RL} which is considered as a standard dataset for most offline RL algorithms. For comparisons, we investigate sample complexity of the following algorithms :  \emph{Batch-Constrained deep Q-learning} (BCQ) \cite{BCQ}, \emph{Behavior Cloning} (BC, implemented in \cite{Bear}) and TD3-BC \cite{TD3-BC}. We run all of our experiments for seed 0-4 and trained for 1M gradient updates. For all the algorithms we use the default network architecture and hyper-parameters. We share our further results in the Appendix. 

\subsection{How does performance vary based on dataset size?}
\label{sec:performance varying datasize}
Given training data, we compare performance for different sizes of the dataset, ranging from $1M$ samples (which is the standard sample size always used), to $100K$ and decreasing to $5000$ samples. 

\begin{figure}[hbt!]
\centering
   \subfloat[]{\includegraphics[width=0.32\linewidth]{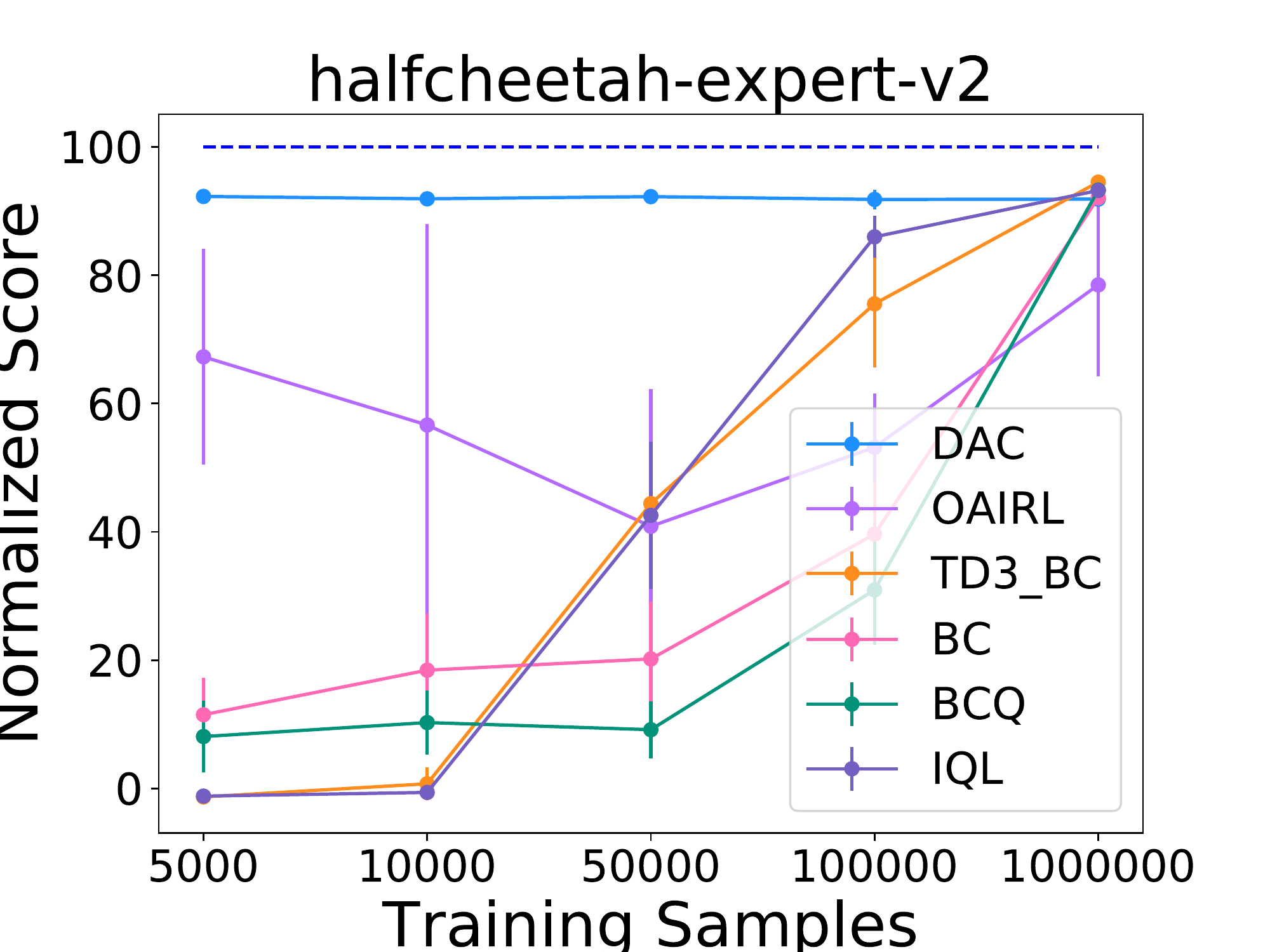}}
   \subfloat[]{\includegraphics[width=0.32\linewidth]{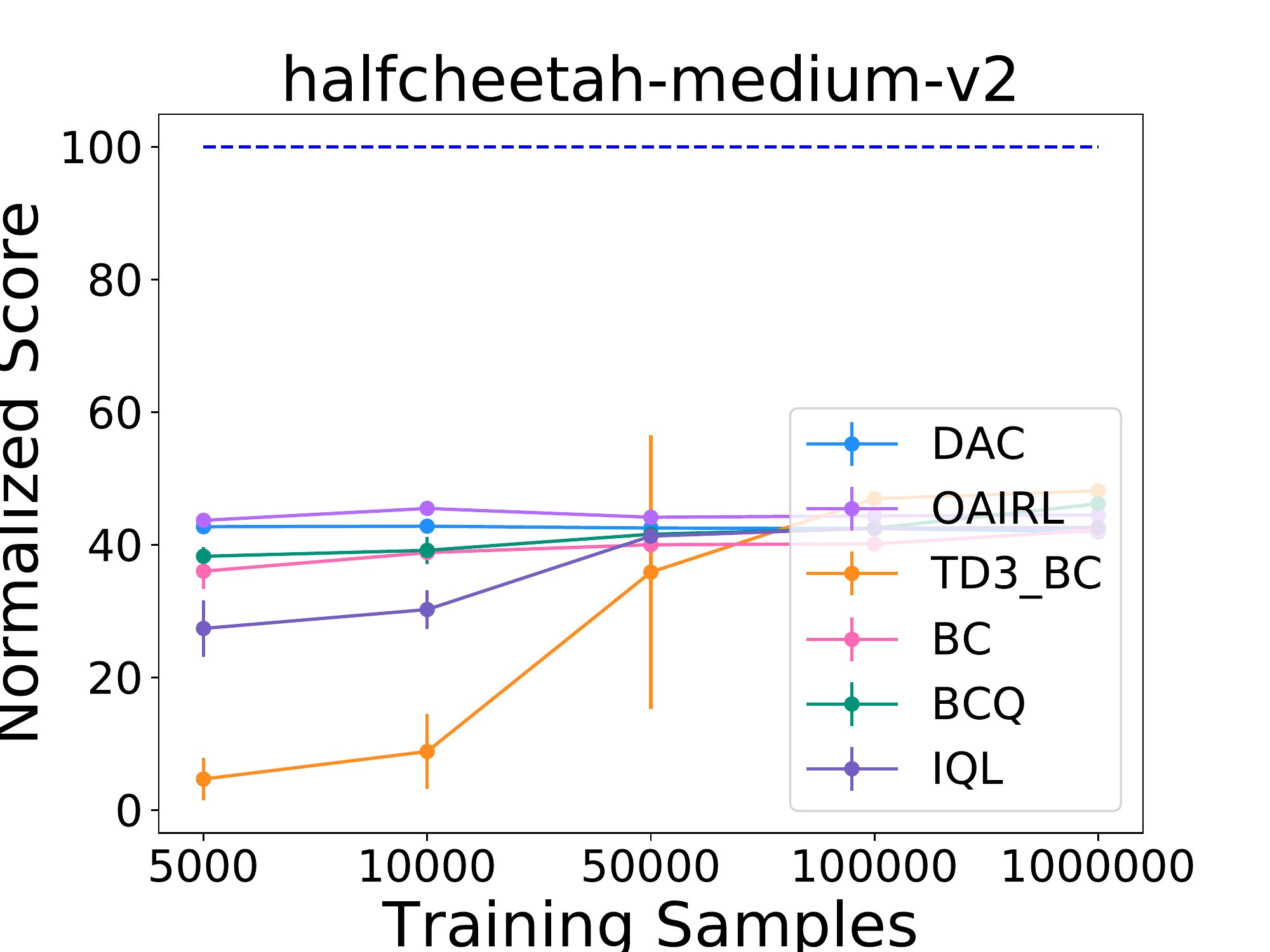}}
   \subfloat[]{\includegraphics[width=0.32\linewidth]{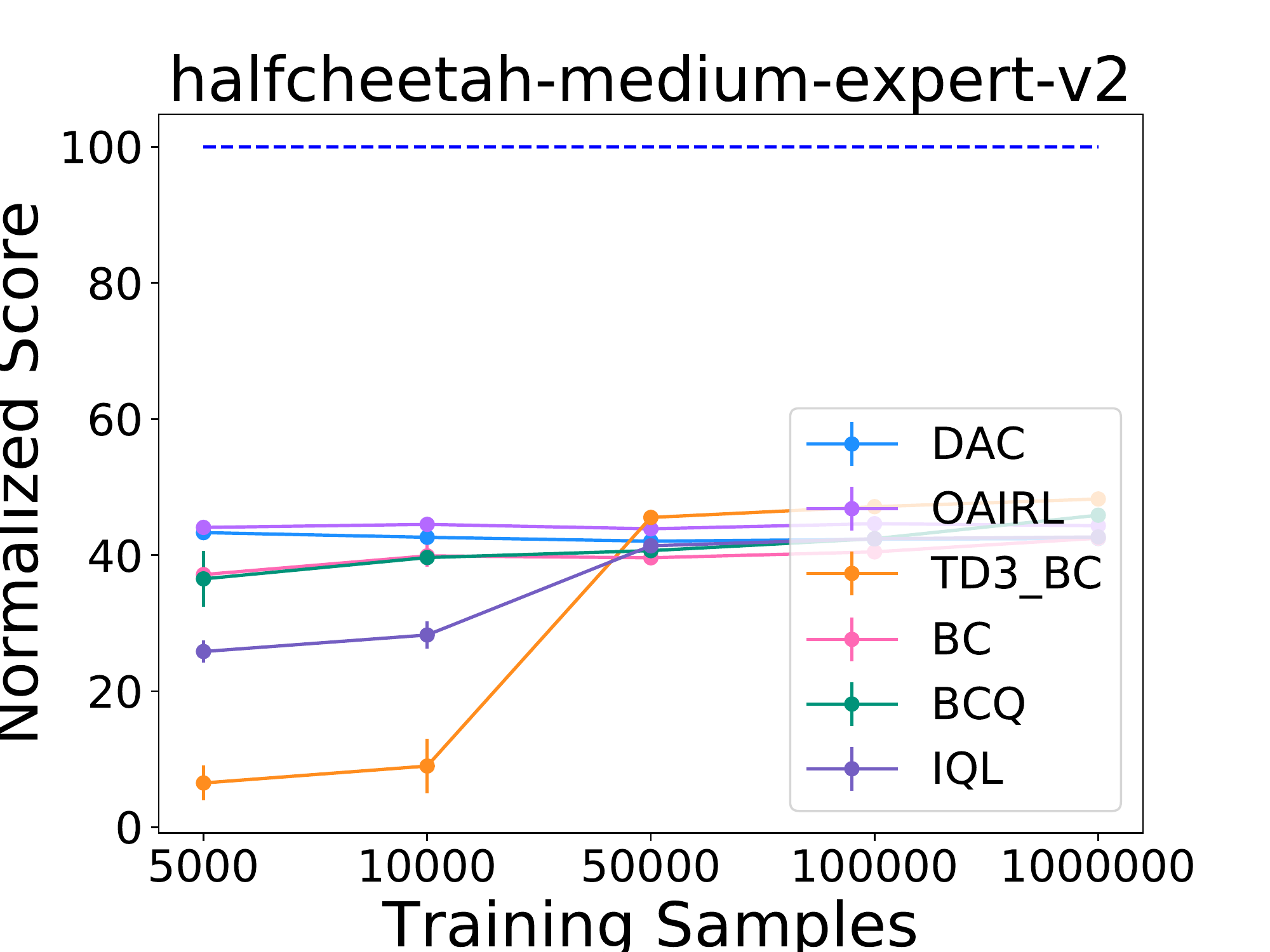}}
   
     \vspace{-0.45cm}
  \subfloat[]{\includegraphics[width=0.32\linewidth]{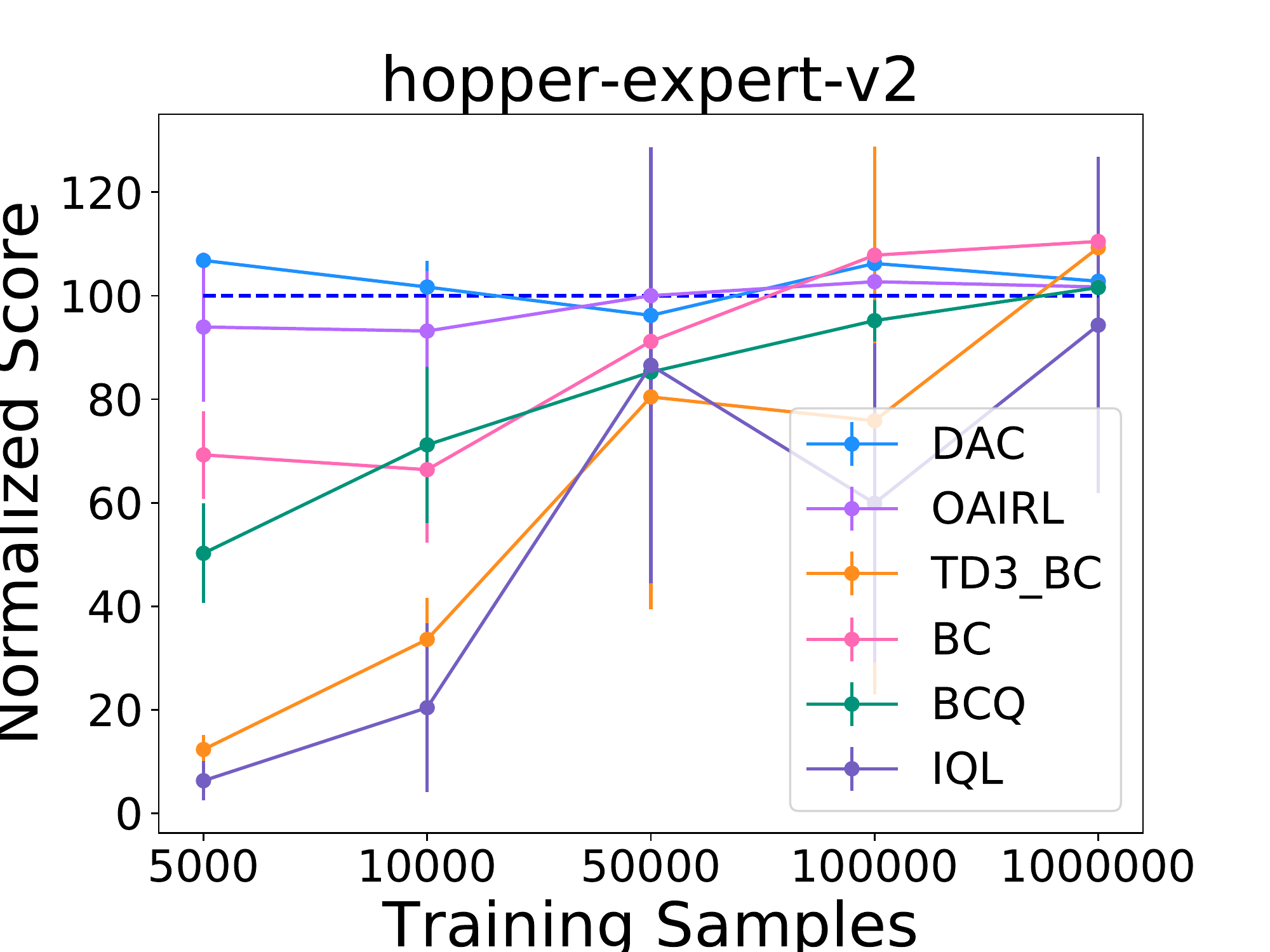}}
    \subfloat[]{\includegraphics[width=0.32\linewidth]{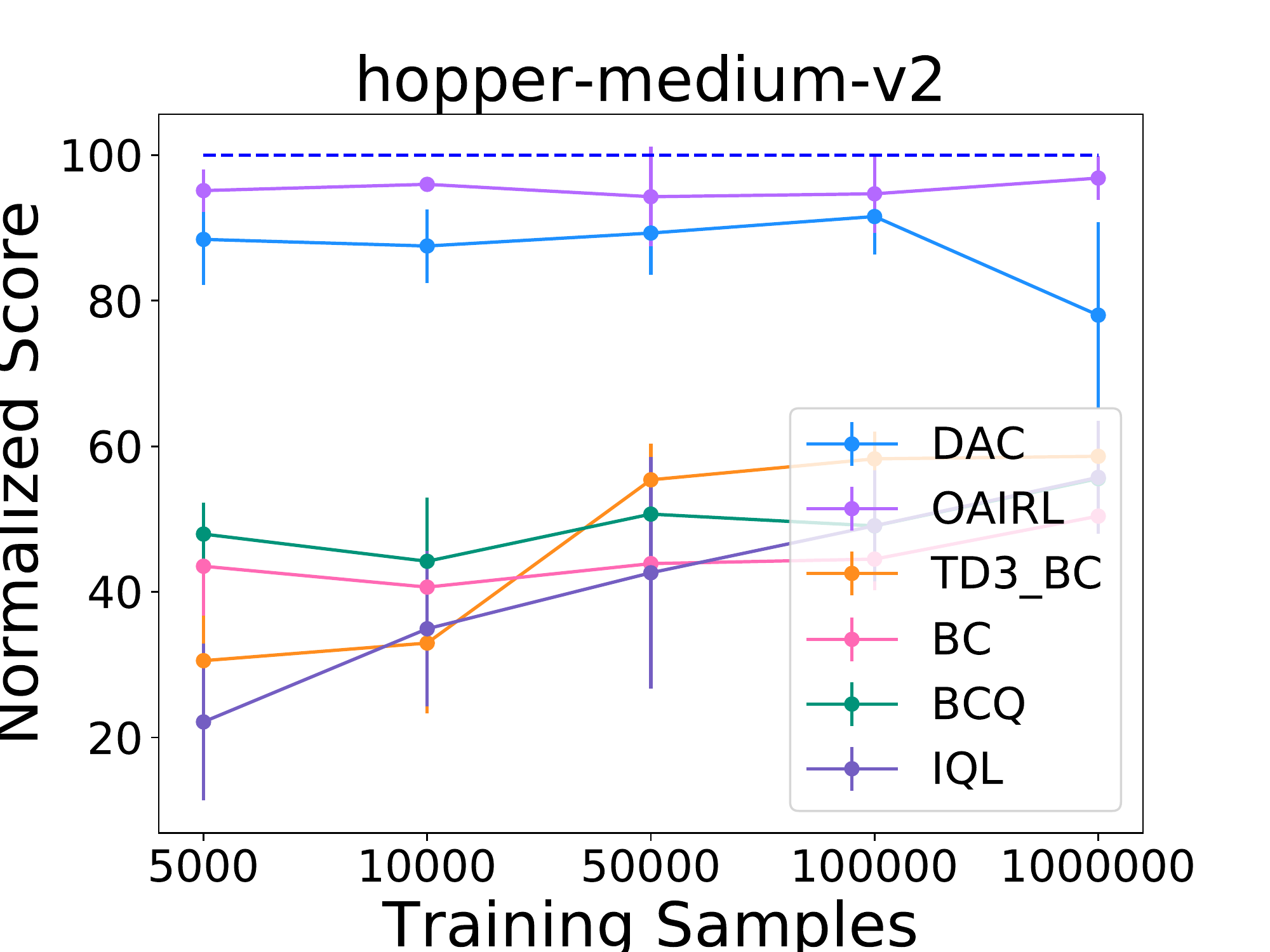}}
   \subfloat[]{\includegraphics[width=0.32\linewidth]{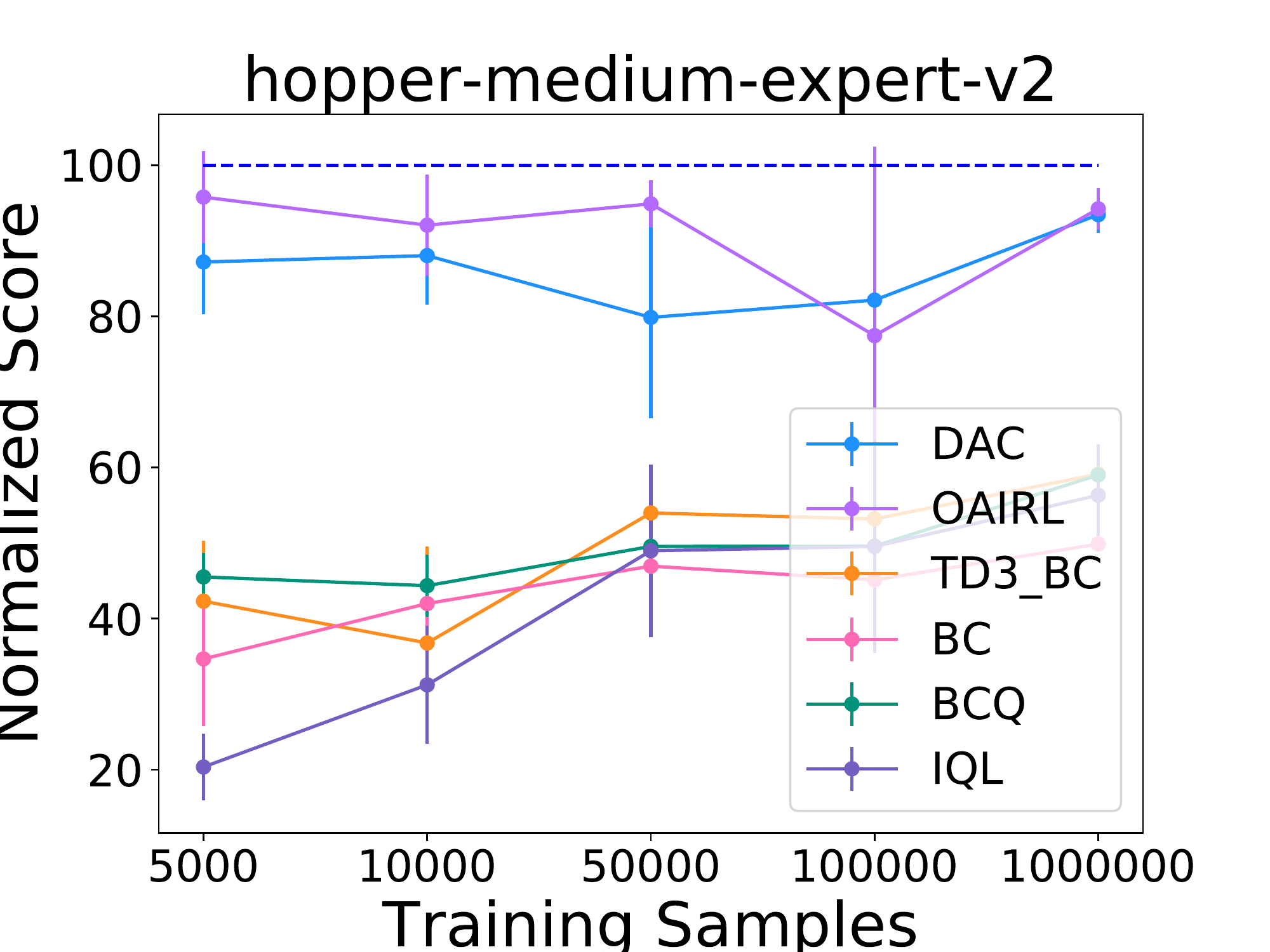}}
   
     \vspace{-0.45cm}
   \subfloat[]{\includegraphics[width=0.32\linewidth]{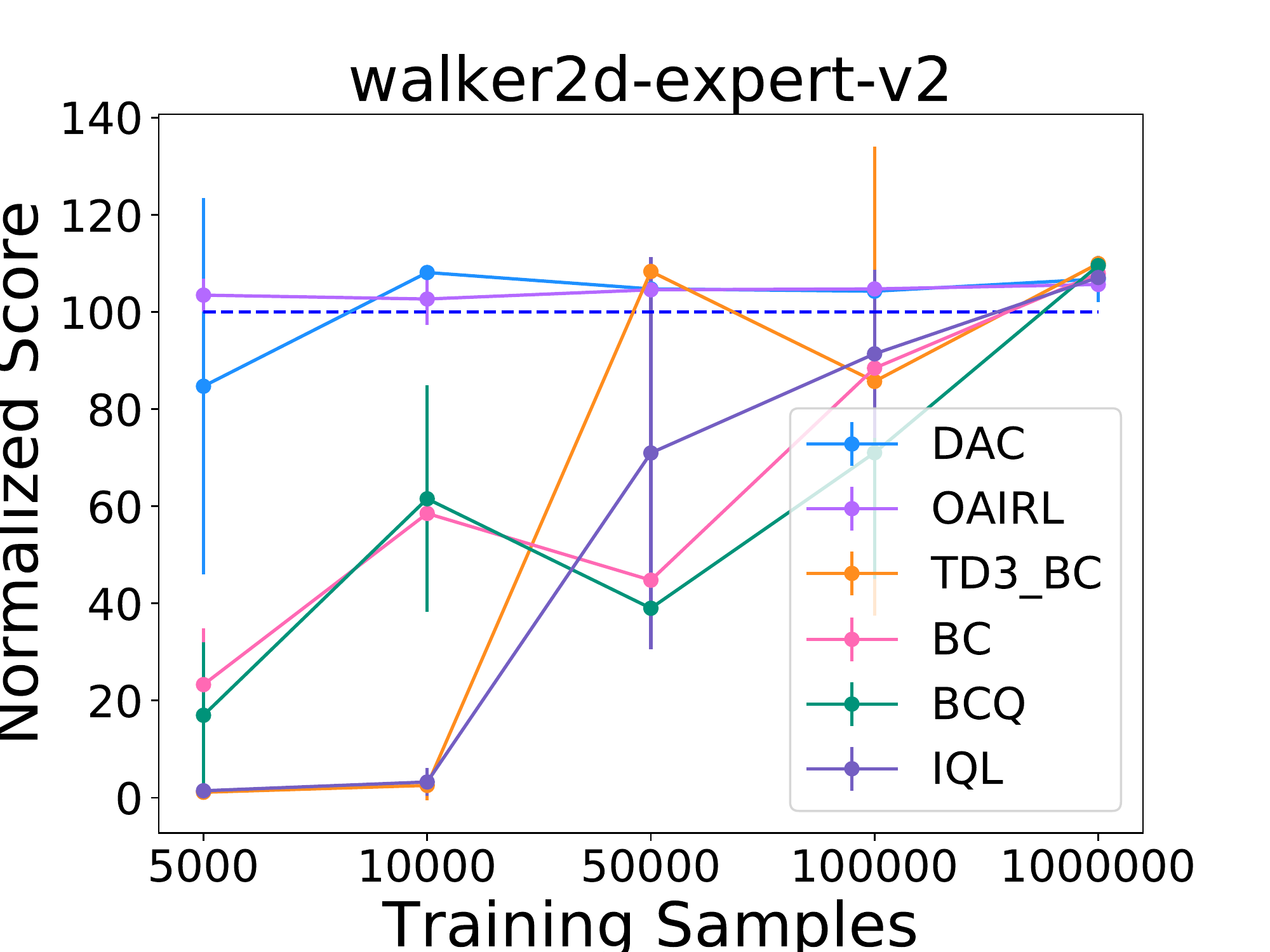}}
  \subfloat[]{\includegraphics[width=0.32\linewidth]{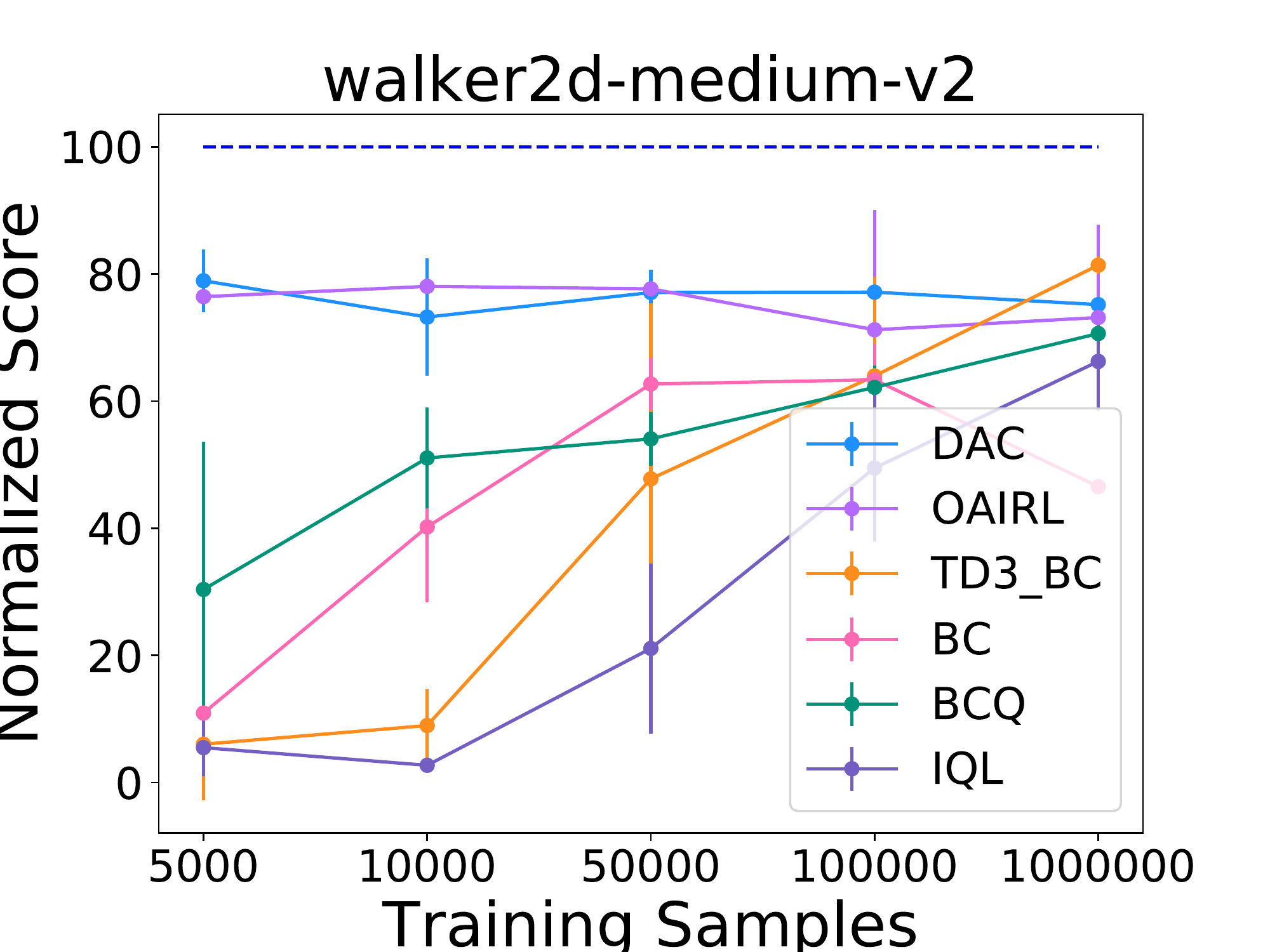}}
  \subfloat[]{\includegraphics[width=0.32\linewidth]{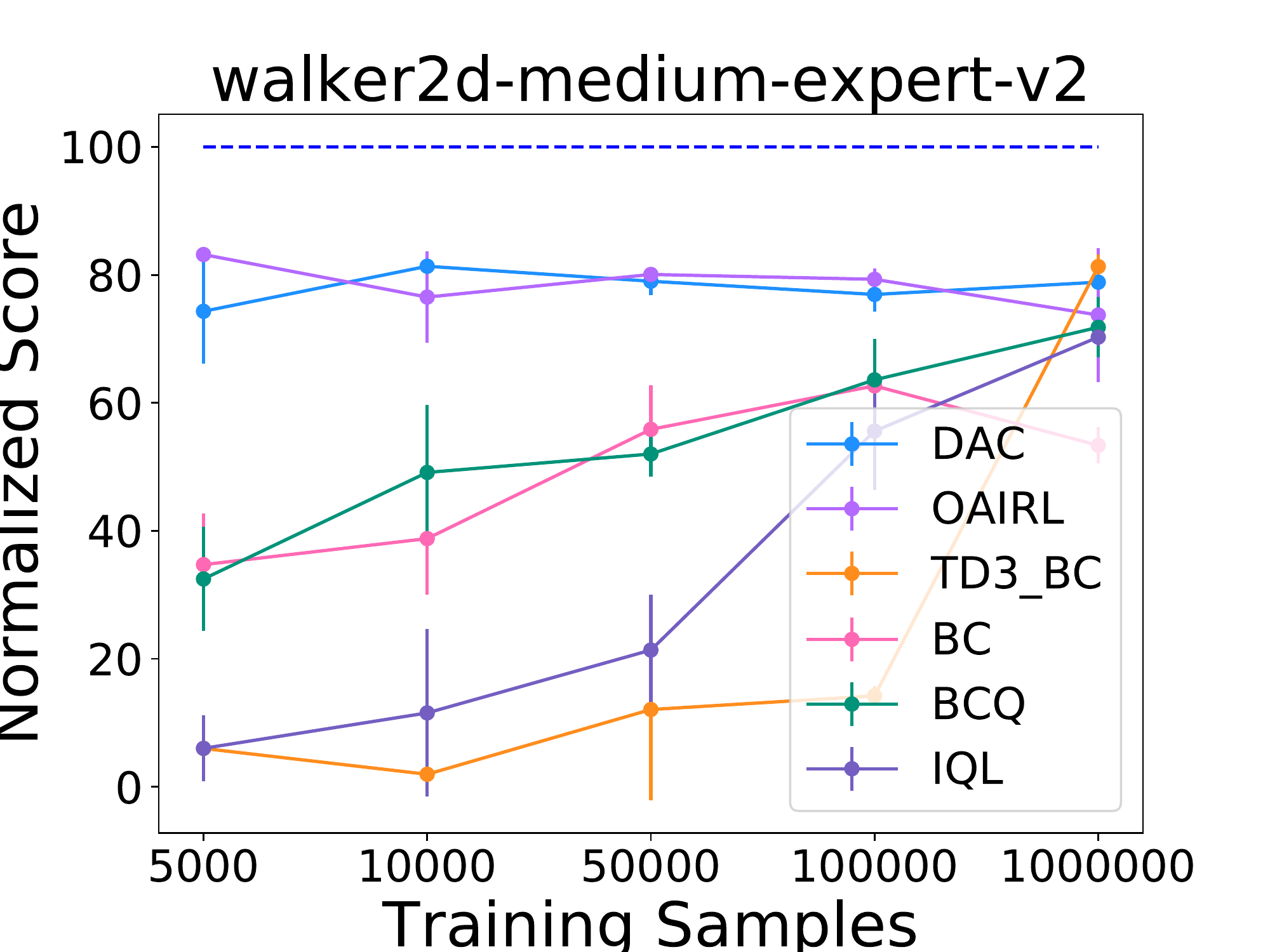}}
\caption{Performance Comparison (D4RL Normalized Score) of DAC with offline-RL Varying Expert data.}
\label{DAC_vs_offline_bar_graph}
\end{figure}

For each of the algorithms and given the training data size, we train for 1M training updates and measure the \textit{normalized score} metric as done in \textit{D4RL} \cite{D4RL}. Experimental results comparing performance dependent on the total number of offline samples is presented in figure \ref{DAC_vs_offline_bar_graph}. %\Riashat{Samin, check the description above}. 

%Note that this is different than using the batch size typically used for training the policy and value networks (we use standard batch size of $32$)

Our experimental results show that the performance drops for each Offline-RL algorithm as we reduce the number of training dataset. For all our offline RL algorithms, we compare the performance with \emph{Discriminative Actor Critic} (DAC) \cite{DAC} - adversarial imitation learning and \emph{Off-policy Adversarial Inverse RL} (OAIRL) \cite{OAIRL} method, which use the same number of expert samples but with the advantage of 1 million environment interactions. The advantage of environment interactions makes the comparison unfair. But the idea is to show, even with smaller expert samples adversarial imitation and IRL methods manages to get consistent performance. Comparetive experiments on these algorithms has proven to be significant later in the paper to support our claim that validation performance always correlates with policy's actual online evaluation improvement discussed in \ref{overfitting_disussion} ( further experiments are in Appendix \ref{appendix:comapre_train_valid} ).

While this is a result that one would typically expect, we find an interesting phenomenon in our results.  Note that the performance varies for each algorithm depending on the training data size. For example, while the recent state-of-the-art algorithm TD3-BC performs significantly better for 1M training sample, this algorithm is in fact worse for $5000$ samples. This phenomenon can be seen in almost all of our experiment in figure \ref{DAC_vs_offline_bar_graph} (except in $1(f)$), where even though TD3-BC performs best for 1M standard sample size, it is the worst performing algorithm as we reduce the size of the dataset. The reason is due to the MSE regularization term in it's actor loss dictates the actor gradient update and thus overfits very easily with smaller training samples and we proof our hypothesis through validation performance in following section \ref{val_performance_offlineRL}. We also see the similar trend in IQL's \cite{IQL} performance but the reason is not so apparent, we need further experiments to hypothesize or come to a solid conclusion. This tells us that the performance of each of these algorithms can vary significantly, and comparisons are not always consistent, as to the best performing algorithm, depending on the training dataset size. This is exactly why we can not guarantee consistent performance with abundant training dataset. 

In offline RL benchmark we compare algorithms on different categories of training samples i.e. expert, medium, medium-expert, random and the intuition is that, in real-world application we can not always guarantee to collect optimal-expert, thus we want to pick an algorithm that guarantees a better performance for any kind of dataset. Similarly, for any real world application there is no way to quantify the \emph{"sufficient amount"} of data that we must collect so that training agent can provide expected performance. Thus we consider \emph{sample complexity}, sensitivity of algorithms performance to training dataset size, as a metric to evaluate the offline-RL performance. An Offline-RL algorithm that give better performance with smaller training samples are more reliable in real-world application than the others. We find the sample complexity analysis to be a very useful metric to evaluate the reliability of Offline-RL algorithm.

\subsection{Does existing offline RL algorithms have overfitting phenomenon?}
\label{val_performance_offlineRL}
 
%  In other words, the different algorithms are likely to overfit in different ways  given the sample size, even though we use the complexity class for each of the policy and value function networks.
%  We hypothesize that algorithms, for example, like TD3-BC are more likely to overfit for small data regime, while algorithms such as BCQ and TD3 are more likely to overfit for large data regime. 

We conjecture that the phenomenon observed in figure \ref{DAC_vs_offline_bar_graph} hints to an overfitting phenomenon for offline RL algorithms. For offline RL, we consider an agent is overfitted over the training dataset when the training objective reduces the divergence between the policy action and expert action over the observed training states and yet fails to provide performance improvement in the oracle (online evaluation).

% agent optimize over the training dataset and fails to provide performance improvement in the oracle (online evaluation). A discrepancy between decreasing training loss and performance improvement is a clear indication of overfitting. 

We emphasize that, to the best of our understanding, no previous works studied similar complexity analysis for different offline RL algorithms. Since most prior works only evaluate performance for 1M sample sizes on D4RL benchmarks, we emphasize that this is not always a good measure, as we see in our analysis in this section. In the subsequent sections, we provide a measure to study the overfitting phenomenon in offline RL, and want to emphasize the readers, that since the goal of offline RL is similar to supervised learning, such characterization of overfitting and sample complexity is necessary for any offline RL algorithm empirically. 

\subsubsection{Evaluating Overfitting in Offline RL}
\label{overfitting_disussion}
% how do we prove hypothesis
To prove our overfitting hypothesis, similar to supervised learning, we propose to use separate validation dataset. 
% how do we get this validation set
We held-out $2000$ expert trajectories (which is approximately $2000,000$ $\{s_{V},a_{V},r_{V},s^{'}_{V}\}$ tuples) from the D4RL dataset \cite{D4RL} during training. We perform evaluation over the validation dataset, which provide an unbiased and the true progress of the learning agent. 
% Since we have the complete expert trajectories, we compute the true value estimation $Q^E(s_{V},a_{V})$ (expected cumulative future discounted reward) w.r.t the expert and we compare the critic's action-value estimation $Q\phi(s_{V},a_{V})$ to evaluate the critic.

% bar graph - True actor loss

\begin{figure}[hbt!]
\centering
  \subfloat[]{\includegraphics[width=0.32\linewidth]{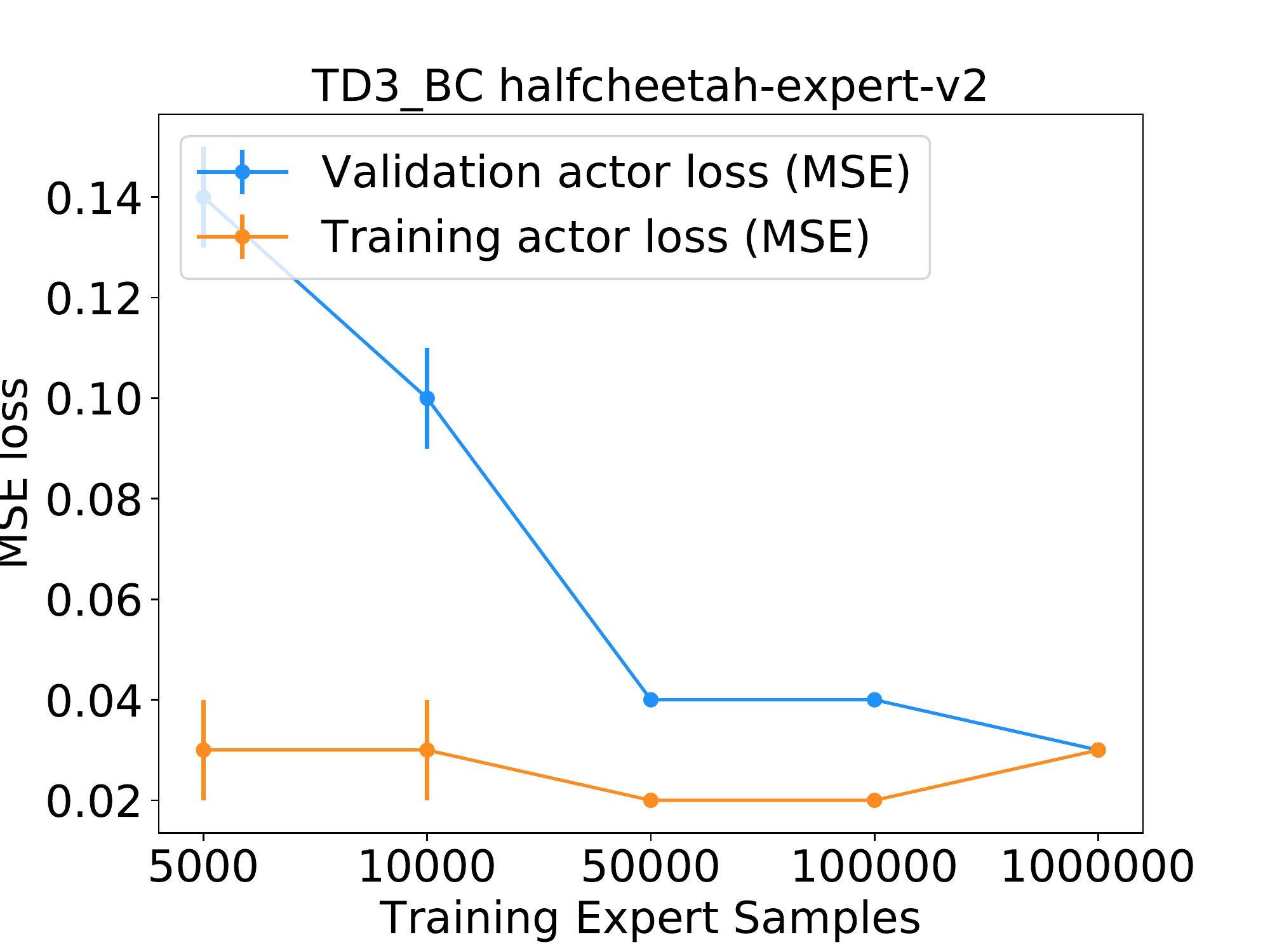}}
    \
    \subfloat[]{\includegraphics[width=0.32\linewidth]{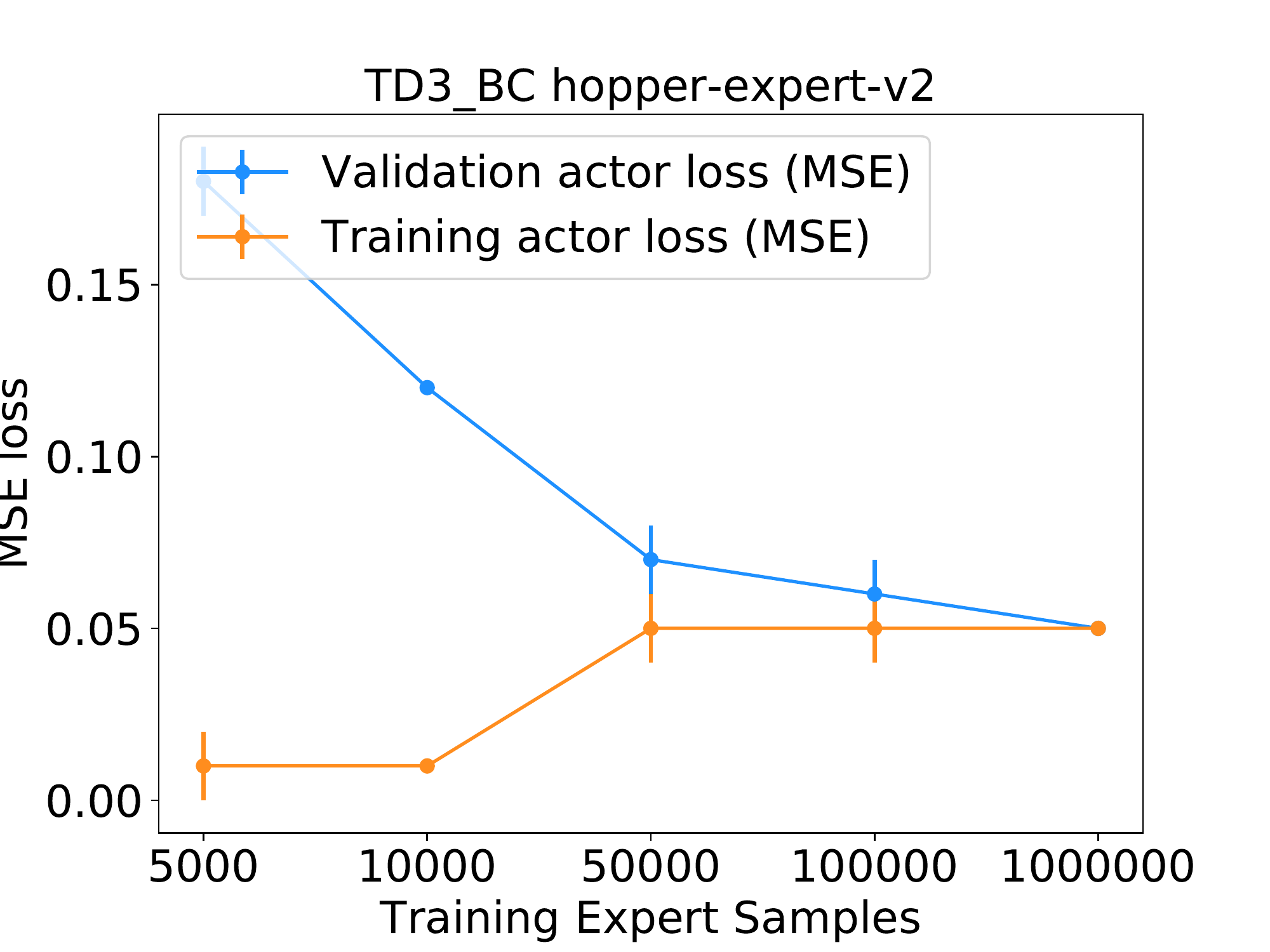}}
    \
    \subfloat[]{\includegraphics[width=0.32\linewidth]{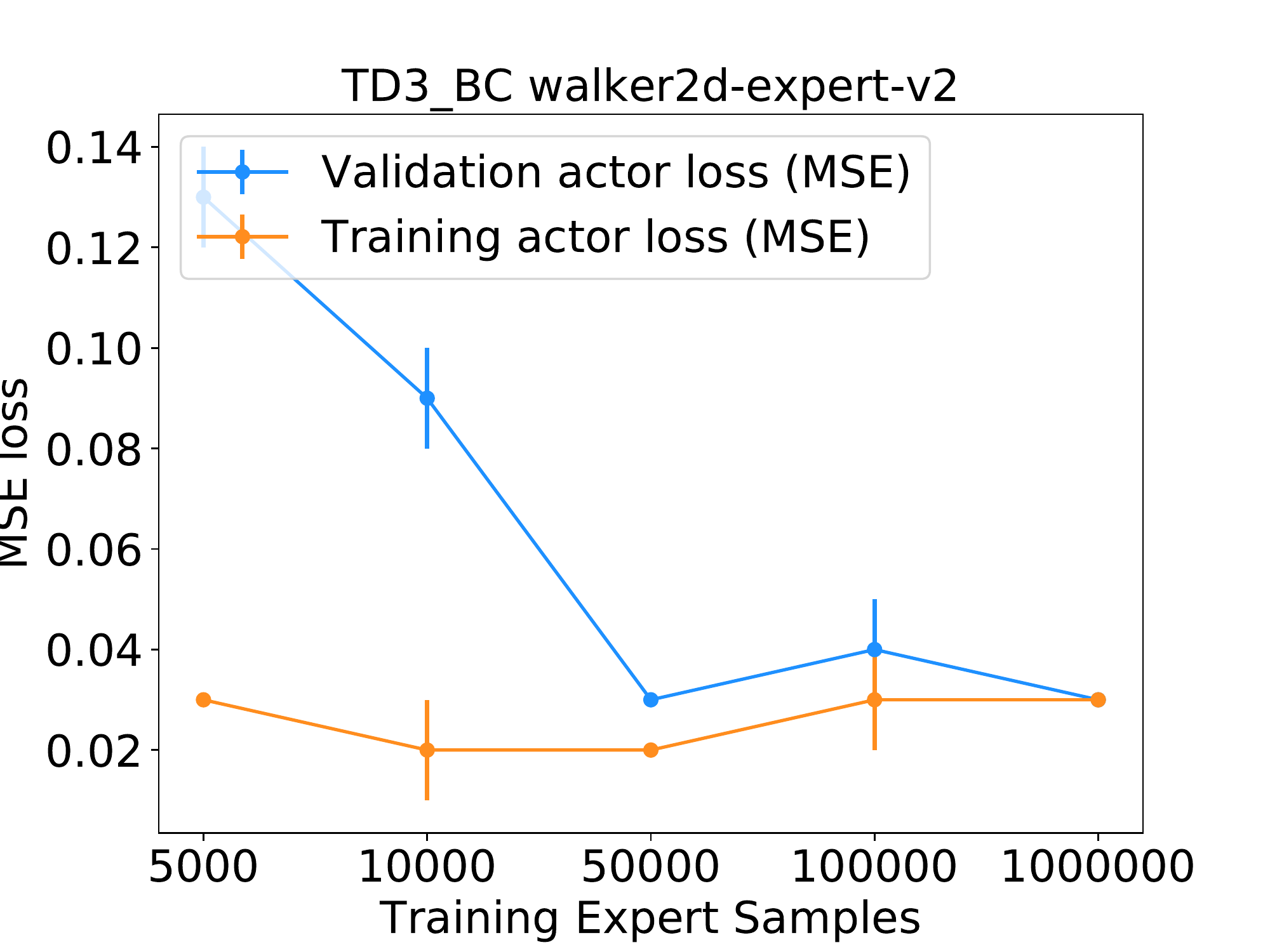}}

        \vspace{-0.45cm}
 \subfloat[]{\includegraphics[width=0.32\linewidth]{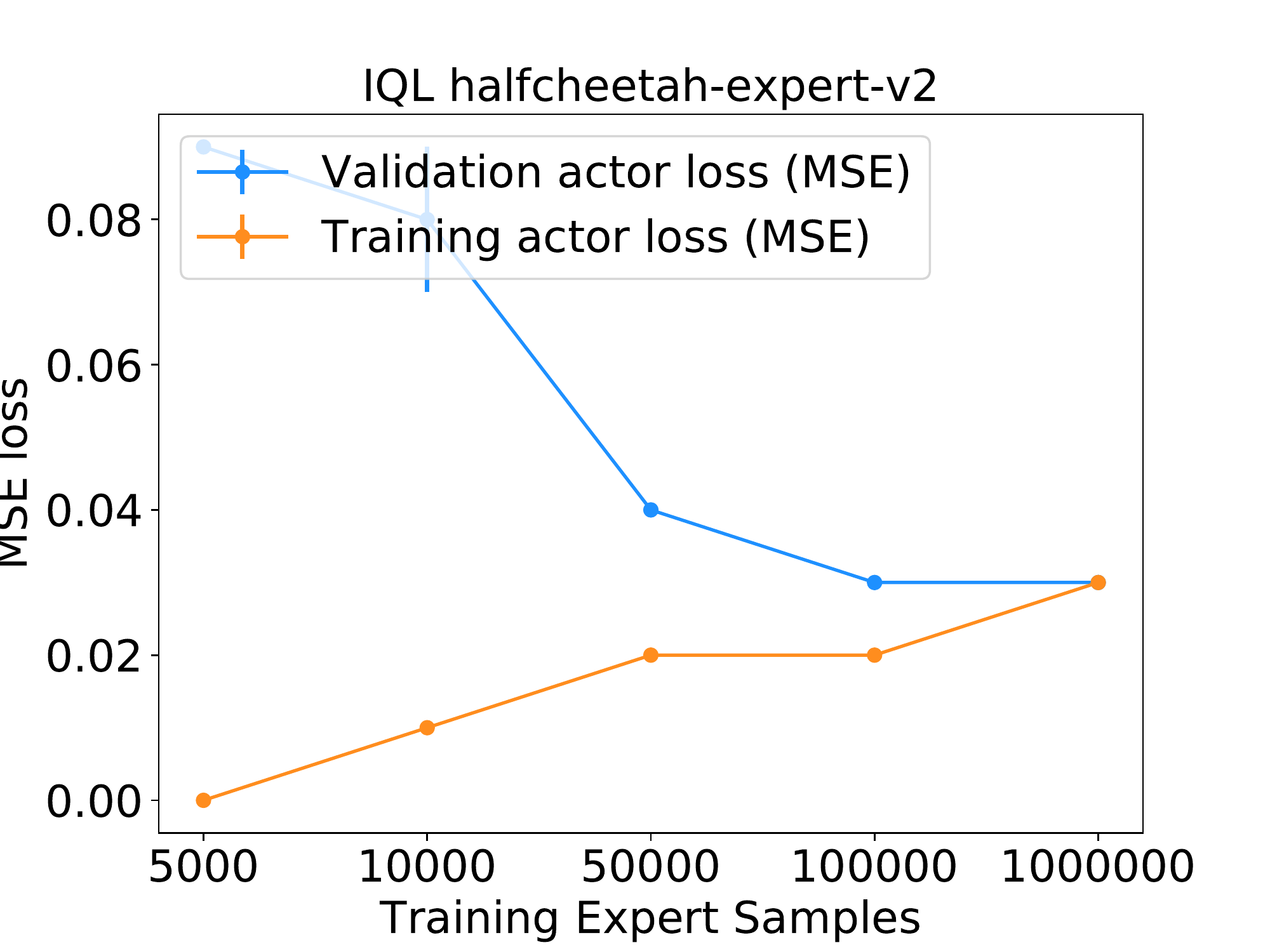}}
 \
  \subfloat[]{\includegraphics[width=0.32\linewidth]{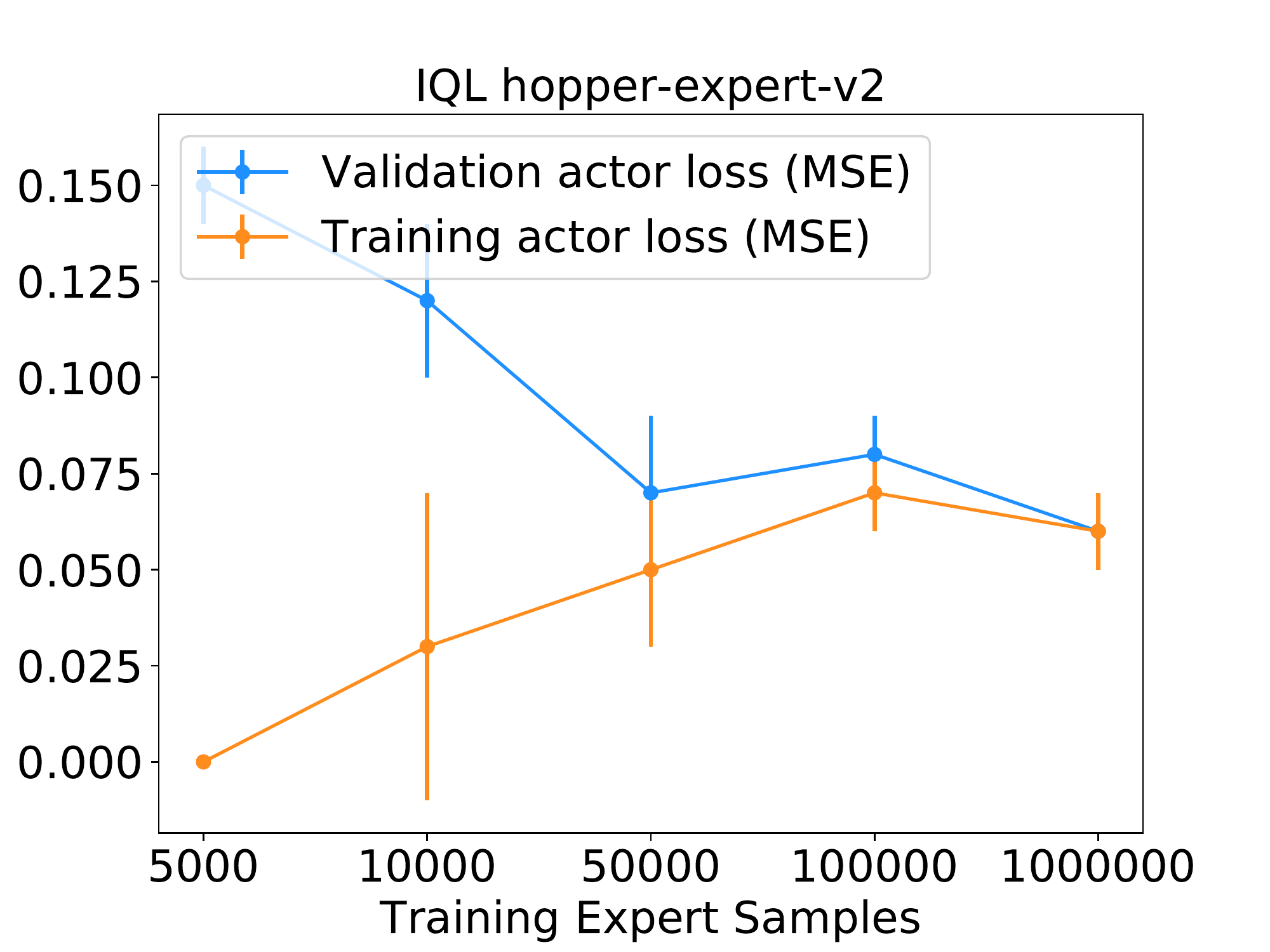}}
 \
  \subfloat[]{\includegraphics[width=0.32\linewidth]{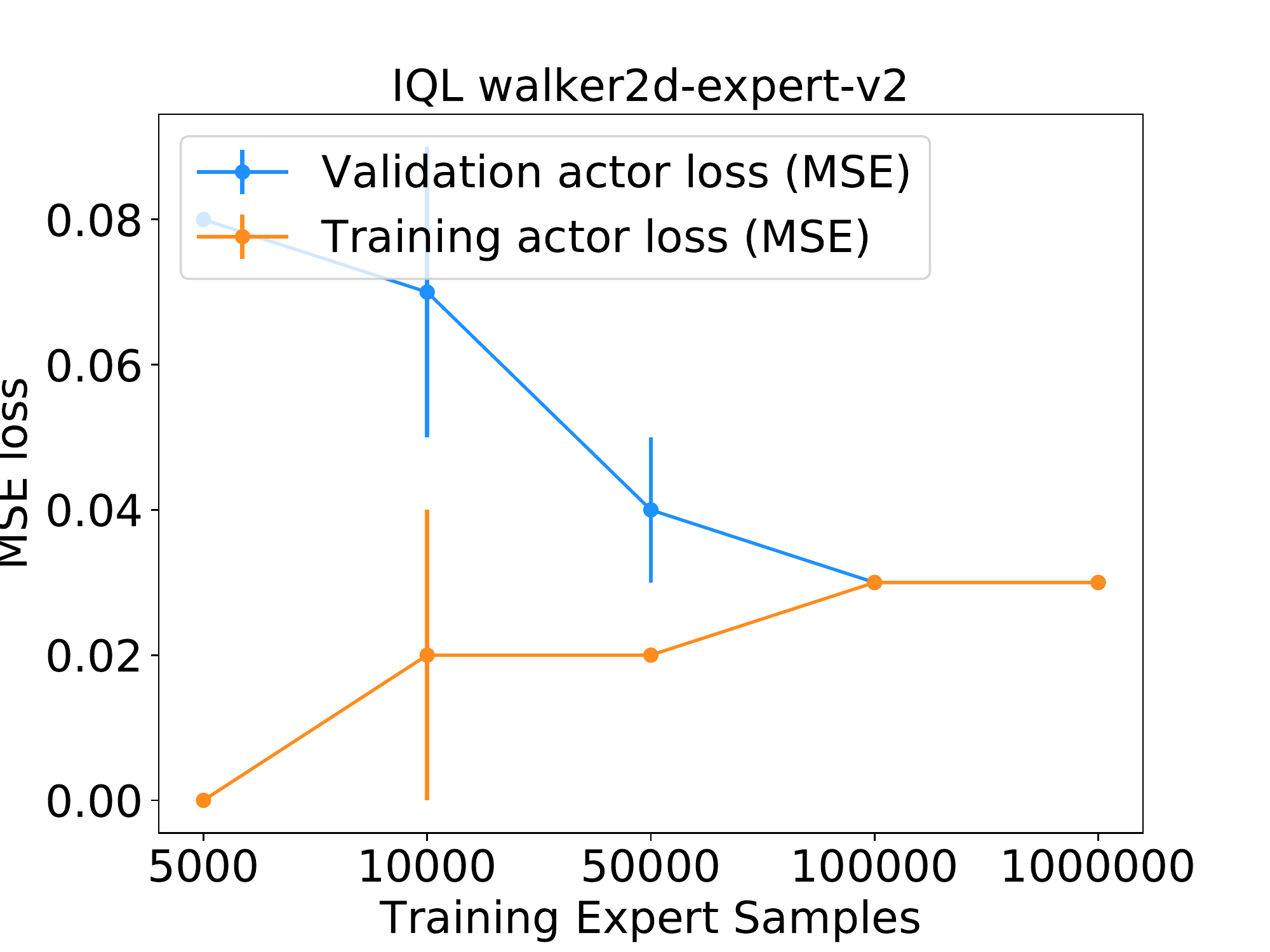}}
    
    \vspace{-0.45cm}
 \subfloat[]{\includegraphics[width=0.32\linewidth]{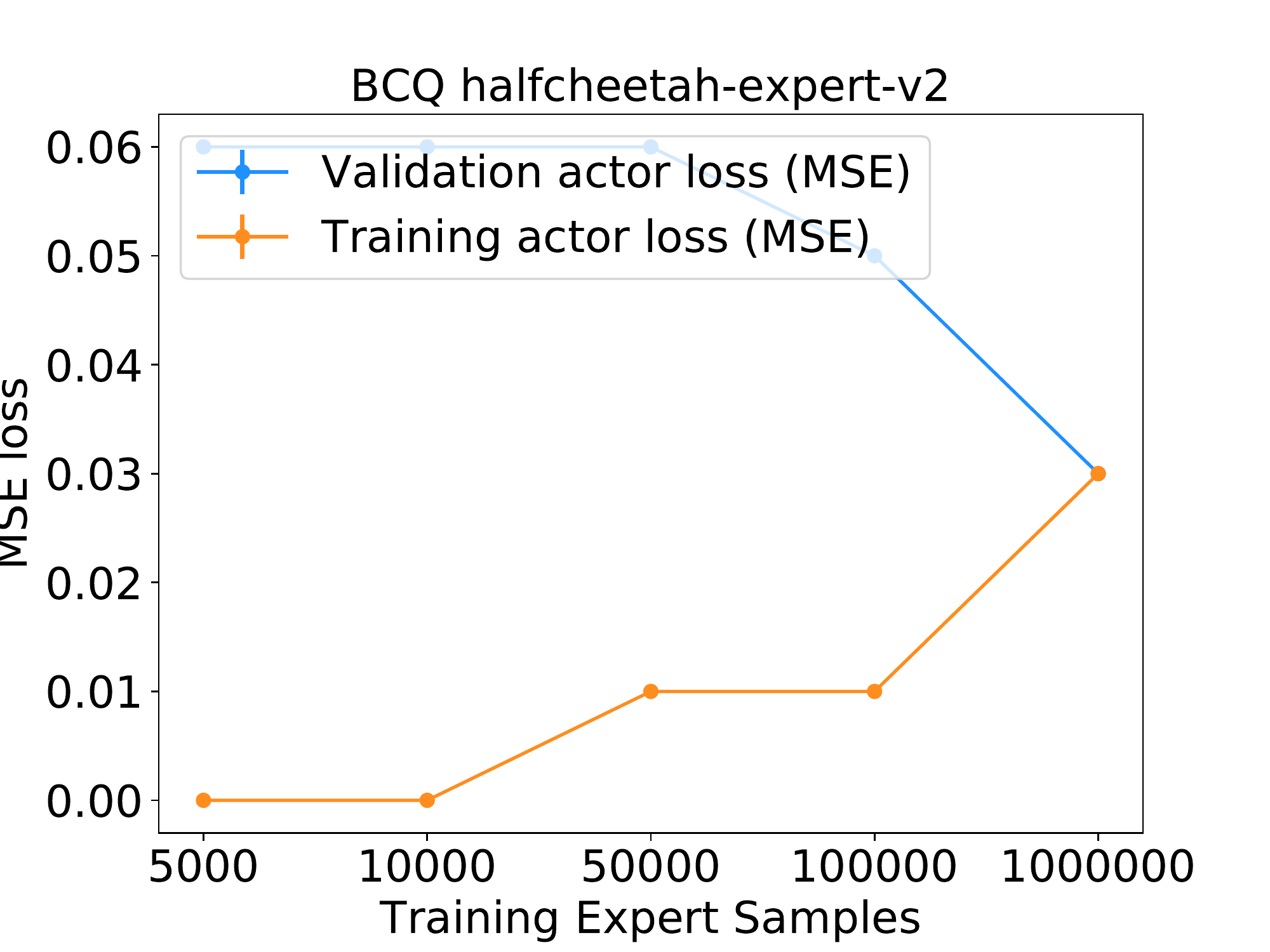}}
 \
  \subfloat[]{\includegraphics[width=0.32\linewidth]{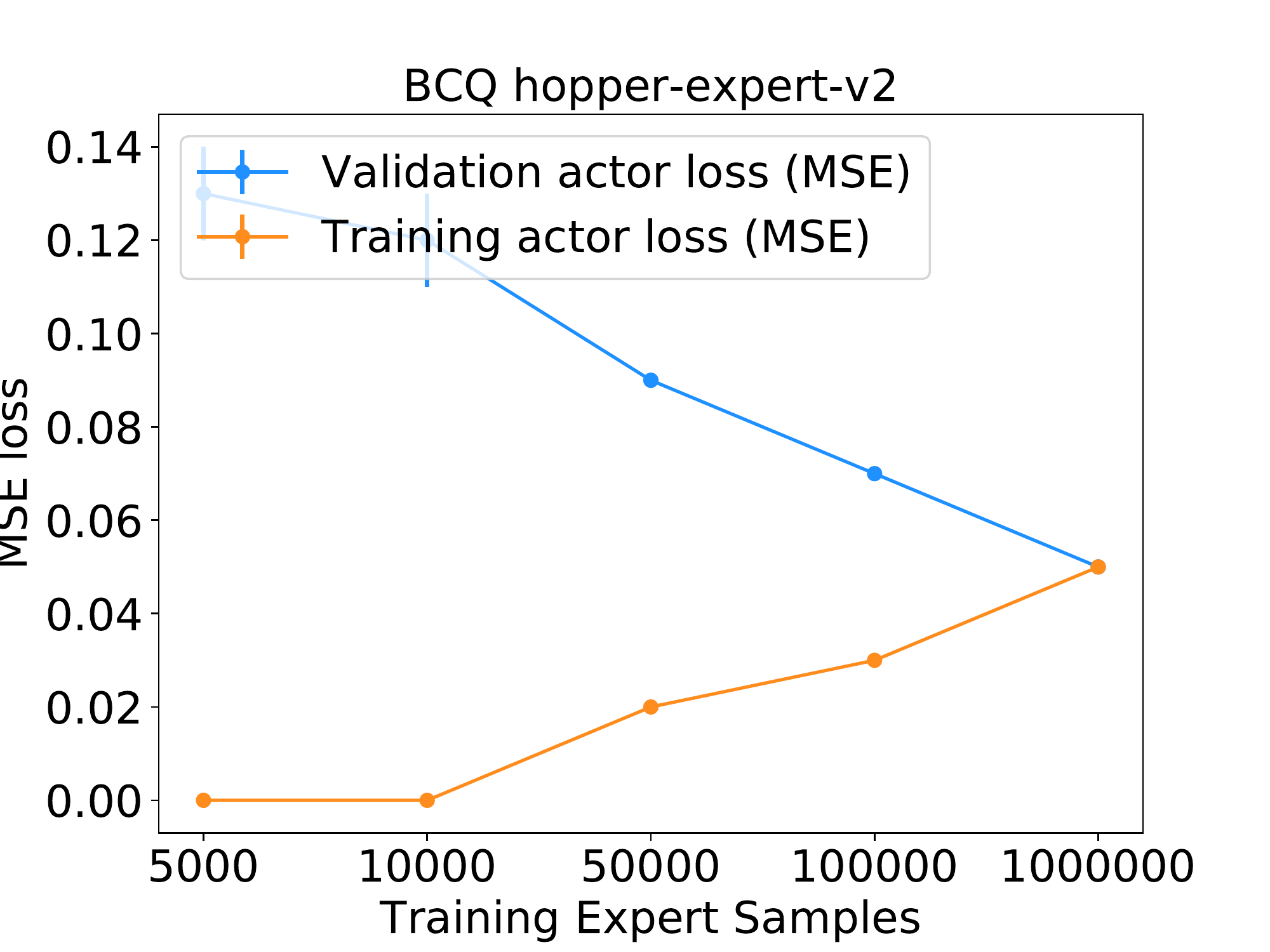}}
 \
  \subfloat[]{\includegraphics[width=0.32\linewidth]{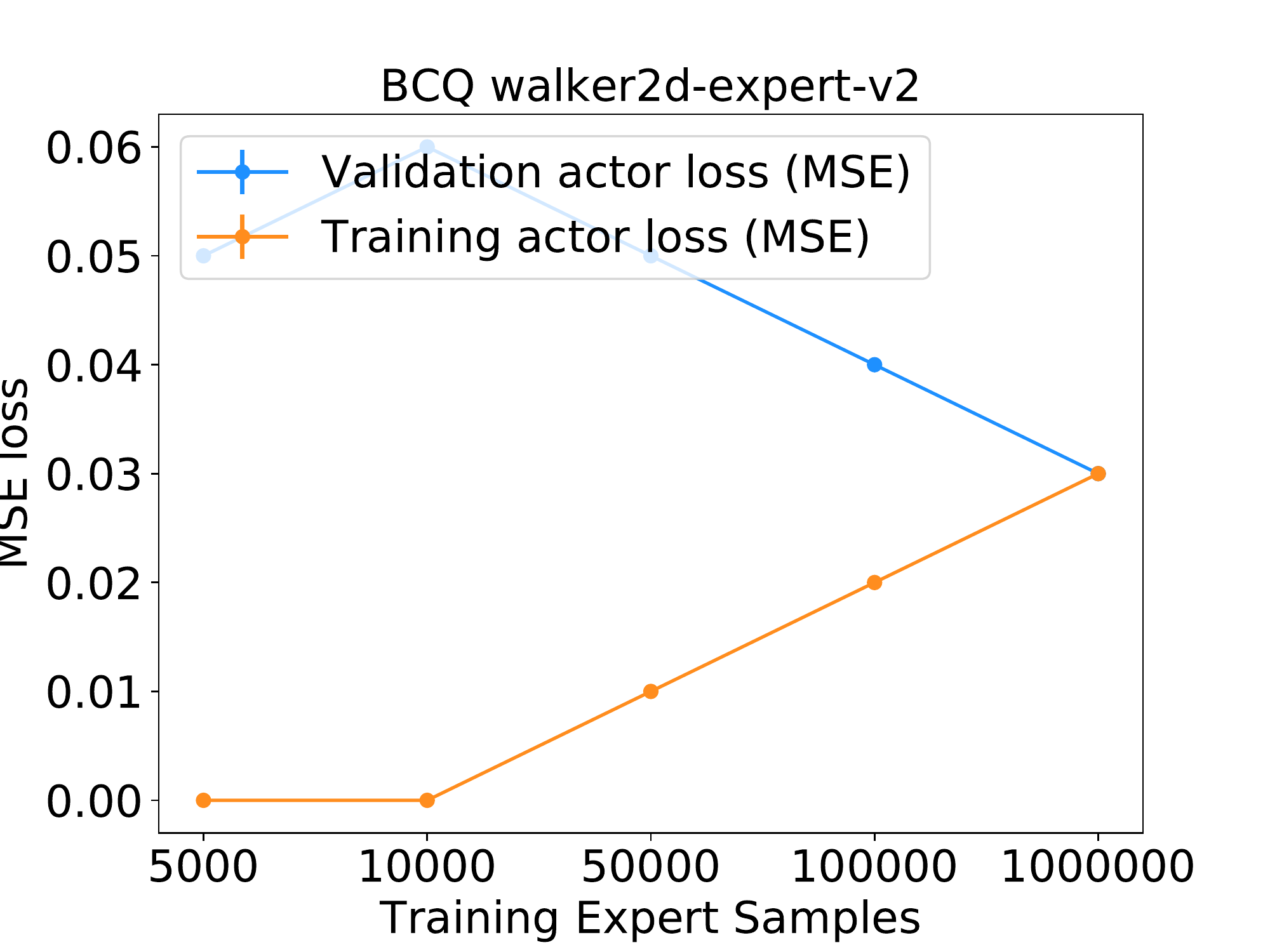}}
 
   \vspace{-0.45cm}
\subfloat[]{\includegraphics[width=0.32\linewidth]{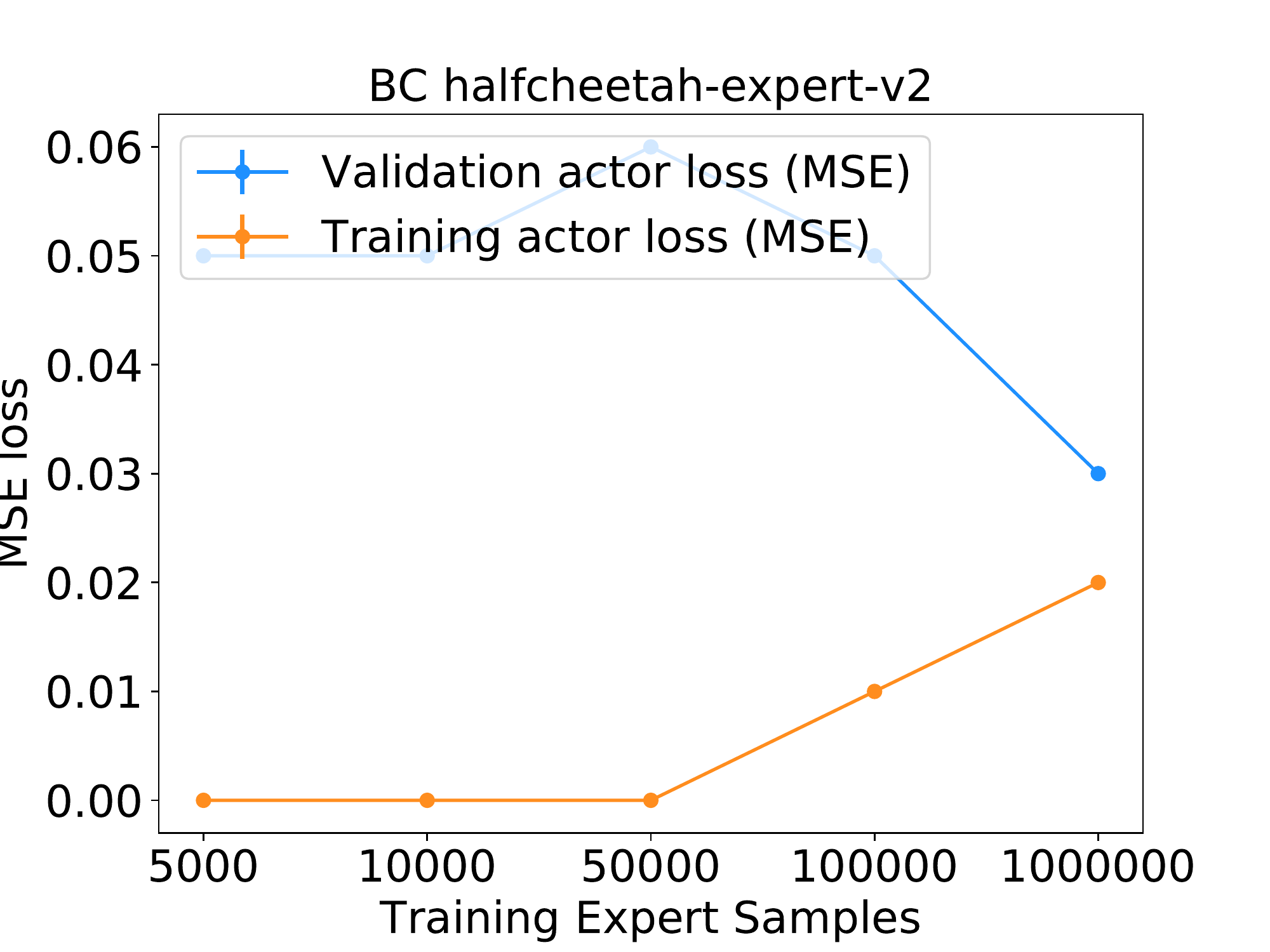}}
\
\subfloat[]{\includegraphics[width=0.32\linewidth]{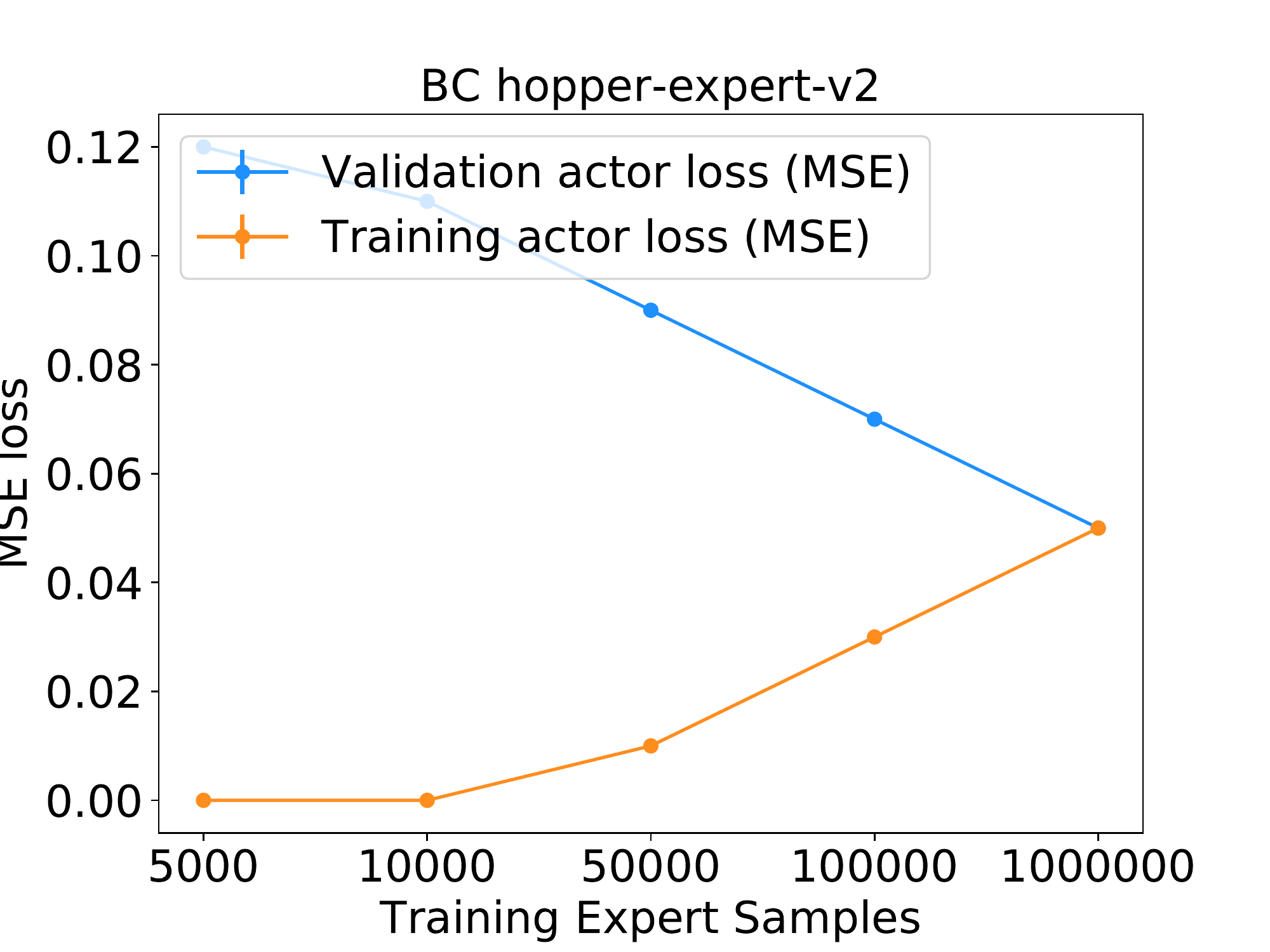}}
\
\subfloat[]{\includegraphics[width=0.32\linewidth]{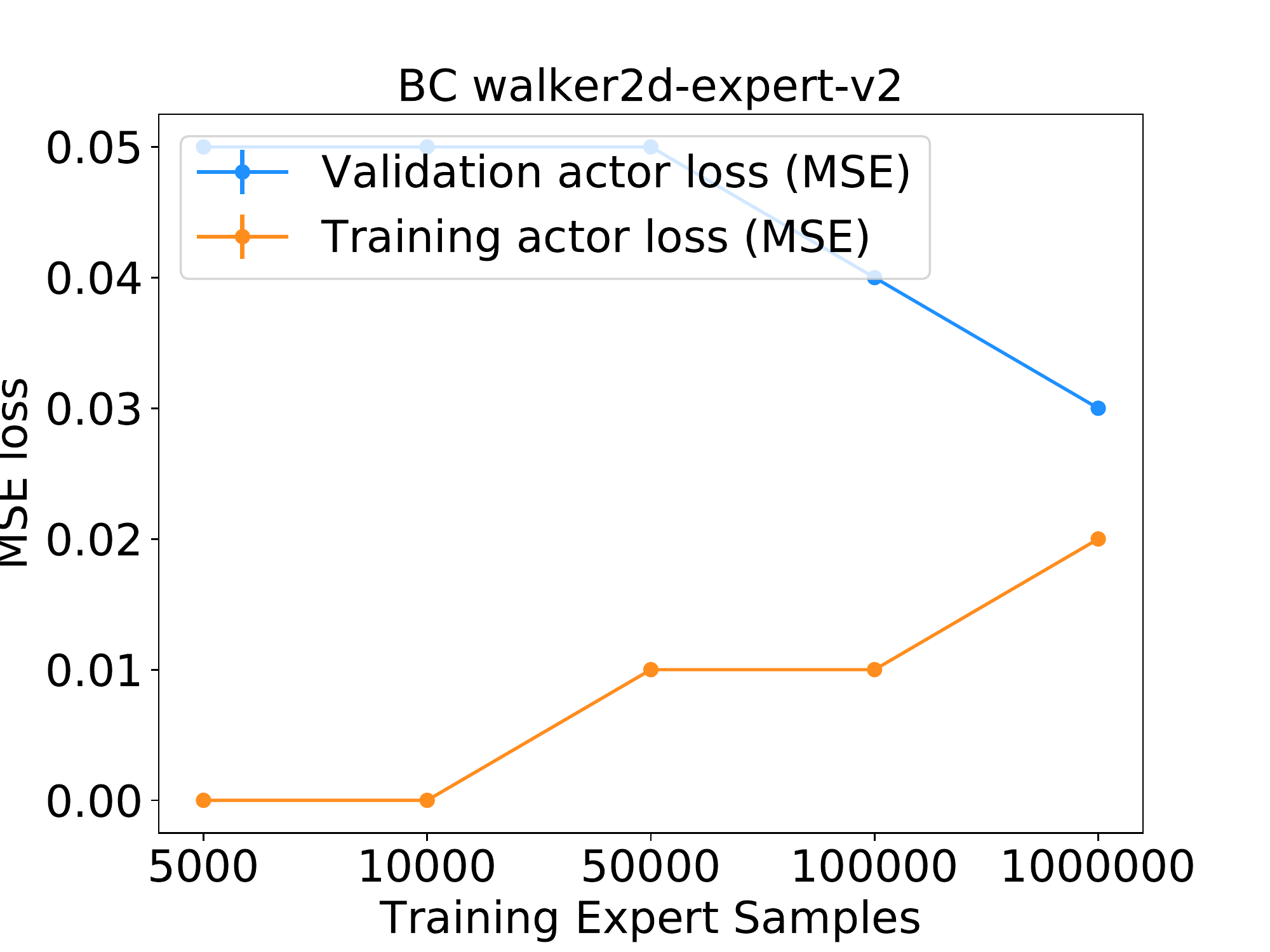}}
\caption{ MSE loss between $\pi_\theta(s_E)$ and $a_E$ for different Offline-RL algorithms over the training (orange) and the validation (blue) dataset as we vary number of training expert samples}
\label{bar_graph_true_actor_loss_train_validation}
\end{figure}

% -------------------------------------------
\textbf{Metric on Training and Validation Dataset : } 
\label{actor_train_valid_performance}
We provide a metric for measuring training and validation performance in offline RL, akin to the standard loss typically studied in supervised learning. As an evaluation criterion, we use the Mean-Square-Error (MSE) loss between expert-action $a_{V}$ and policy-action $\pi_\theta(s_{V})$ as to measure actor's deviation from the expert. Note that we use MSE instead of the KL divergence metric here, since most offline RL algorithms that we study are based on deterministic policies, as typically in BCQ \cite{BCQ} and other algorithms. 

Figure \ref{bar_graph_true_actor_loss_train_validation} shows the overfitting phenomenon for different offline RL algorithms. We plot the MSE loss over the training and validation dataset, and vary the sample size. For each algorithm, we train up to 1M iterations (as typically done in standard experiments), but with different training data sizes. We find that as the training data decreases, the difference between training and validation error increases significantly, which shows that the algorithms are more likely to overfit (due to a more complex policy class compared to the dataset size).

% give a context how we know model is overfitting
We get a good generalization in estimation when we make improvement in estimating both the training and validation dataset. We know our training model is overfitting over the training the dataset when the training loss gets reduced with each gradient update but the performs worse on the validation set.
For 1 million expert samples, algorithms performs lowest validation error. For 5000 training dataset the Actor gets the lowest training (orange) error but gets the highest validation error. It suggests that the Actor overfits the expert samples and we see the consequence in the policy performance (figure \ref{DAC_vs_offline_bar_graph}). 

The largest deviation in training-validation performance in found for TD3-BC. This confirms our hypothesis for TD3-BC's performance drop with smaller training sample discussed in section \ref{sec:performance varying datasize} and consolidates the fact that validation performance and Offline policy evaluation are correlated. We further show how the actors training loss gives a false sense of improvement in appendix \ref{appendix:comapre_train_valid}.

\subsubsection{Validation Performance of Offline RL algorithms}
\label{Compare True Actor/Critic Loss}

\begin{figure}[hbt!]
\centering
\hspace*{-.6in}
%\advance\leftskip-4cm
   \subfloat[]{\includegraphics[width=0.3\linewidth]{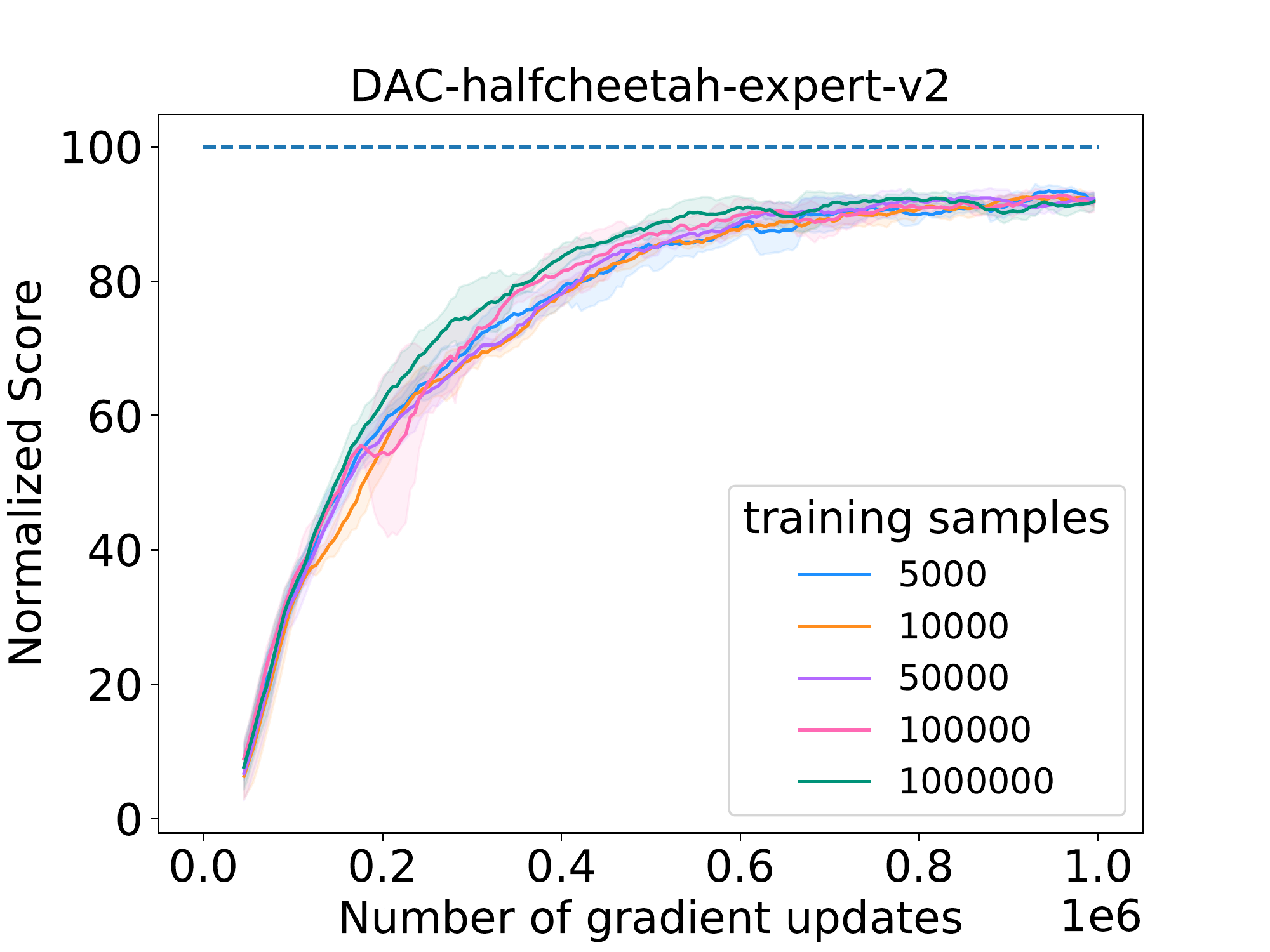}}
  \subfloat[]{\includegraphics[width=0.3\linewidth]{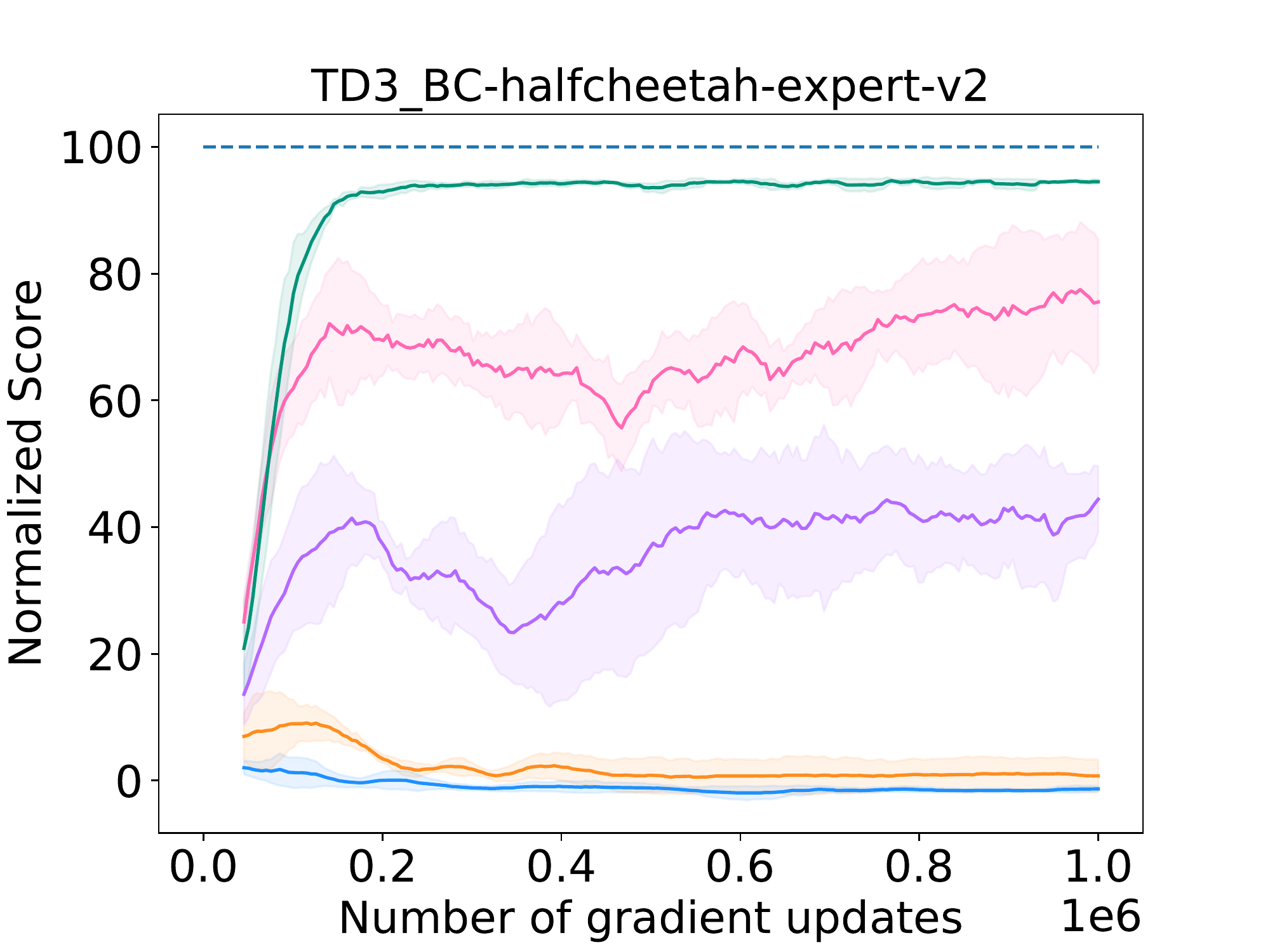}}
  \subfloat[]{\includegraphics[width=0.3\linewidth]{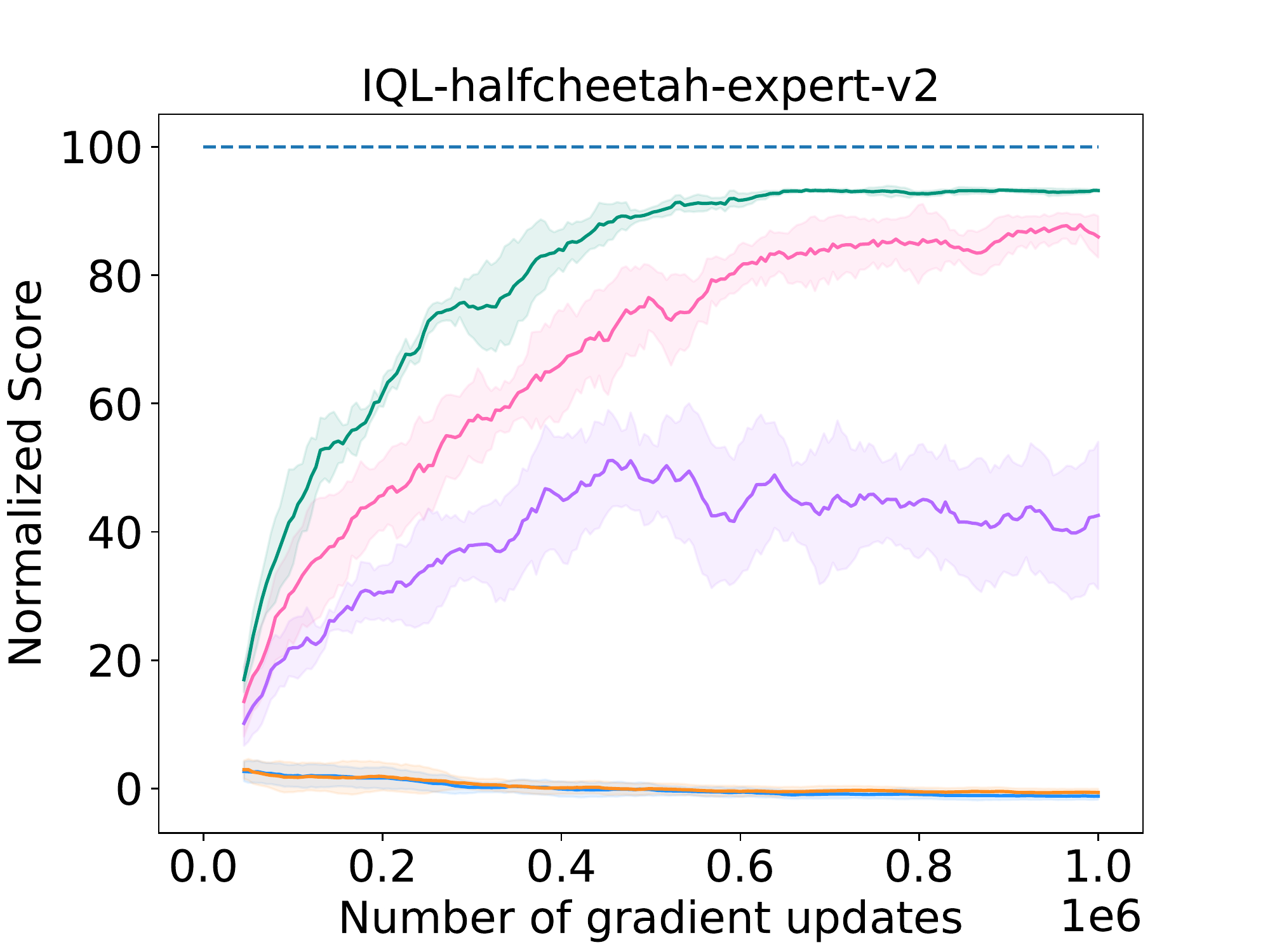}}
  \subfloat[]{\includegraphics[width=0.3\linewidth]{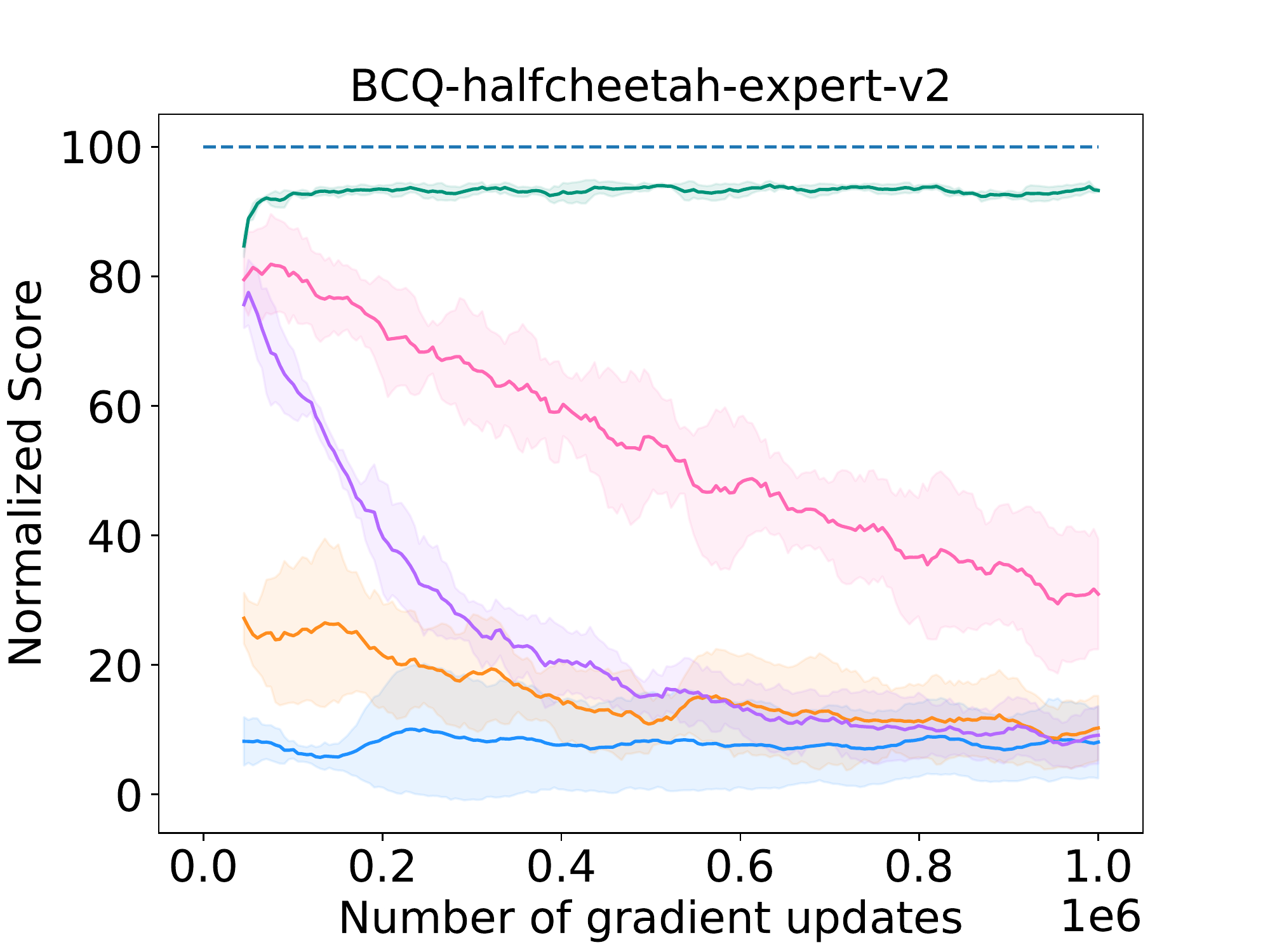}}

  \vspace{-0.45cm}
 \hspace*{-.6in}
   \subfloat[]{\includegraphics[width=0.3\linewidth]{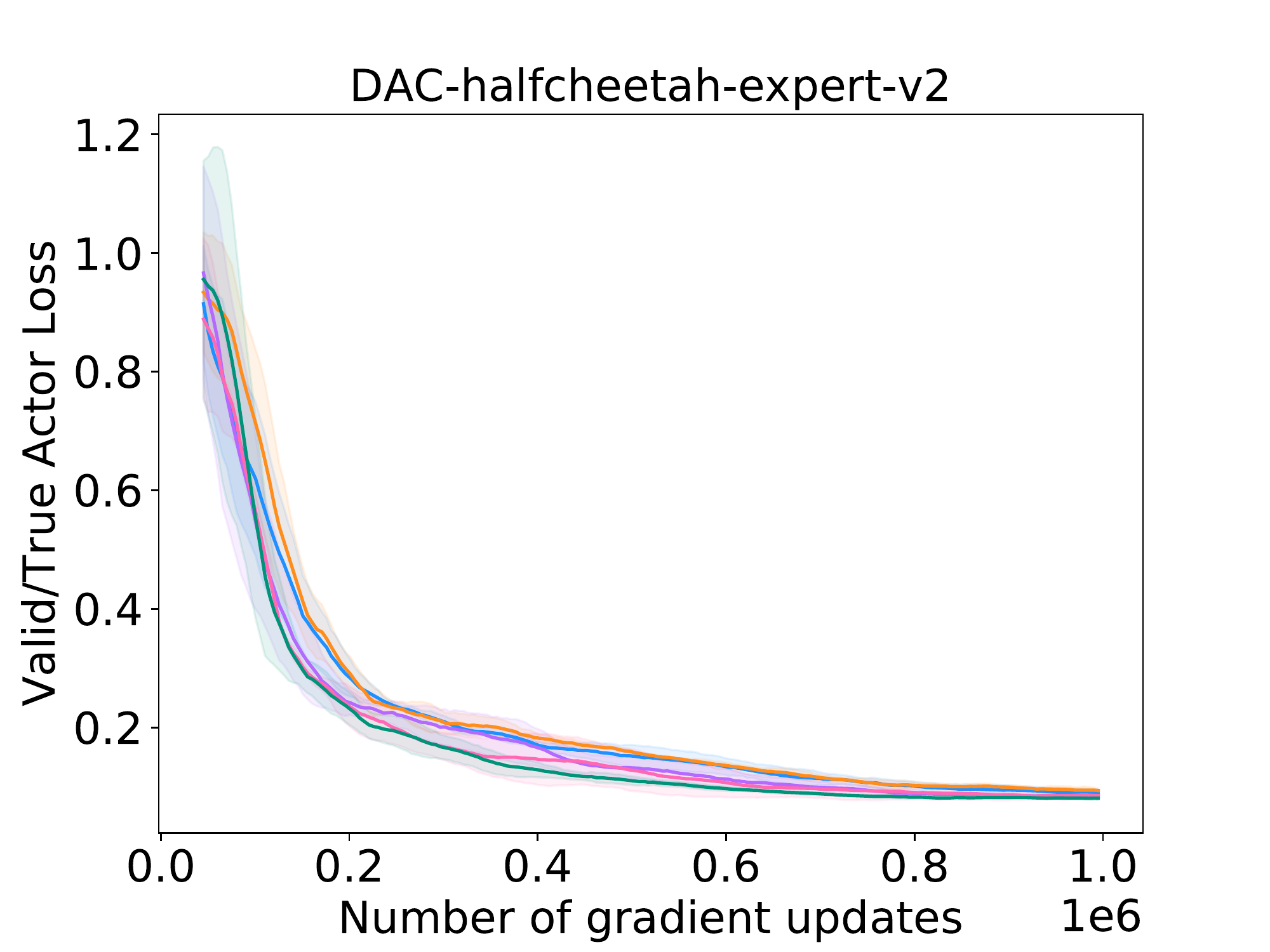}}
  \subfloat[]{\includegraphics[width=0.3\linewidth]{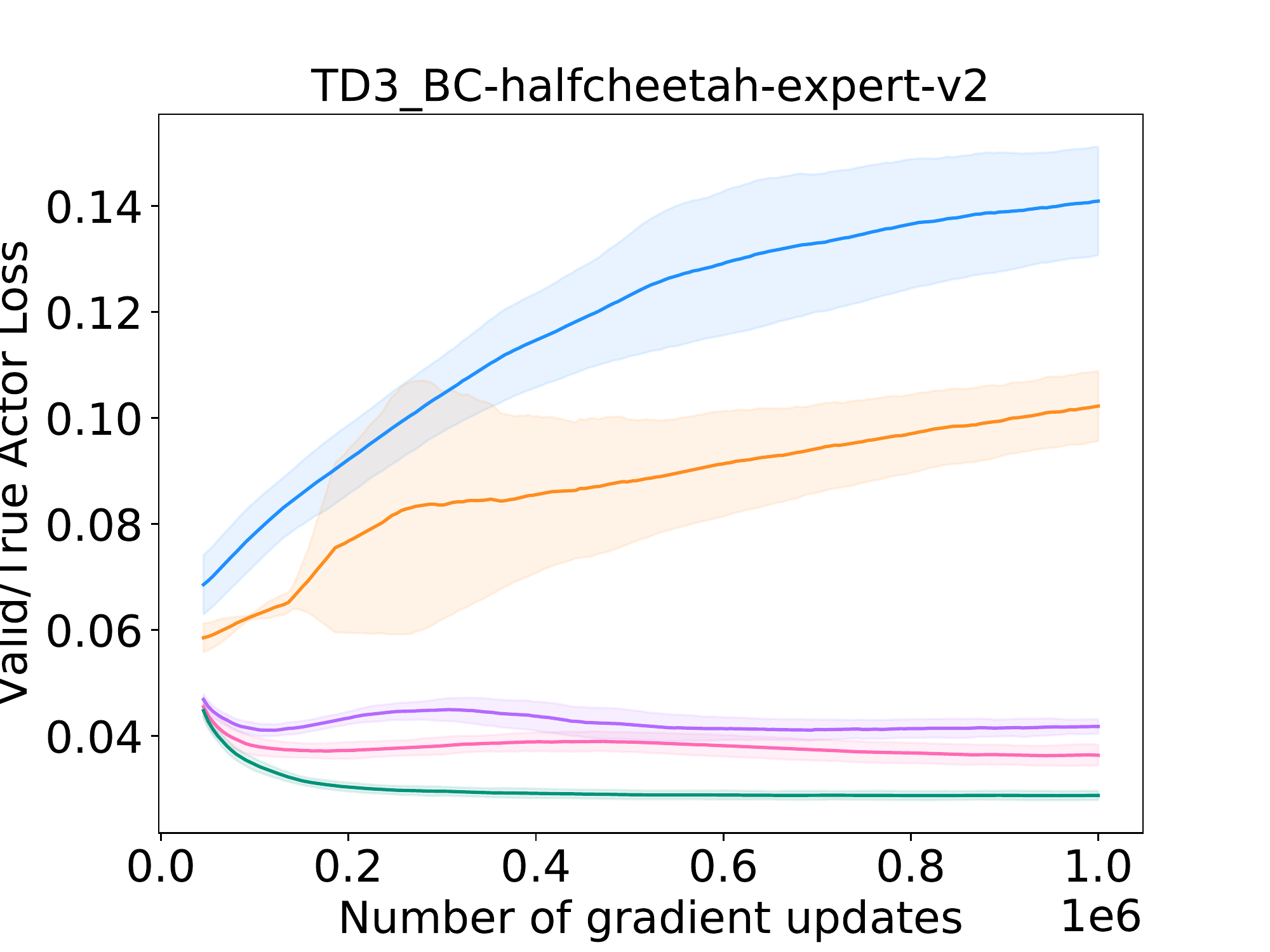}}
   \subfloat[]{\includegraphics[width=0.3\linewidth]{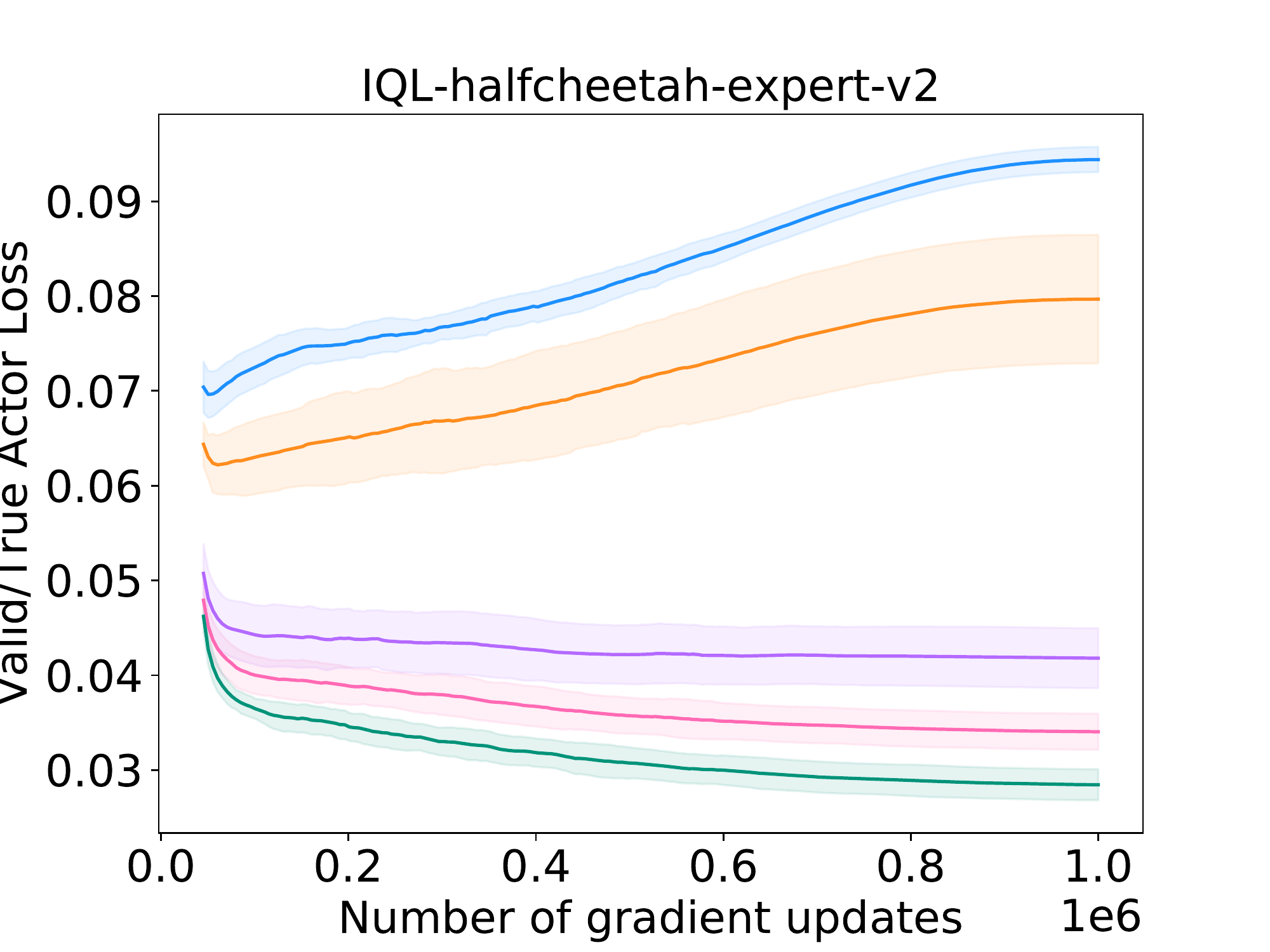}}
   \subfloat[]{\includegraphics[width=0.32\linewidth]{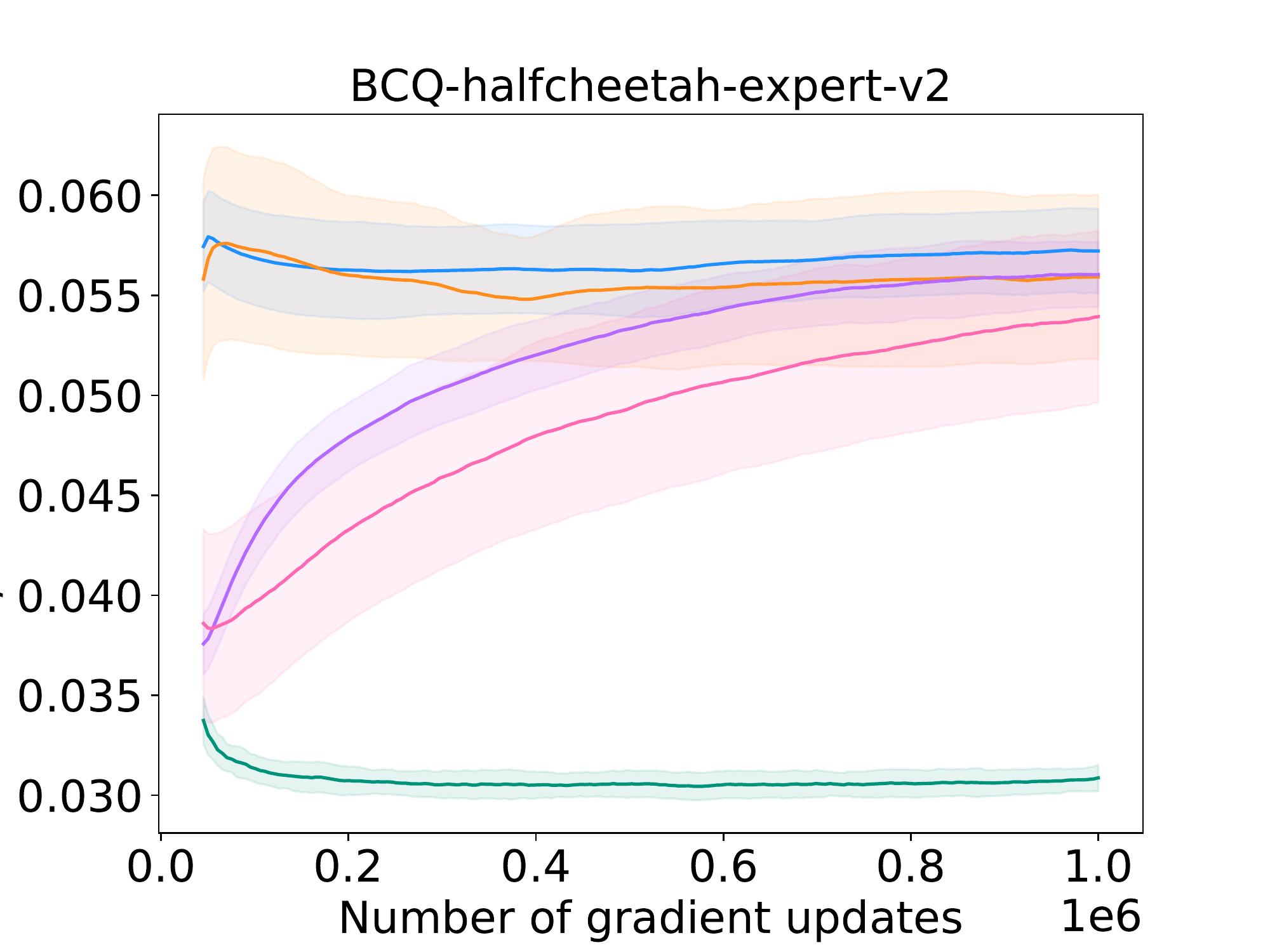}}
%   \!
% \includegraphics[width=0.32\linewidth]{figures/valid_true_critic_loss/BC_halfcheetah-expert-v2_vary_expert_sample.pdf}

\caption{ Performance curve evaluated over 1M gradient updates of (a) DAC, (b) TD3-BC, (c) BCQ, (d) BC and corresponding MSE loss (e-h) between $\pi(s_V)$ and $a_V$ over the validation dataset as we vary number of training dataset.}
\label{noramlized_score_validation_performance}
\end{figure}

This section further confirms our conjecture above - the validation dataset is a useful metric to truly measure the performance improvement for different algorithms. Figure \ref{noramlized_score_validation_performance} further confirms this. We plot the cumulative performance return over 1M training iterations, for each of the sample size of the dataset over the HalfCheetah environment for different algorithms. We find that the validation performance is consistent with the cumulative return metric - for example, in figure \ref{noramlized_score_validation_performance} $(b)$ and \ref{noramlized_score_validation_performance}$(f)$ for the TD3-BC algorithm, the performance improvement is highest when validation loss is the lowest; similarly for sample size of 5000, the validation error for TD3-BC is highest which leads to the lowest performance of this algorithm, as measured by the cumulative returns. 
Without evaluations in between training, we can further guarantee of an improvement using the validation performance. The evaluation on the \emph{validation dataset} provide a clear indication whether training agent is improving or diverging from expected behavior.
% -----------------------------------------
% Offline-RL algorithms are specially useful when it's expensive or risky  (i.e. medical trial, self driving car) to evaluate in the real world. As proved in section the \ref{actor_train_valid_performance}, validation datatset inform about the false sense of improvement, this as well as can be used as further assurance towards algorithm's performance before we evaluate in the real-world. The following figures \ref{noramlized_score_validation_performance} show how algorithms true performance can be visualized with the help of validation performance. 

% "MSE critic loss" indicates, for smaller training samples, it tense to overestimate the action-values and "MSE actor loss" shows ascending curve. On the other hand, actor-loss function of the algorithm for the validation set seems to decline thus clearly misleading the gradient updates.

\subsubsection{Further Discussion on the Validation Performance}
\label{Further Discussion on the Validation Performance}
\begin{figure}[hbt!]
\centering
  \includegraphics[width=0.32\linewidth]{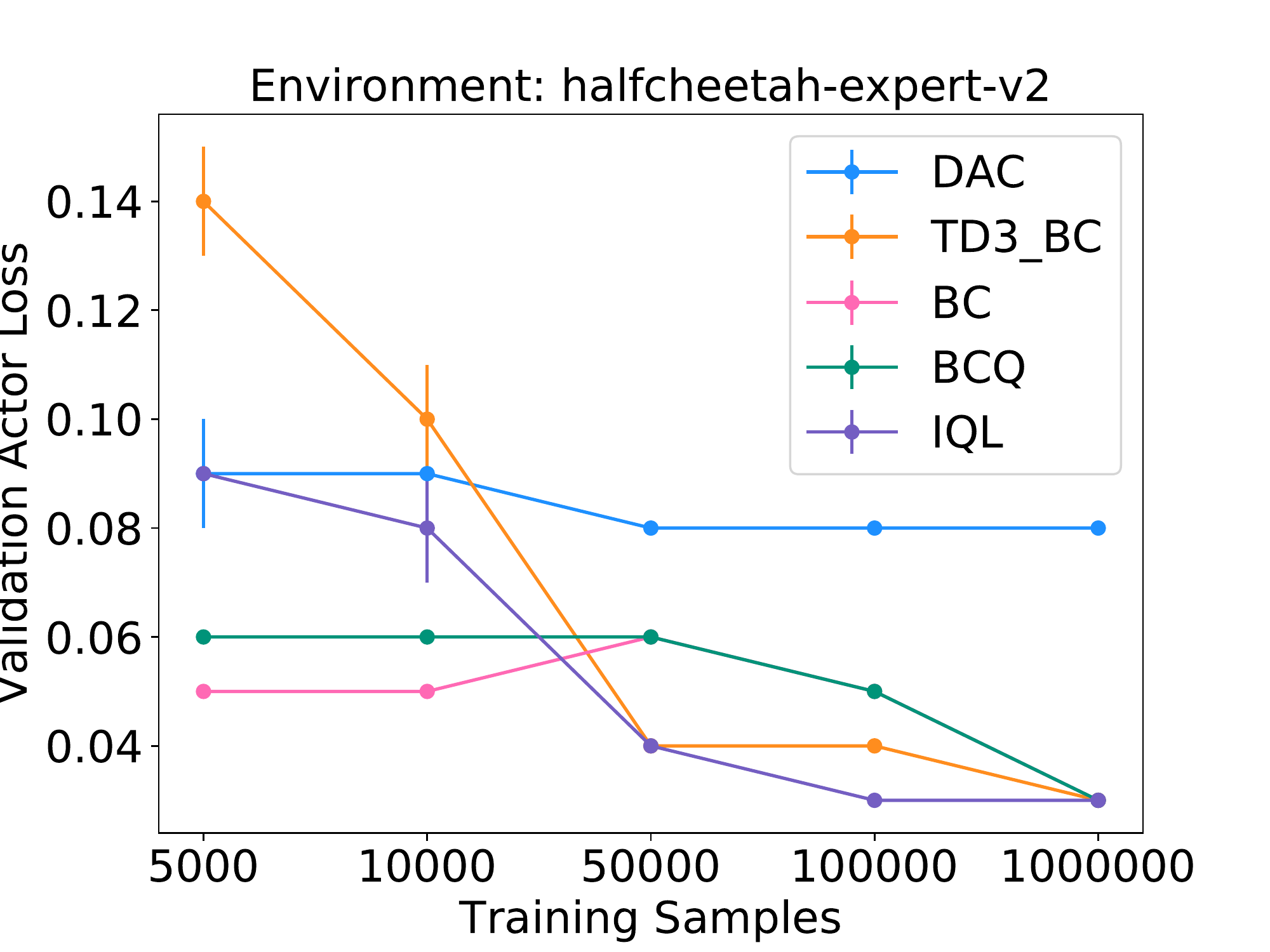}

\caption{Compare Validation loss of different learning algorithms }
\label{bar_graph_validation_true_actor_critic_loss_2}
\end{figure}

In the figure \ref{bar_graph_validation_true_actor_critic_loss_2} we see a clear deviation in actor performance on the validation set as we decrease the training samples size.  But we do not find any significant change in DAC's estimation with expert sample complexity. Despite providing bad estimation compared to offline-RL algorithms, DAC performs better. The offline-RL algorithms are provided with expert samples and are compelled to mimic the expert behavior. And since the expert samples are collected from the same expert, validation estimation are co-related with algorithms performance. We do not have access to optimal expert $\pi^*$, rather collected expert trajectories are sub-optimal, thus DAC still performs better without proving good validation performance. Thus under sub-optimal expert, validation is most useful when we compare algorithms that mimics expert.  

% DAC agent does not gets exposed to expert behavior in training. Rather, DAC uses discriminator to provide reward signals. VERY IMPORTANT: even though the true-actor and true-critic loss does not performs bter than other offline-rl algorithms, we get consistent good performance from DAC, which means we don't exactly have to mimic expert trajectories (given the expert is not optimal). Thus it's not fair to compare DAC's evaluation with the offlineRL. BUT when we have data collected from same expert of offline RL, we can actually use validation to get some idea about actor critics divergence.

%  compare Validation MSE loss for actor and critic 

% \begin{wrapfigure}{r}{0.5\textwidth}
%   \begin{center}
%     \includegraphics[width=0.8\linewidth]{figures/bar_graph/Bar_graph_compare_valid_True_Actor_Loss_halfcheetah-expert-v2.pdf}
%   \end{center}
%   \caption{Compare Validation loss of different learning algorithms }
% \label{bar_graph_validation_true_actor_critic_loss_2}
% \end{wrapfigure}

% \begin{figure}[hbt!]
% \centering
% \subfloat[Action MSE loss]{\includegraphics[width=0.32\linewidth]{figures/bar_graph/Bar_graph_compare_valid_True_Actor_Loss_halfcheetah-expert-v2.pdf}}
% \subfloat[Action-Value MSE loss]{\includegraphics[width=0.32\linewidth]{figures/bar_graph/Bar_graph_compare_valid_True_Critic_Loss_halfcheetah-expert-v2.pdf}}
% \caption{ Compare Validation loss of different learning algorithms }
% \label{bar_graph_validation_true_actor_critic_loss_2}
% \end{figure}
\section{Conclusion}
We investigated the sample complexity of different offline RL algorithms, by varying the size of the training dataset for the same training procedure for each of the algorithms. Our experimental studies leads to a surprising finding : the cumulative return performance as typically shown in standard offline RL algorithms over 1M dataset size, is \textit{not always} a good indicative measure of whether the algorithm is robust under smaller dataset. Our experiment with smaller training dataset shows, the performance of the state of the art offline RL algorithms fall dramatically since the objective function do not consider improving sample complexity.

% comparing overfitting phenomenon for each of the algorithms, as we find that different algorithms exhibit different validation performance as we reduce the sample size. 

The key contribution of our work is therefore to provide an important message for studying offline RL algorithms empirically. We emphasize that studying sample complexity of offline RL algorithms is important, to truly evaluate the performance comparison for each algorithm. We show that current offline-RL algorithms overfit with smaller dataset and the best performing algorithm can perform very poorly under such condition. Thus to make Offline-RL algorithm more reliable in real-world application, where collecting data is non-trivial and no way to quantify the required amount of the data to achieve expert like performance, we need to consider model overfitting into account. We show how training loss can be misleading. Unlike recent studies \cite{REM, workflow_offlineRL} that use online performance to evaluate overfitting, we propose a complete offline evaluation of the policy leveraging a validation dataset to foresee if agent is improving. Improving performance in validation set shows a consistent online performance improvement in all our experiments. Thus in real-world applications  (i.e. self-driving car, drone auto-pilot, medical trails, controlling power system etc.), where a badly trained agent can be extremely risky or costly to evaluate, a validation performance can  provide performance improvement guarantee.

% In real-world application, where a badly trained (i.e. self-driving car, drone auto-pilot, medical trails, controlling power system etc.) can be extremely risky or costly, a validation performance can  provide additional performance guarantee.

% % In future work we want to prevent such overfitting
% \begin{itemize}
%     \item Offline RL overfits with smaller dataset.
%     \item Sample-complexity is important to analyze (how good truely) offline algorithms, makes the offlin RL algorithm practicle when it's expensive to collect expert samples. example: TD3-BC despite give good performance with larger dataset fall hard with smaller samples.
%     \item Validation dataset can be used to identify overfit
%     \item In applications, where evaluation is expensive, we can use validation to visualize algorithms improvement/ decreament.
% \end{itemize}

\bibliographystyle{unsrt}  

%%% Comment out this section when you
\bibliography{references} 

\newpage
\section{Appendix}

\subsection{Performance curve of different algorithm}

\begin{figure}[hbt!]
\centering
   \subfloat[]{\includegraphics[width=0.28\linewidth]{figures/normalized_score/DAC_halfcheetah-expert-v2_vary_expert_sample.pdf}}
   \
    \subfloat[]{\includegraphics[width=0.28\linewidth]{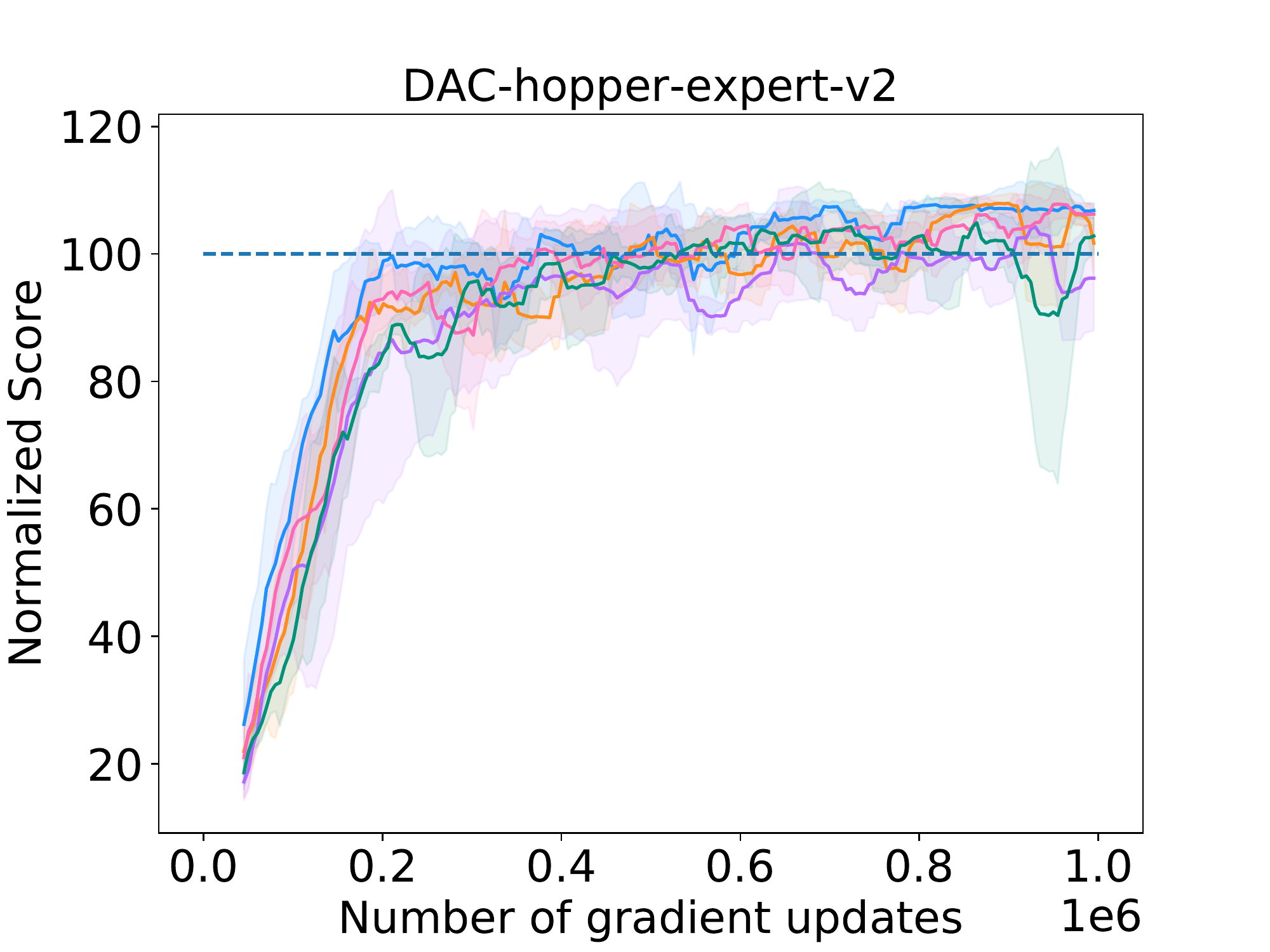}}
   \
    \subfloat[]{\includegraphics[width=0.28\linewidth]{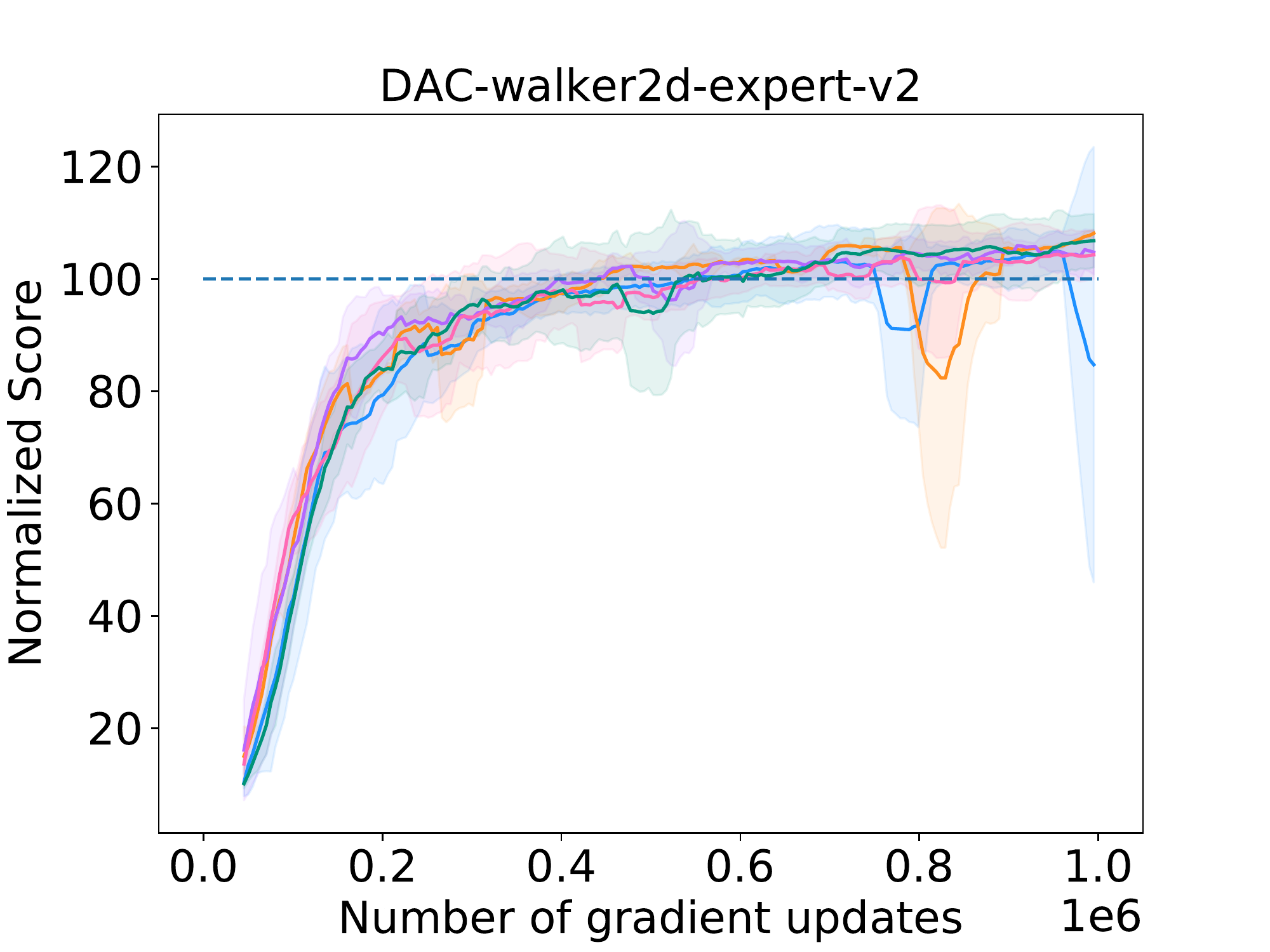}}
   \

\vspace{-0.4cm}
   \subfloat[]{\includegraphics[width=0.28\linewidth]{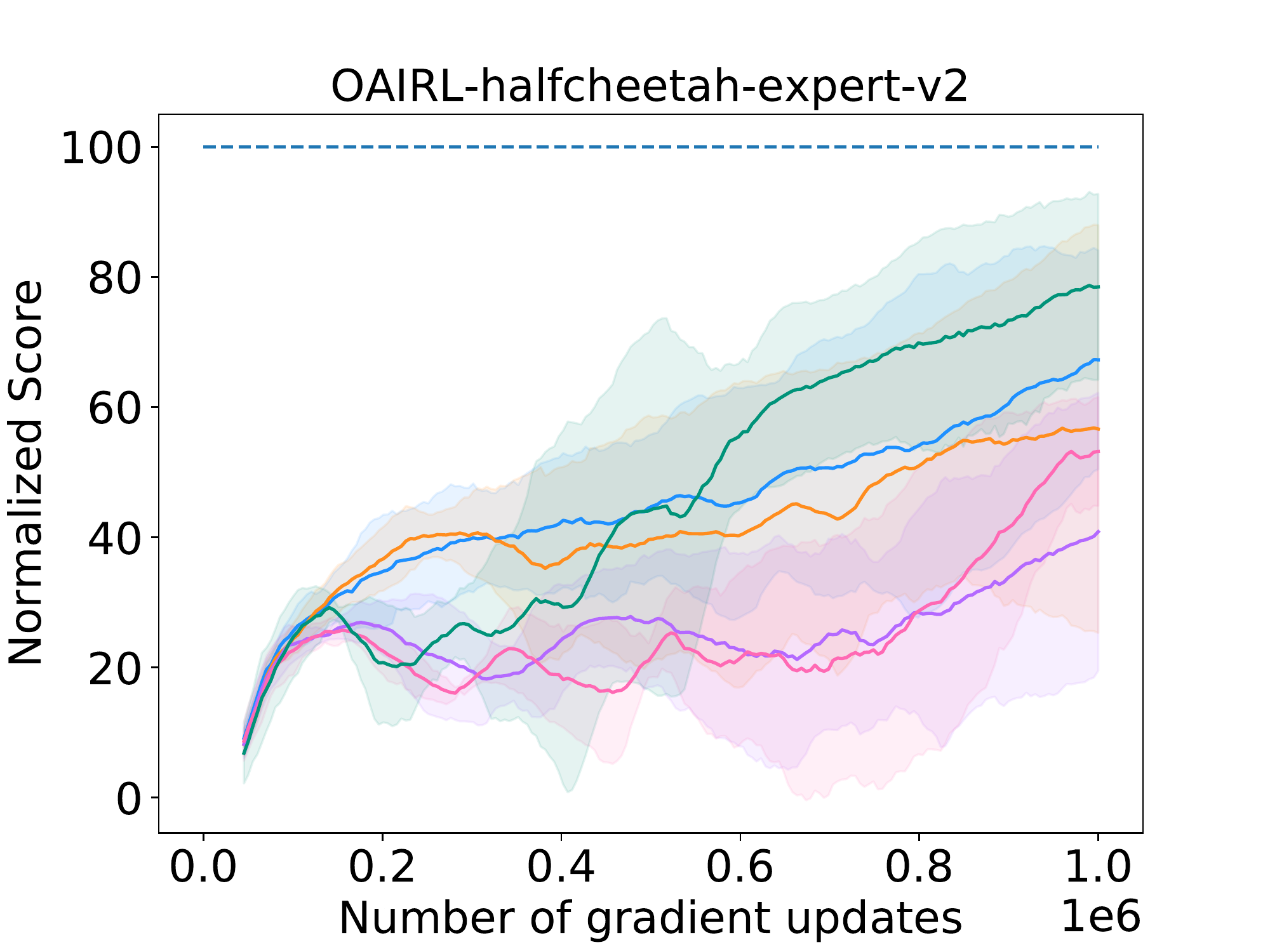}}
   \
    \subfloat[]{\includegraphics[width=0.28\linewidth]{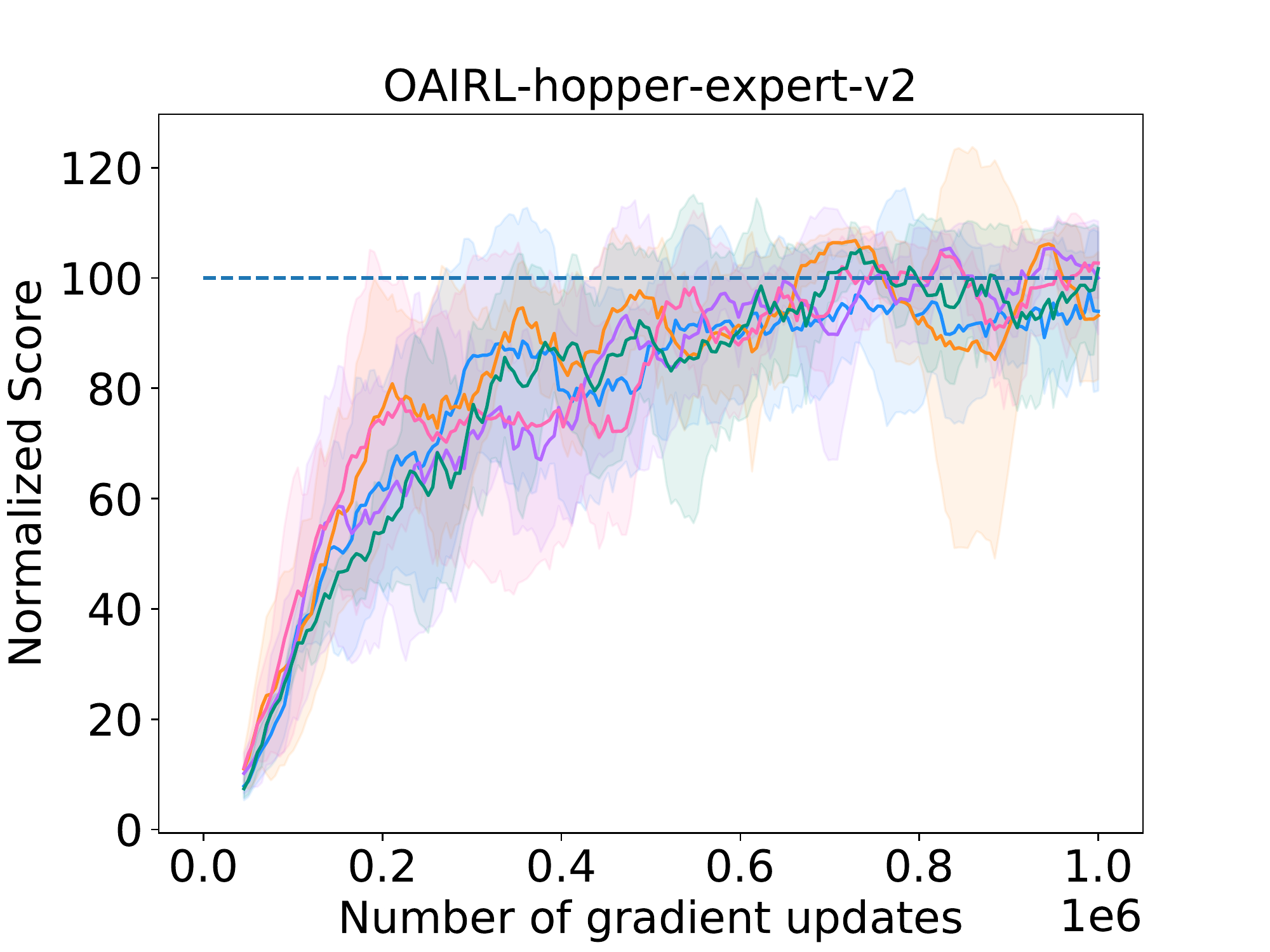}}
   \
    \subfloat[]{\includegraphics[width=0.28\linewidth]{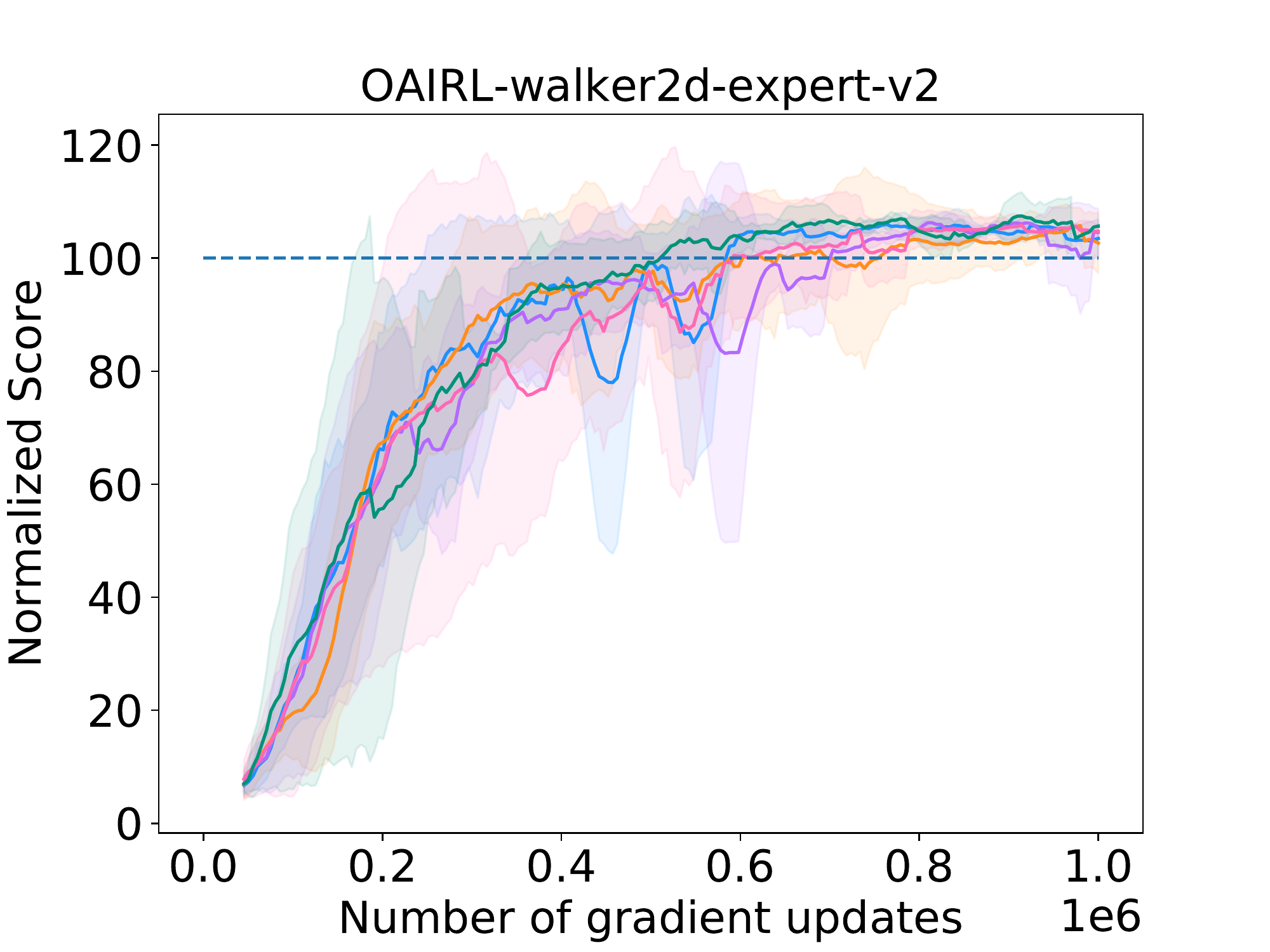}}
   \

   \vspace{-0.4cm}
   \subfloat[]{\includegraphics[width=0.28\linewidth]{figures/normalized_score/IQL_halfcheetah-expert-v2_vary_expert_sample.pdf}}
   \
    \subfloat[]{\includegraphics[width=0.28\linewidth]{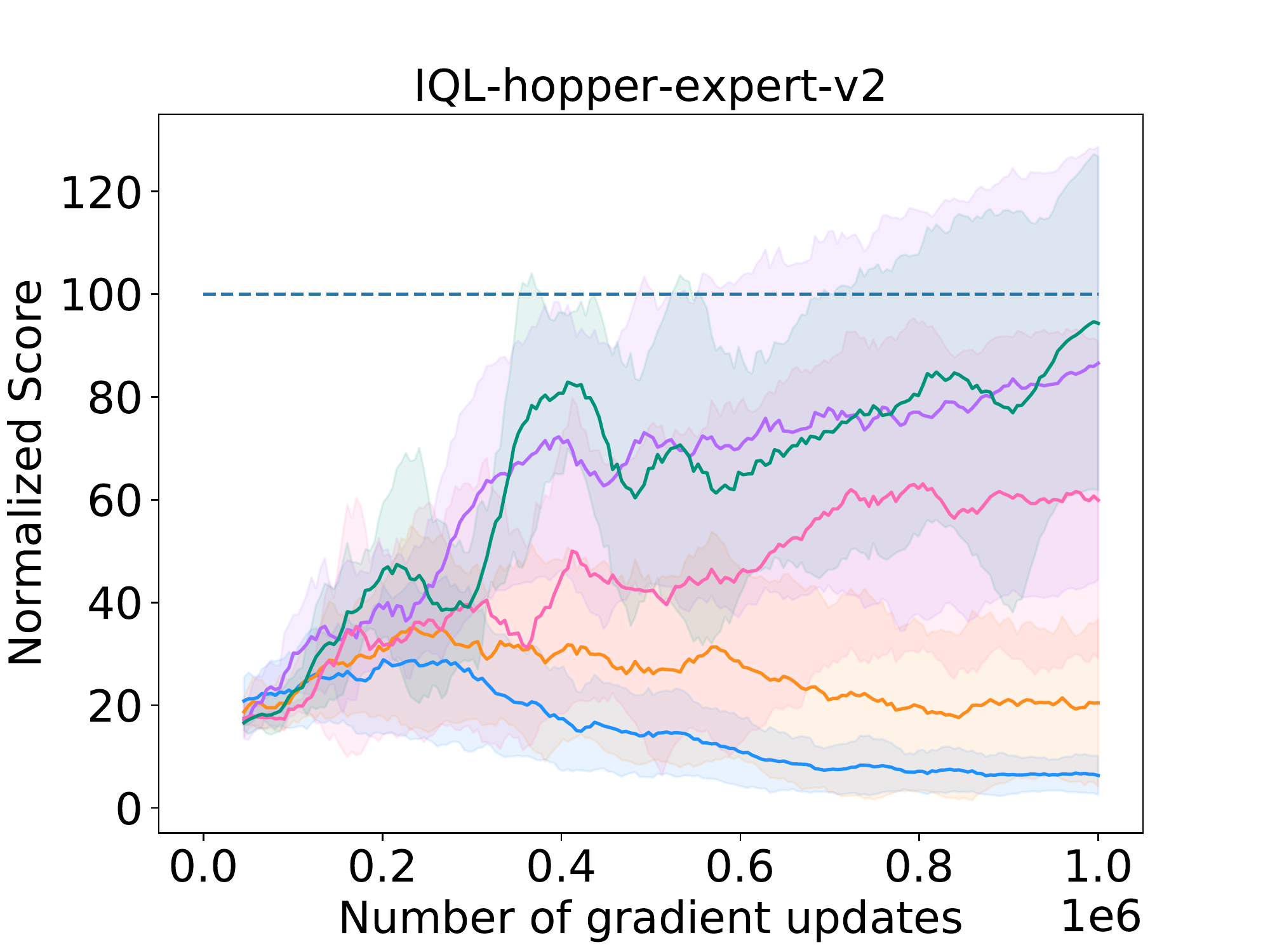}}
   \
    \subfloat[]{\includegraphics[width=0.28\linewidth]{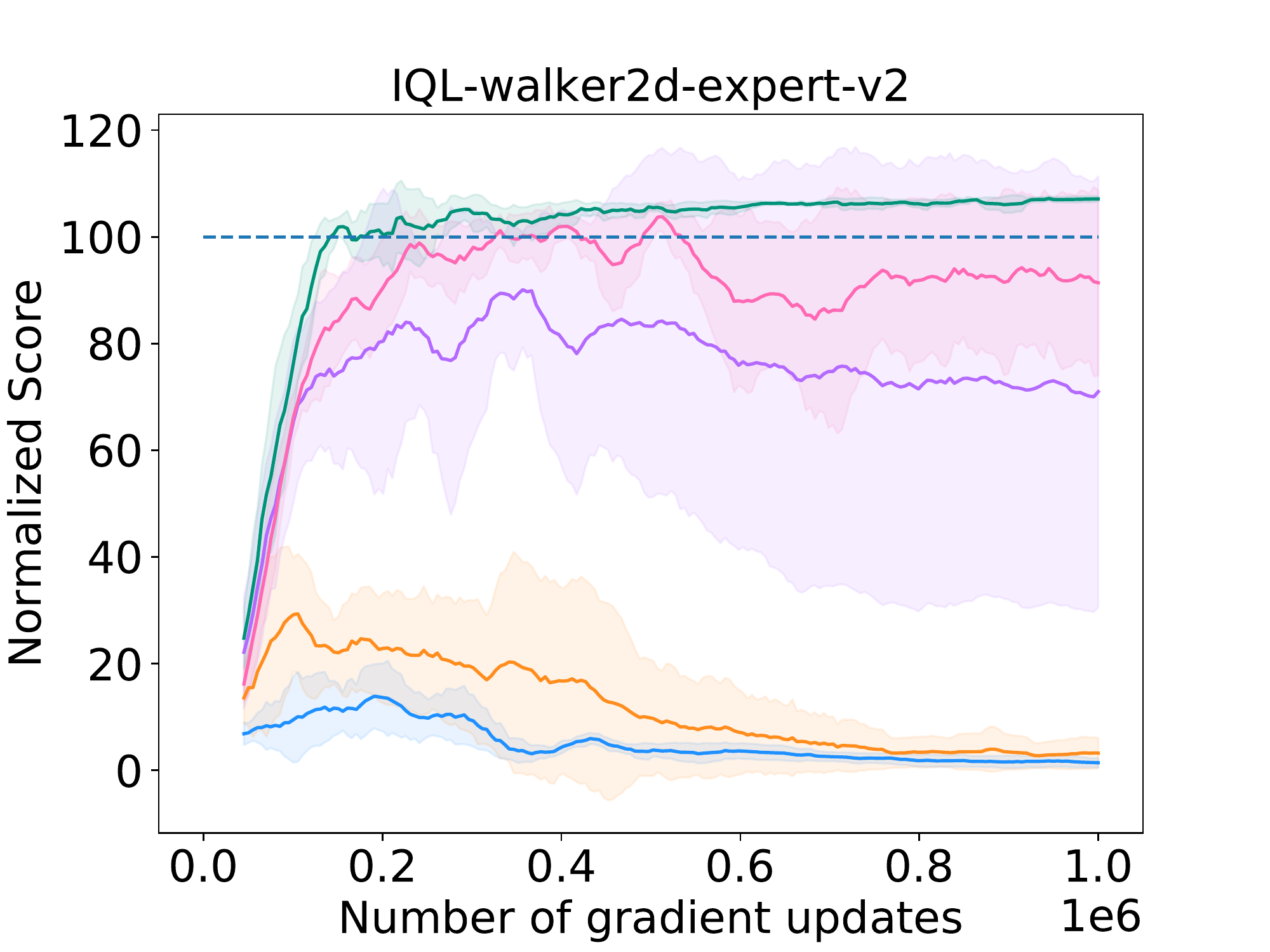}}
   \

\vspace{-0.4cm}
   \subfloat[]{\includegraphics[width=0.28\linewidth]{figures/normalized_score/TD3_BC_halfcheetah-expert-v2_vary_expert_sample.pdf}}
   \
    \subfloat[]{\includegraphics[width=0.28\linewidth]{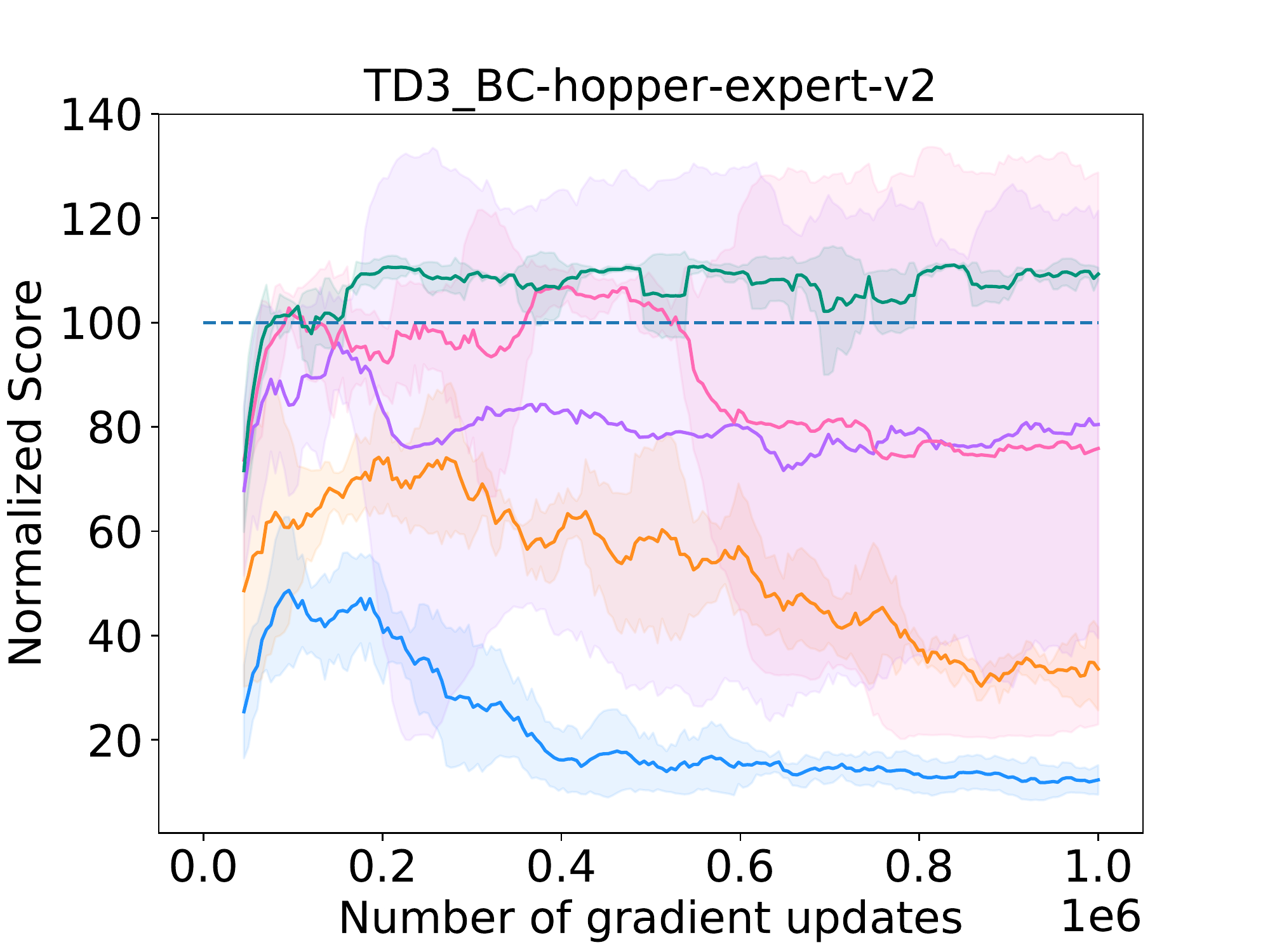}}
   \
     \subfloat[]{\includegraphics[width=0.28\linewidth]{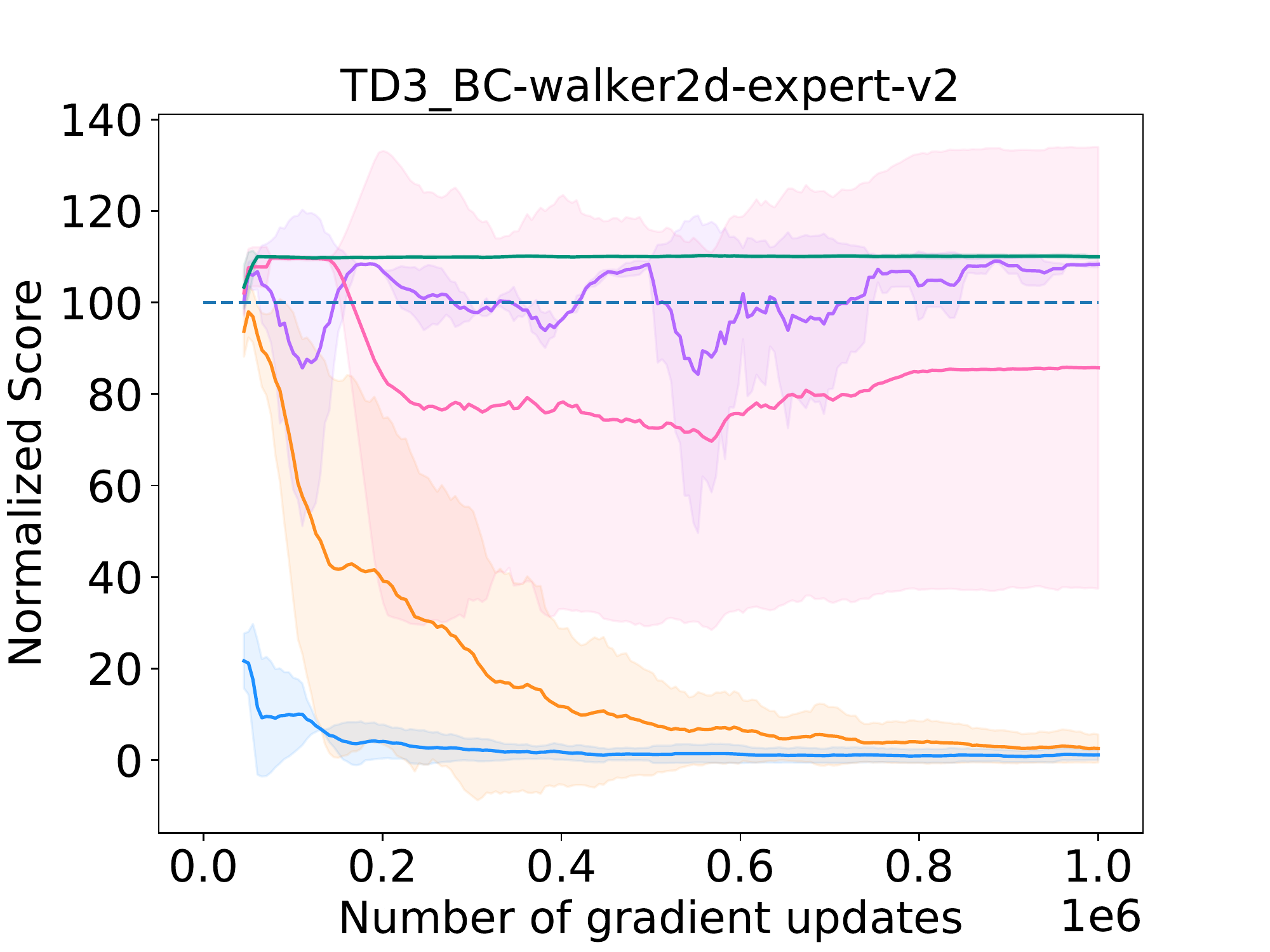}}
   \

\vspace{-0.4cm}
   \subfloat[]{\includegraphics[width=0.28\linewidth]{figures/normalized_score/BCQ_halfcheetah-expert-v2_vary_expert_sample.pdf}}
   \
    \subfloat[]{\includegraphics[width=0.28\linewidth]{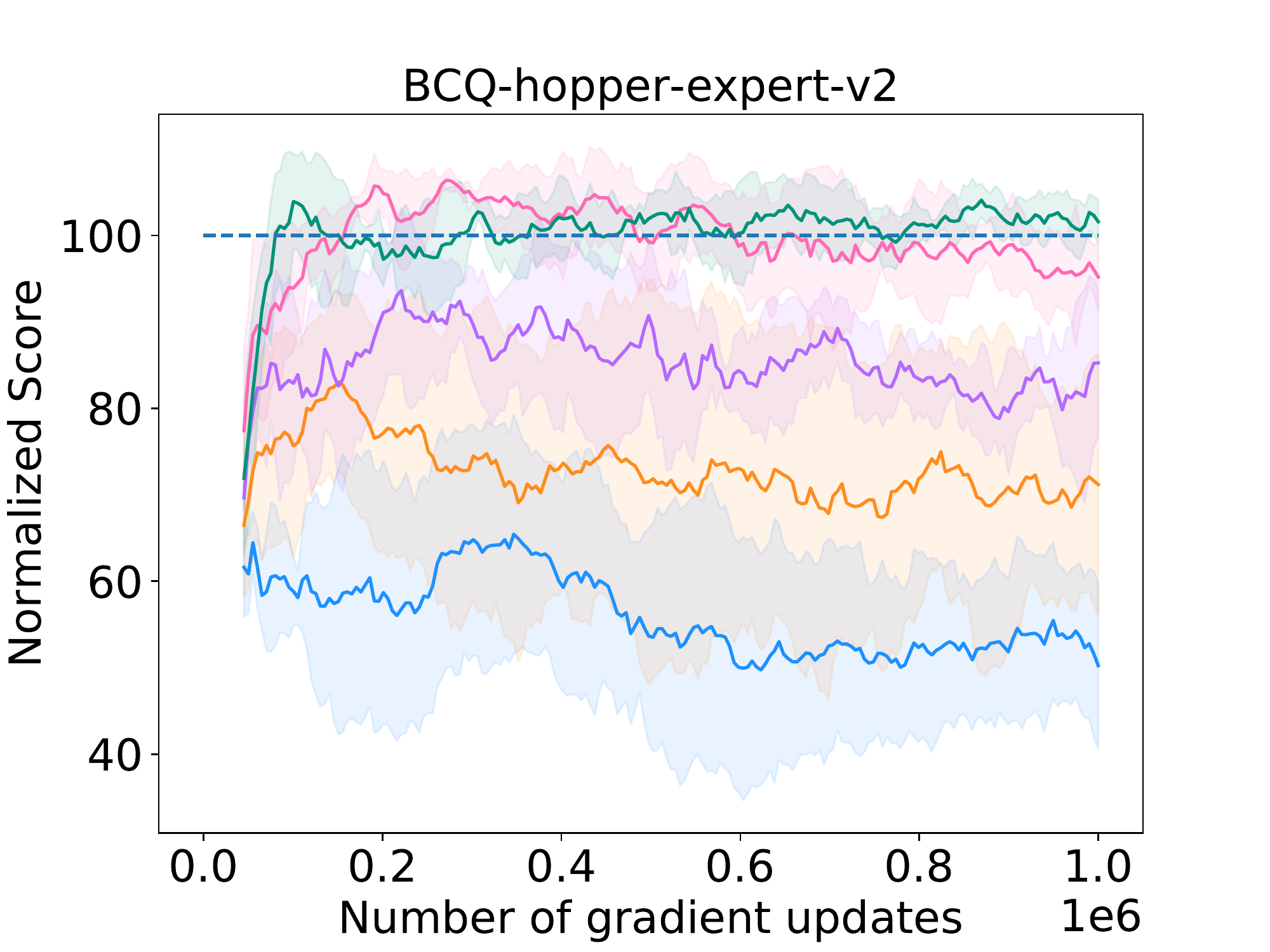}}
   \
    \subfloat[]{\includegraphics[width=0.28\linewidth]{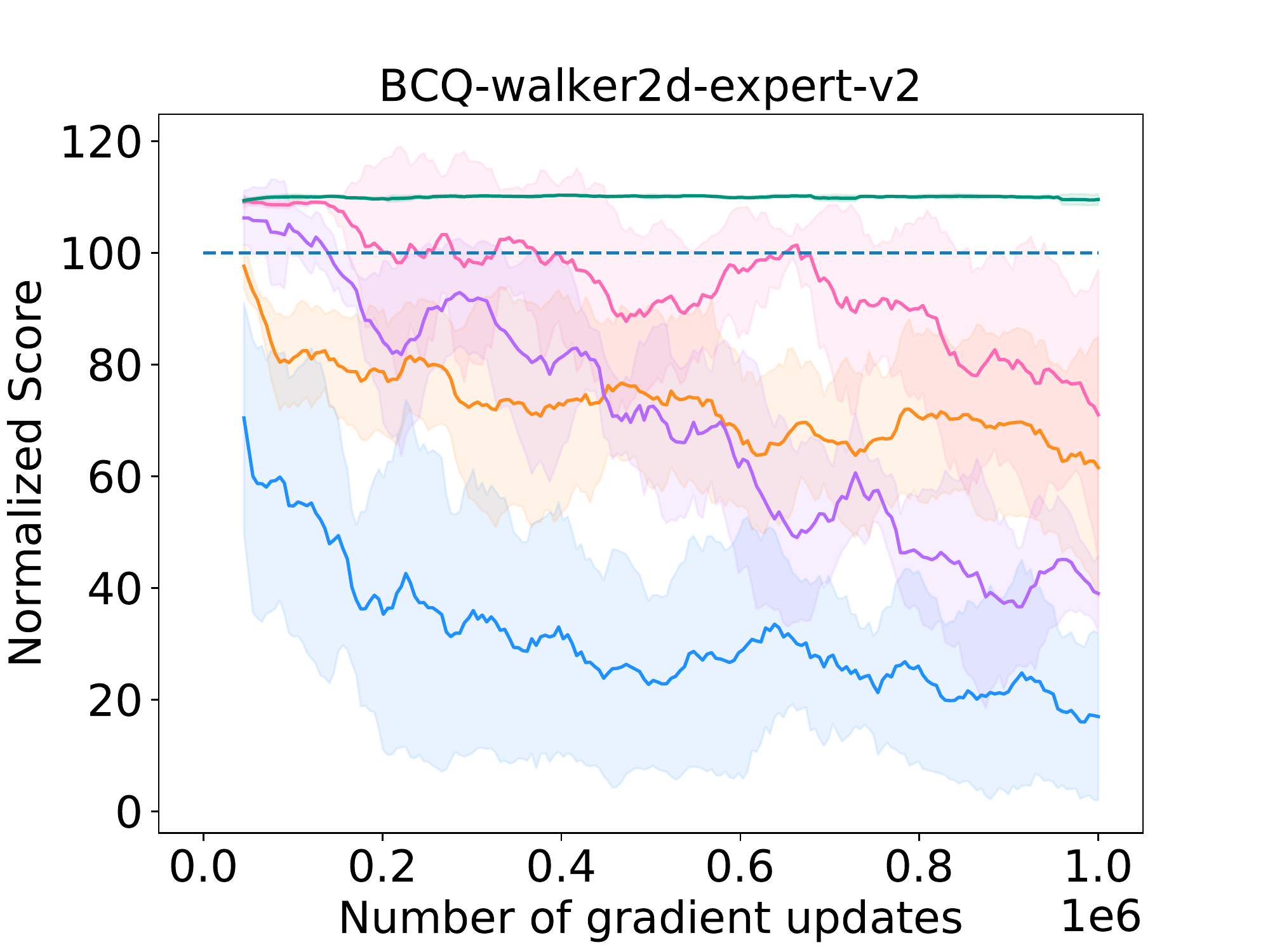}}
   \
  
   \vspace{-0.4cm}
   \subfloat[]{\includegraphics[width=0.28\linewidth]{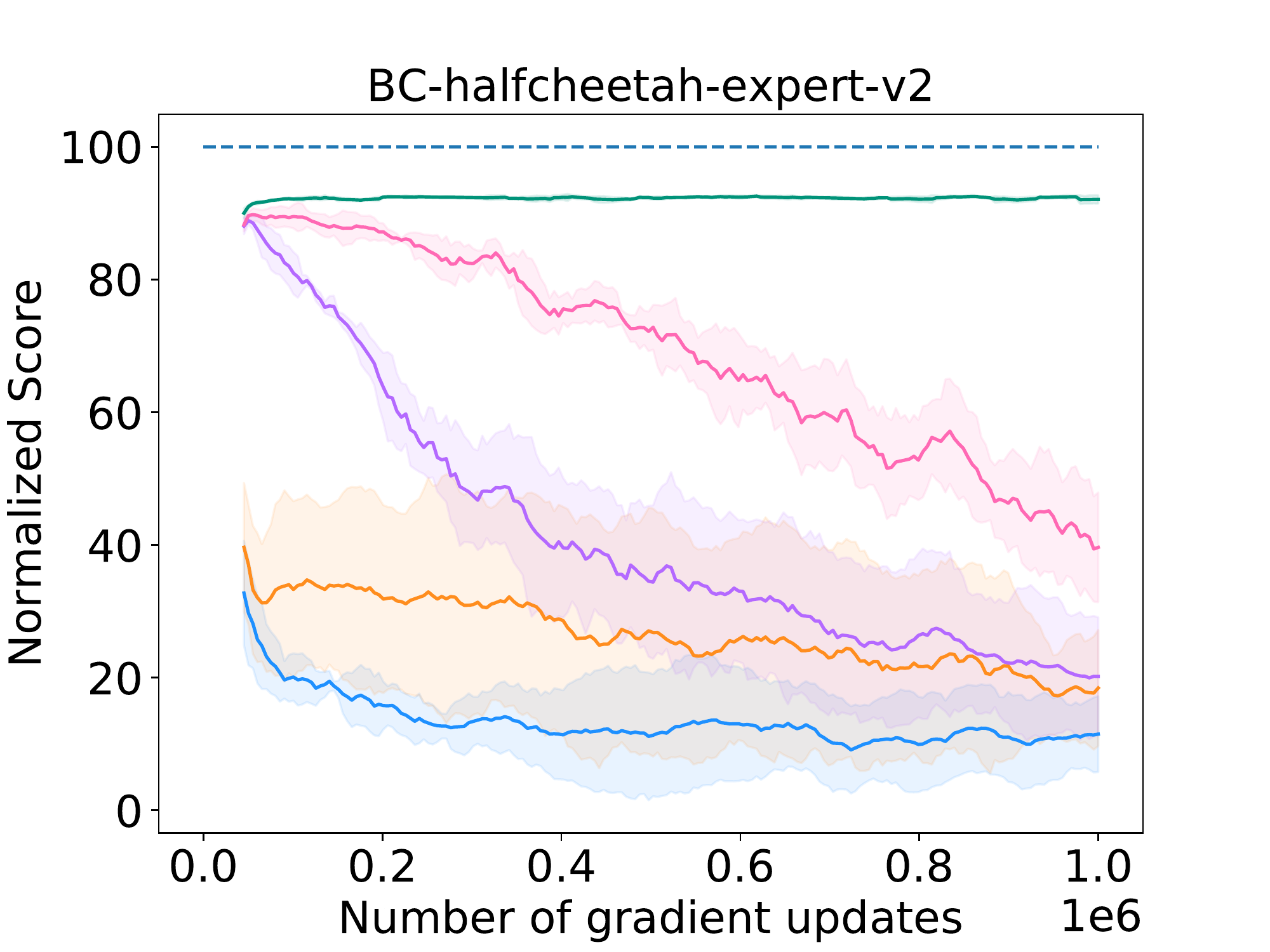}}
   \
    \subfloat[]{\includegraphics[width=0.28\linewidth]{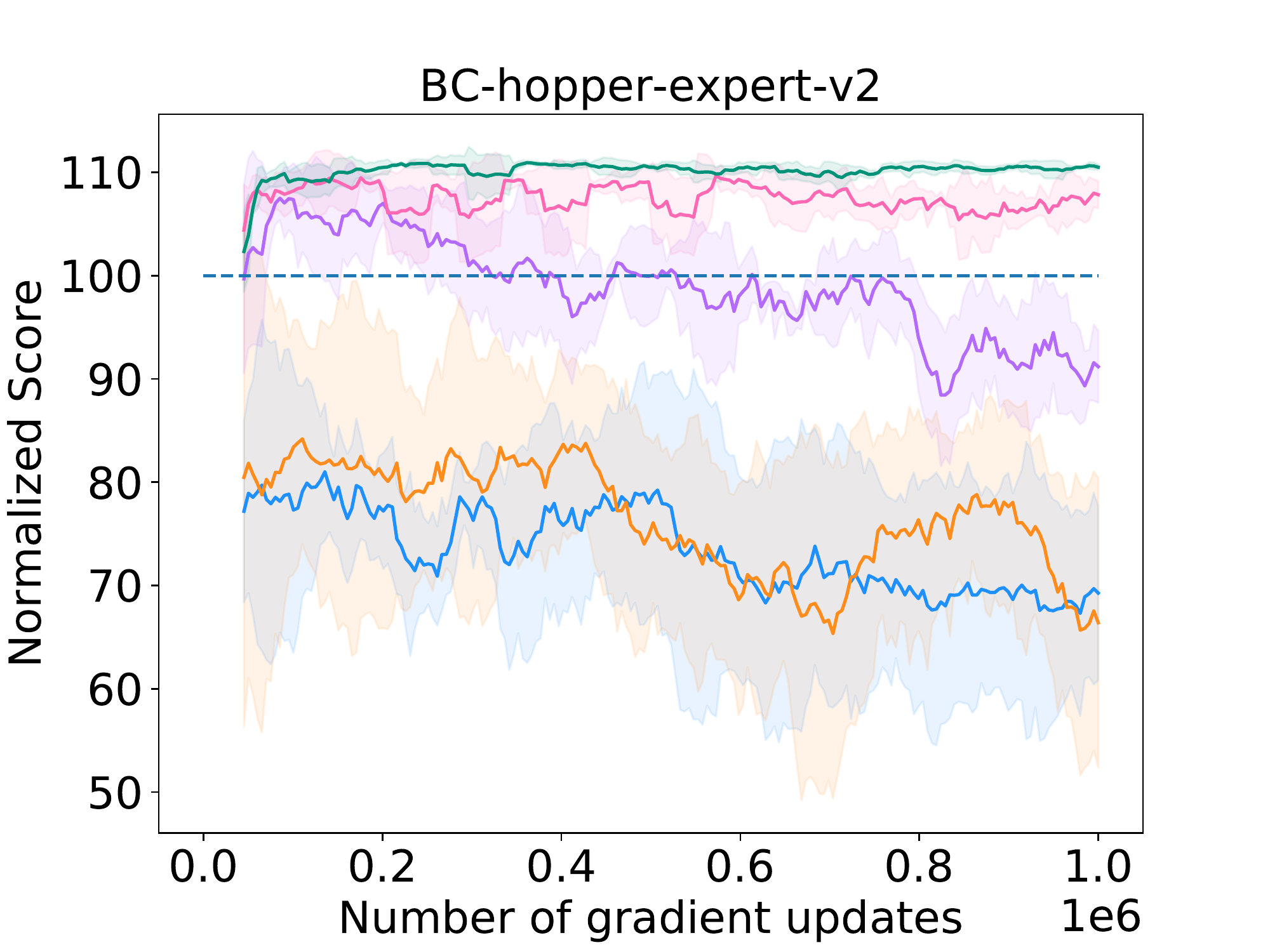}}
   \
    \subfloat[]{\includegraphics[width=0.28\linewidth]{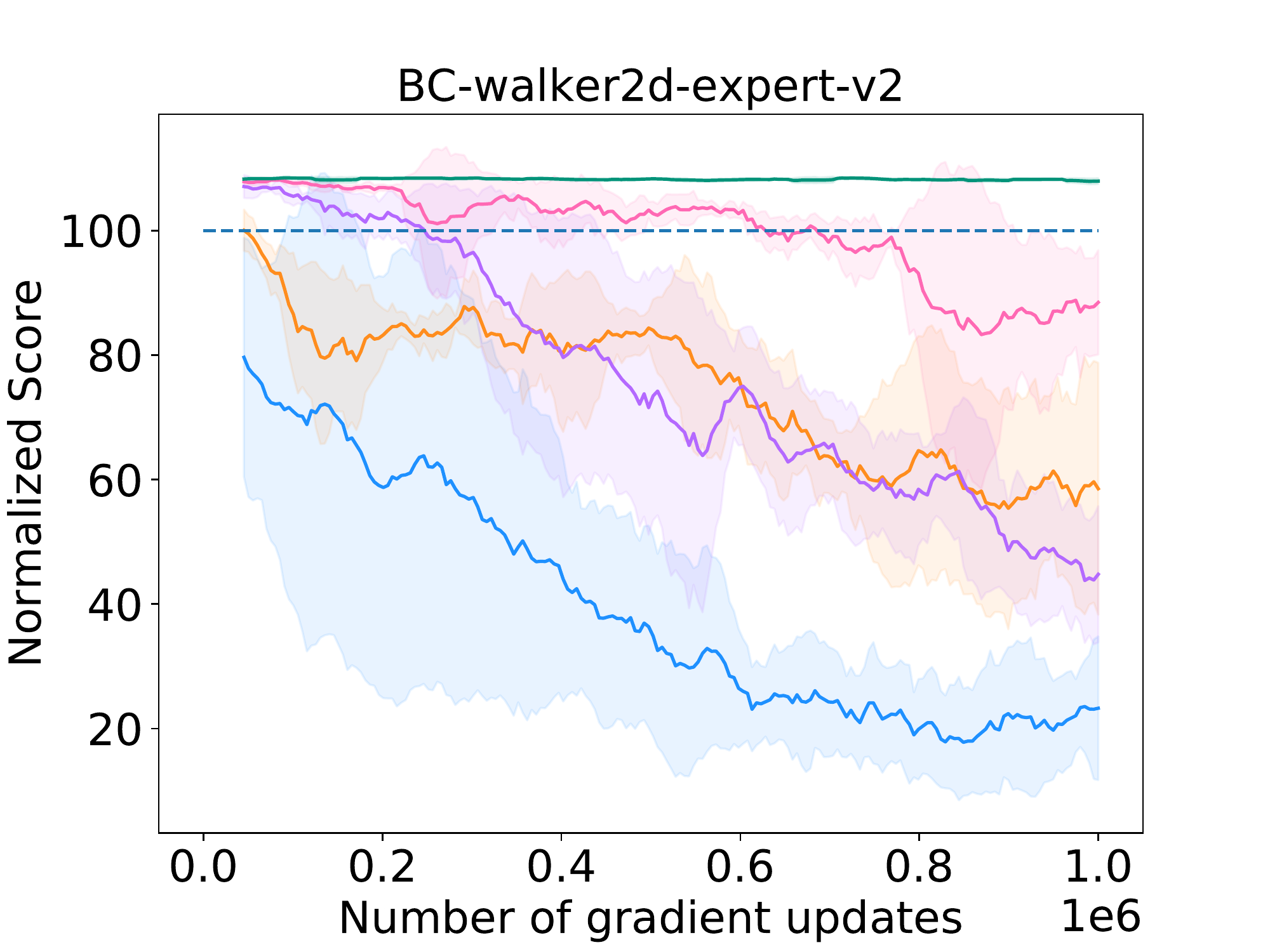}}
   \

\caption{Performance Comparison (D4RL Normalized Score) of DAC with offline-RL Varying Expert data.}
\label{DAC_vs_offline}
\end{figure}

In figure \ref{DAC_vs_offline} we plot the mean performance of the algorithms for seeds 0-4 with $100\%$ confidence interval over 1 million gradient updates. We compare the performance of each algorithm varying training sample size on MuJoCo control tasks.

% ---------------------------------------------
\subsection{Compare Training and Validation Actor Evaluation}
\label{appendix:comapre_train_valid}
% The class of Actor-Critic algorithm uses Critic's evaluation of to improve Actor. To portray what is going wrong in these algorithms, 
We compare the actor's training loss and actor's validation loss (MSE($\pi_\theta(s_V),a_V$)) over 1 million gradient updates. It clearly shows how actors training loss (blue) gives a false sense of imporvement.

%------------ train validation --------
\begin{figure}[hbt!]
\centering
 \hspace*{-.6in}
   \subfloat[]{\includegraphics[width=0.26\linewidth]{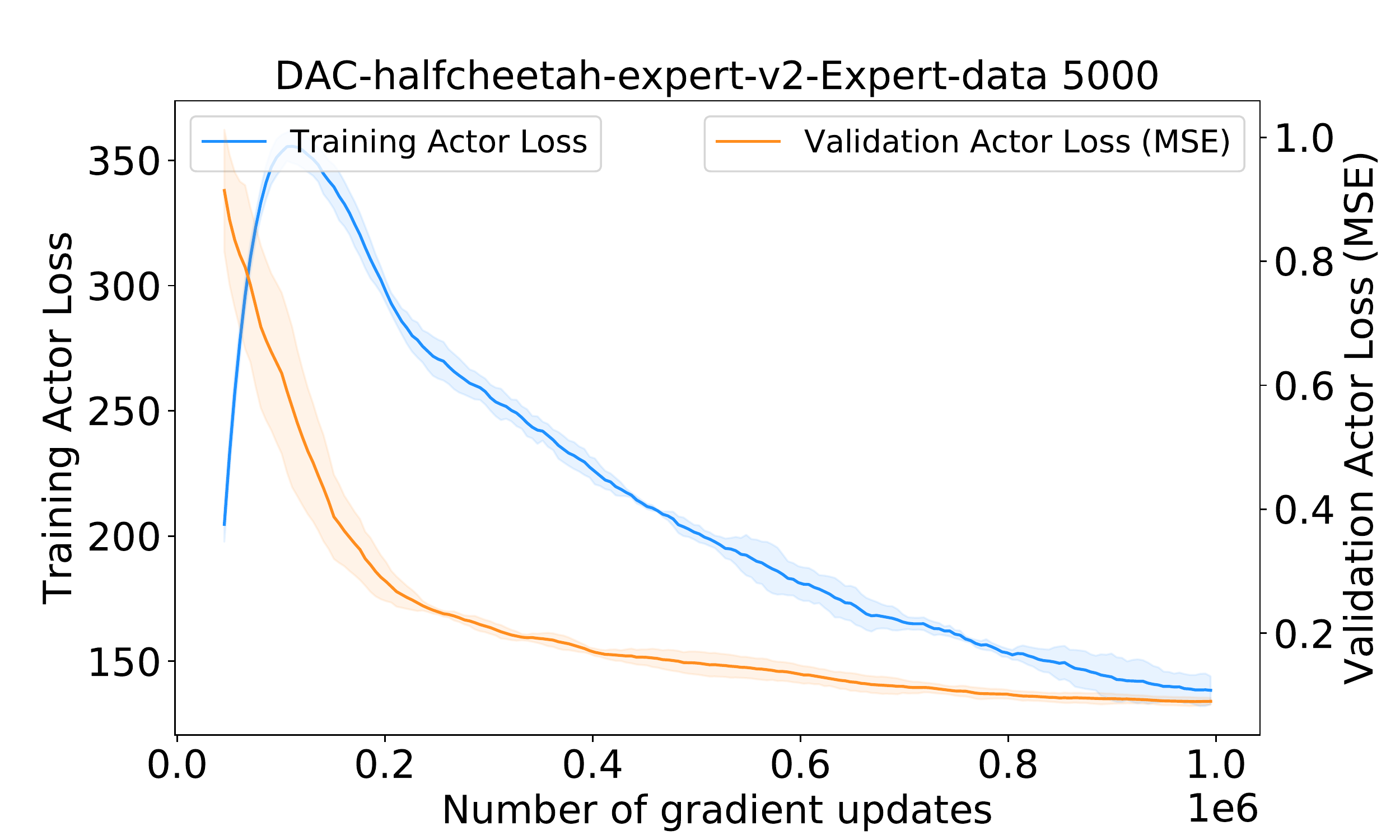}}
   \vspace{0.2cm}
    \subfloat[]{\includegraphics[width=0.26\linewidth]{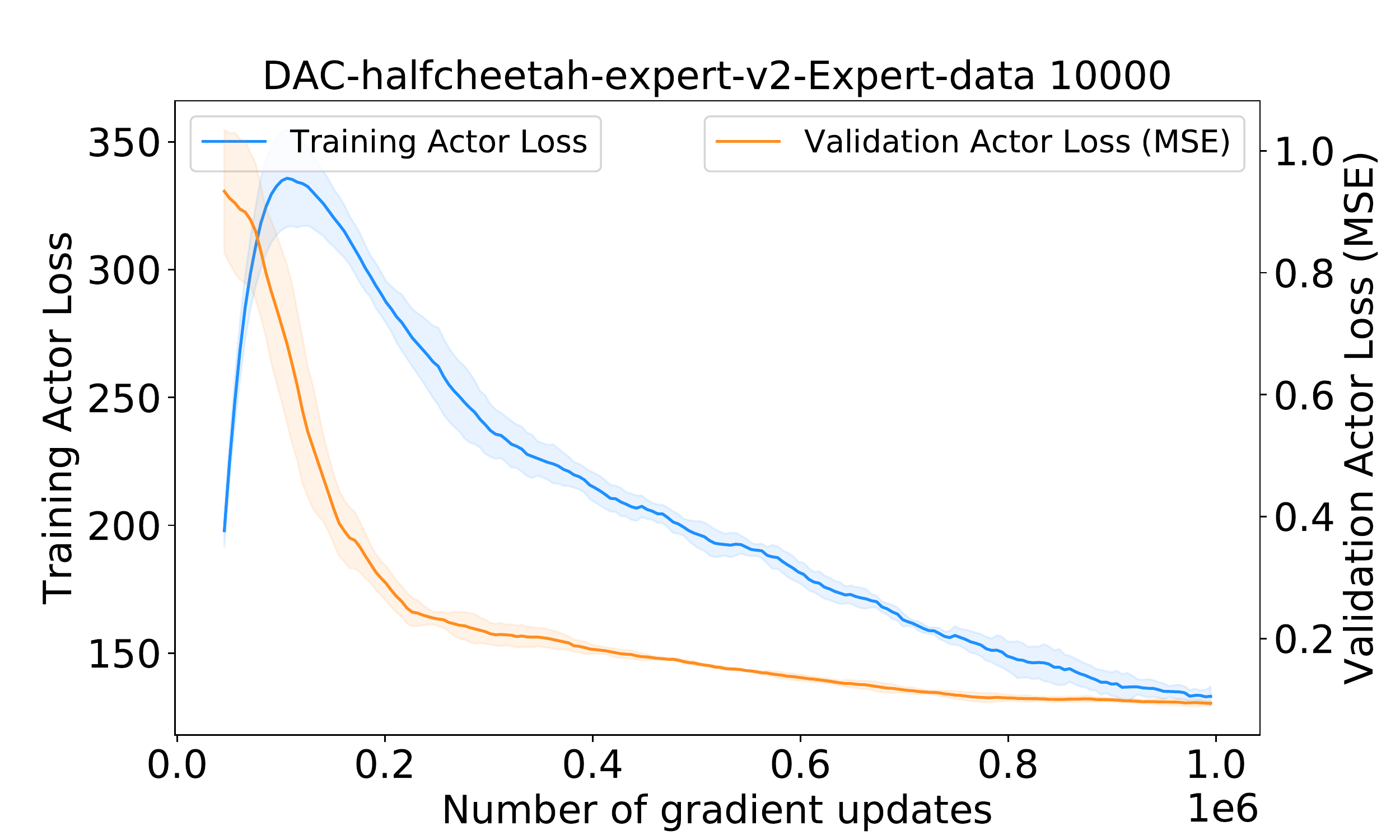}}
    \subfloat[]{\includegraphics[width=0.26\linewidth]{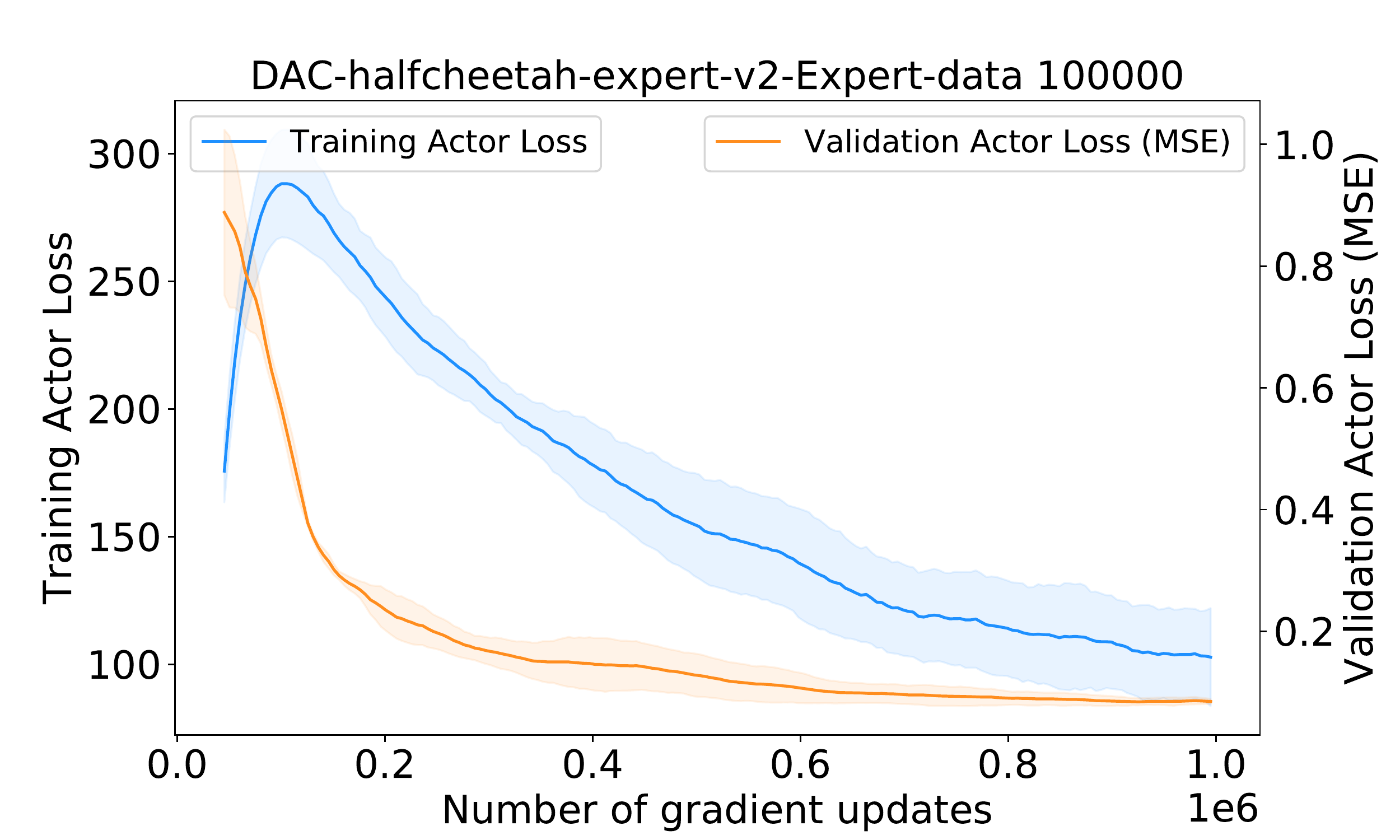}}
    \subfloat[]{\includegraphics[width=0.26\linewidth]{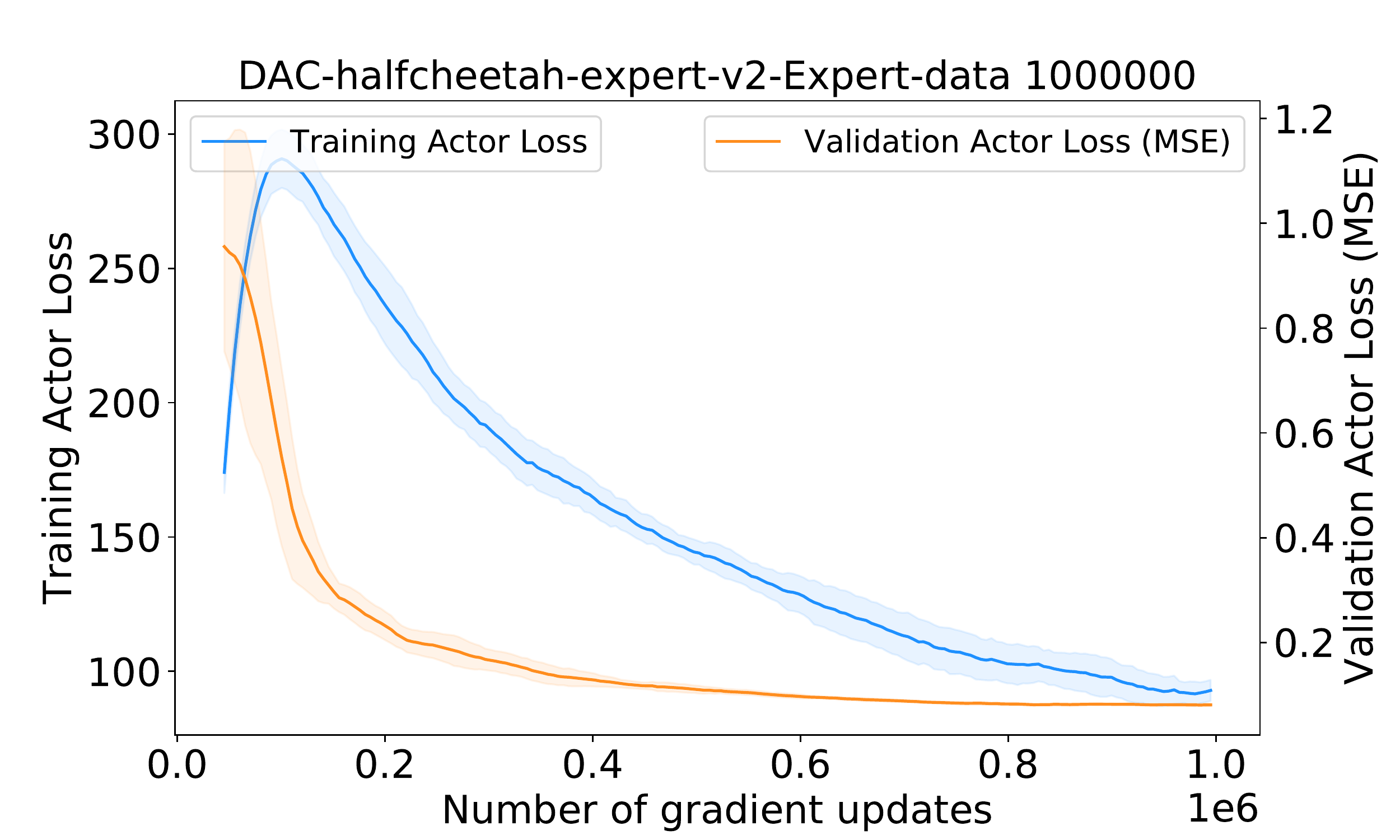}}
   
   \vspace{-0.45cm} 
   \hspace*{-0.6in}
      \subfloat[]{\includegraphics[width=0.26\linewidth]{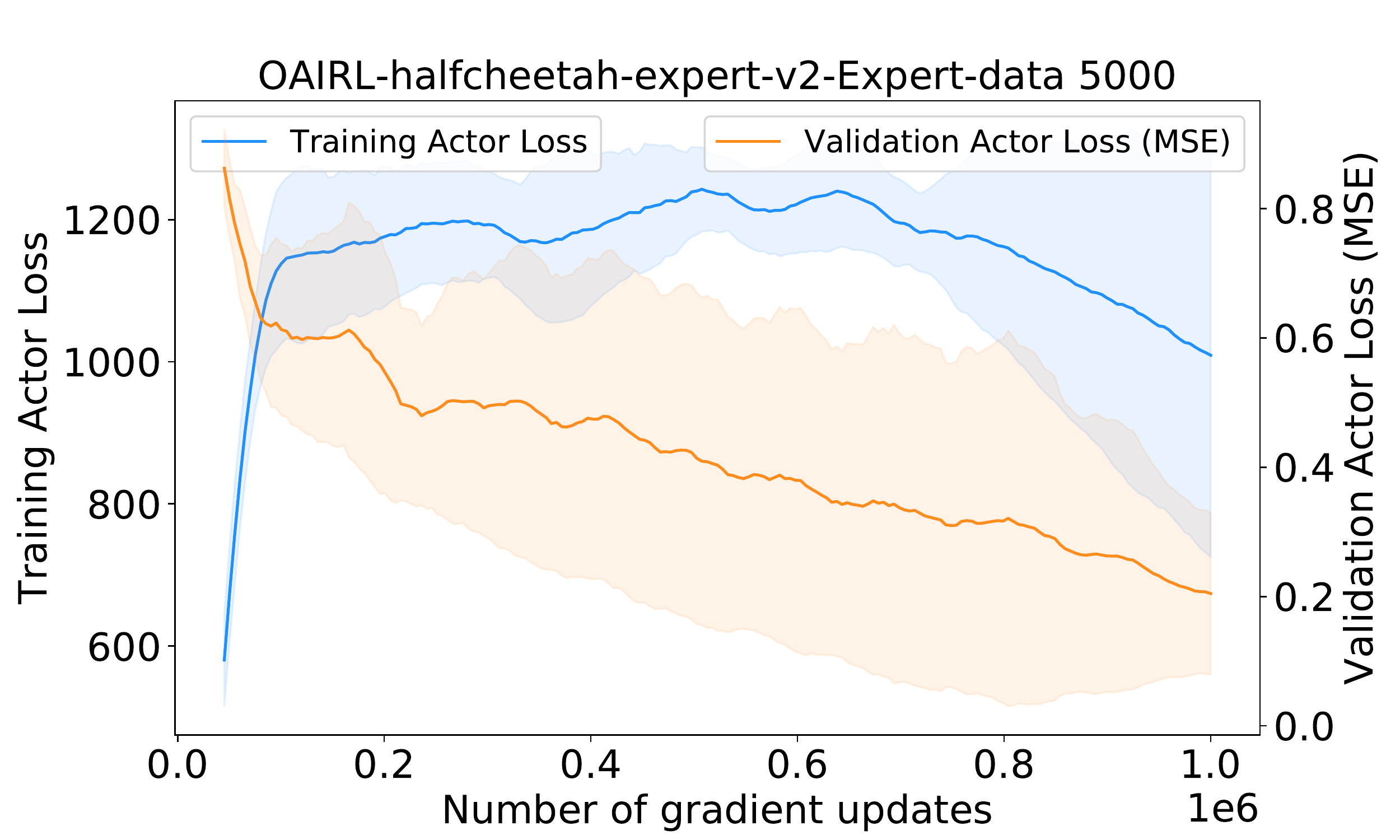}}
   \!
    \subfloat[]{\includegraphics[width=0.26\linewidth]{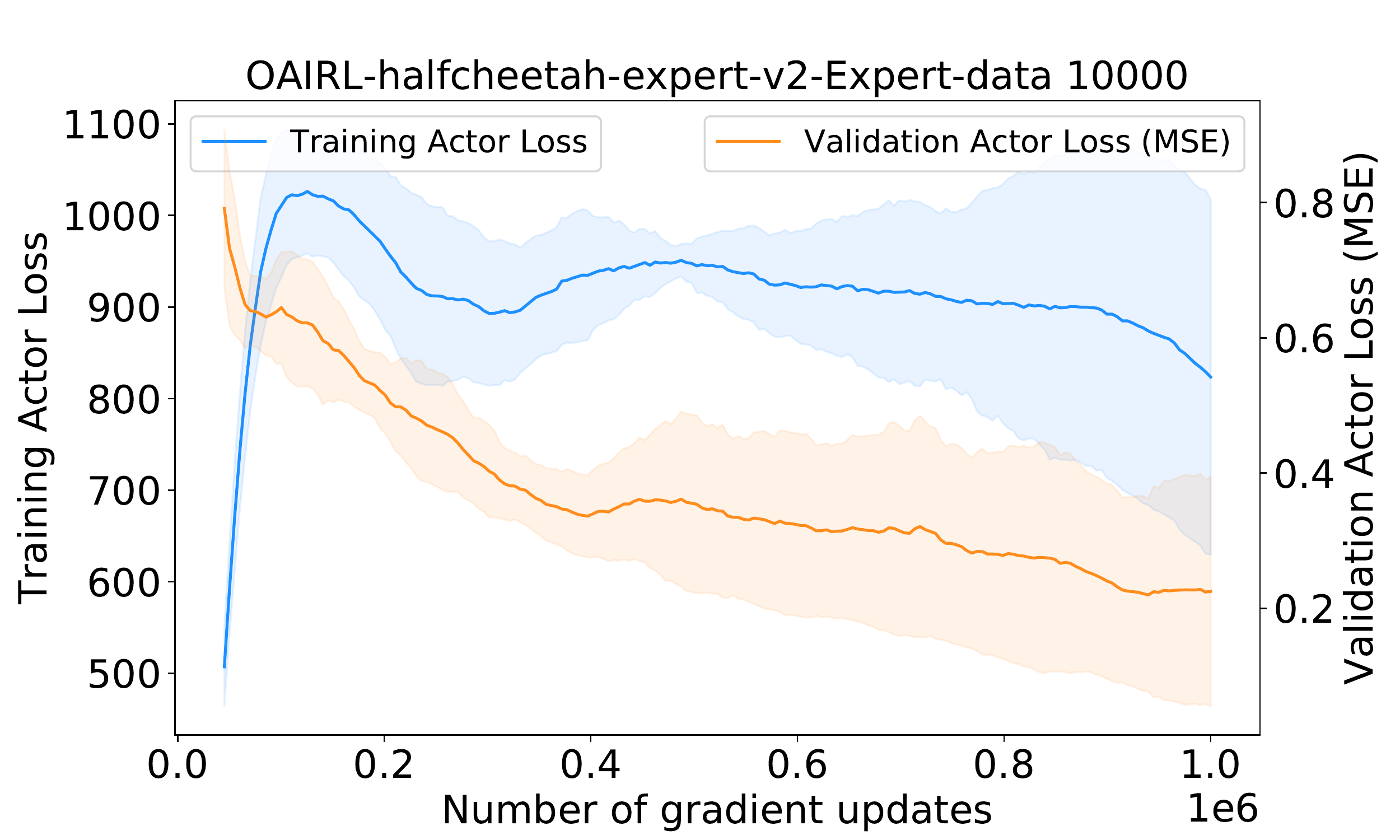}}
   \!
    \subfloat[]{\includegraphics[width=0.26\linewidth]{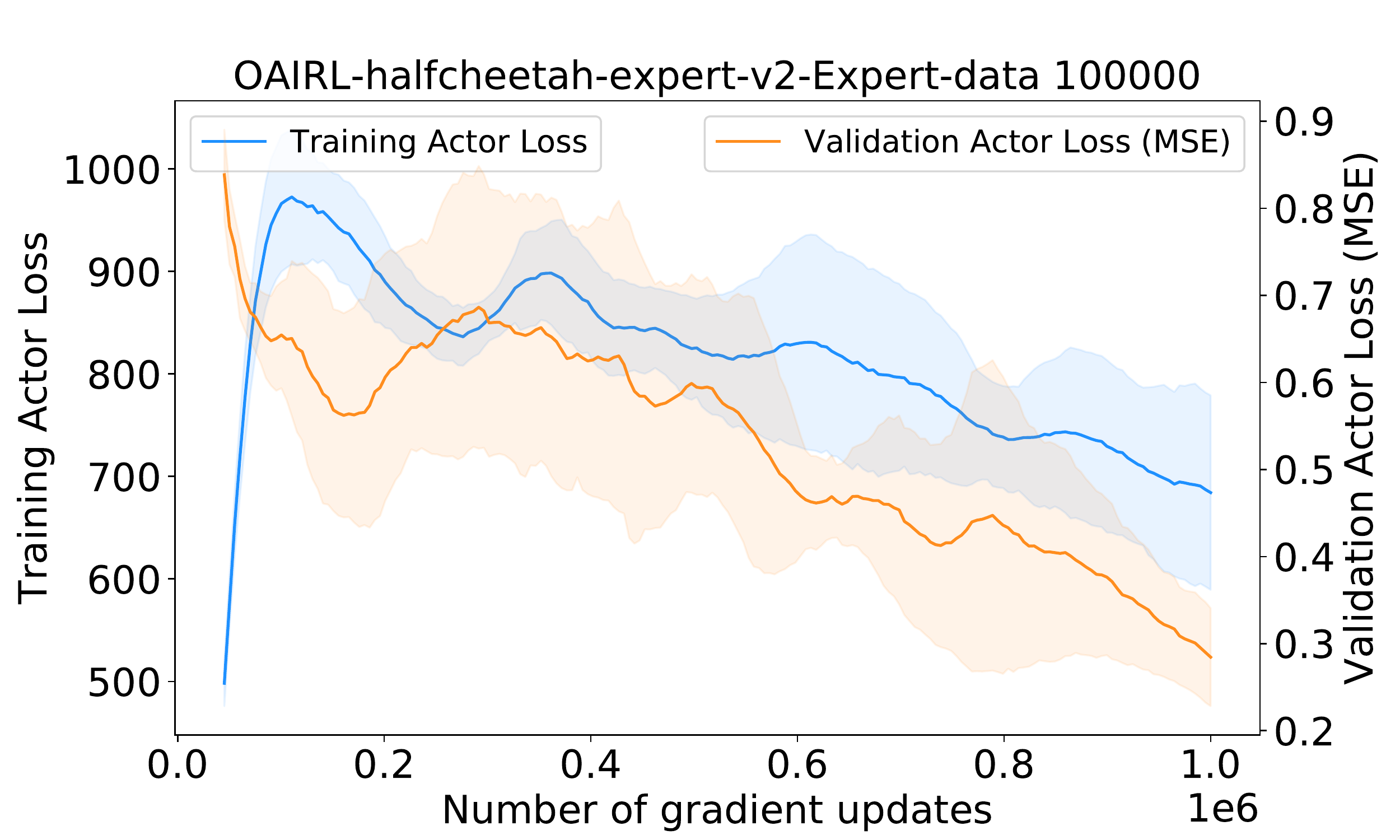}}
   \!
    \subfloat[]{\includegraphics[width=0.26\linewidth]{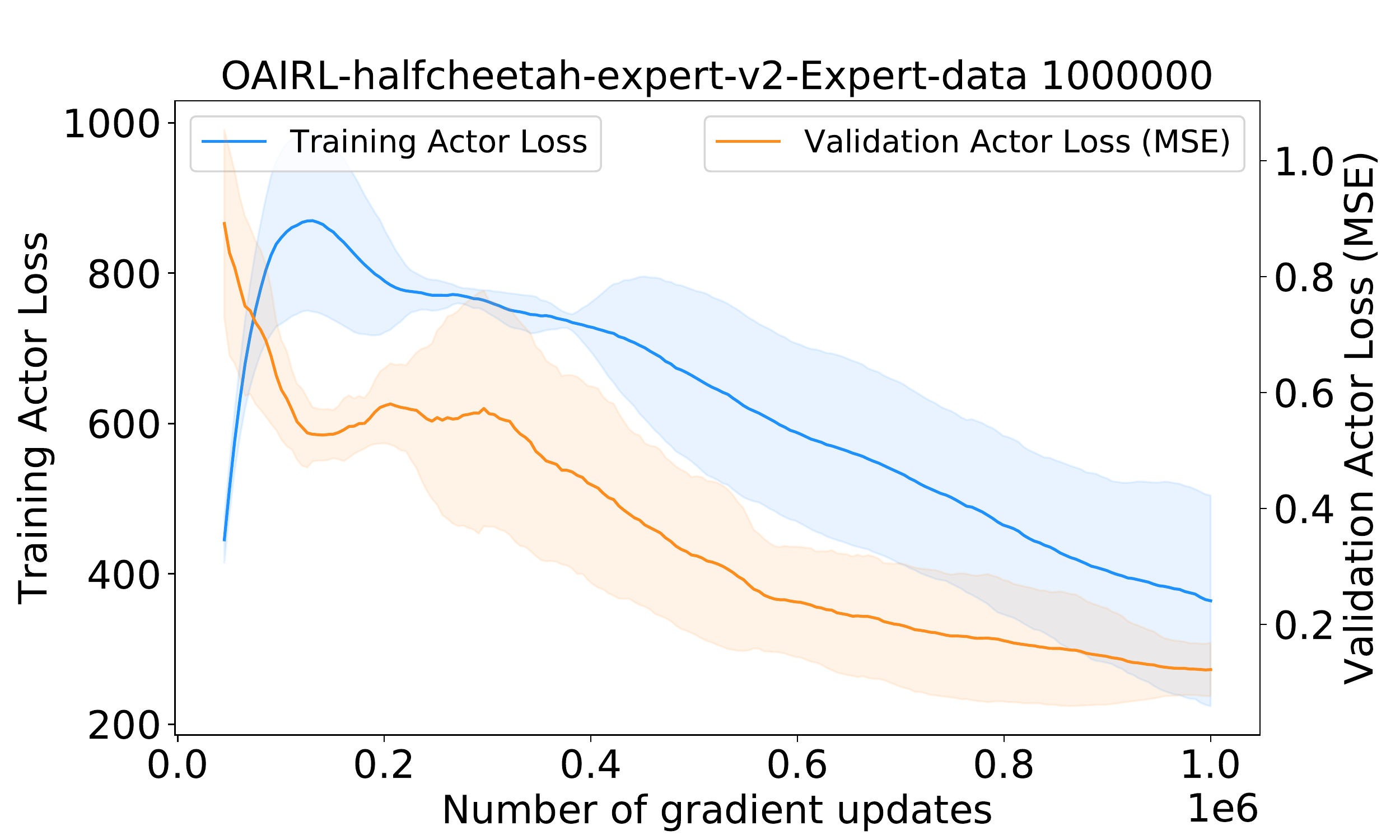}}
   \!
   
      \vspace{-0.45cm}
    \hspace*{-0.6in}
  \subfloat[]{\includegraphics[width=0.26\linewidth]{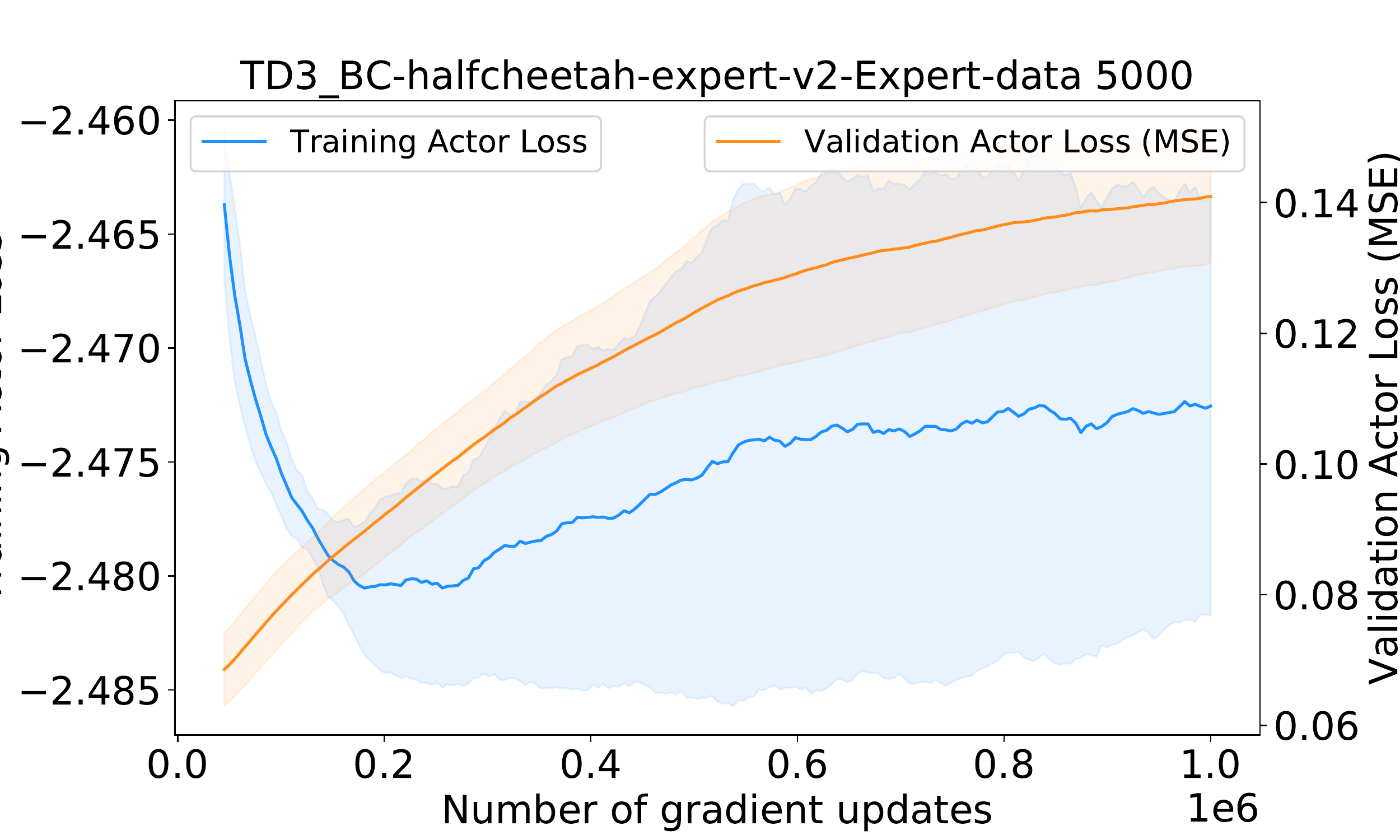}}
     \!
  \subfloat[]{\includegraphics[width=0.26\linewidth]{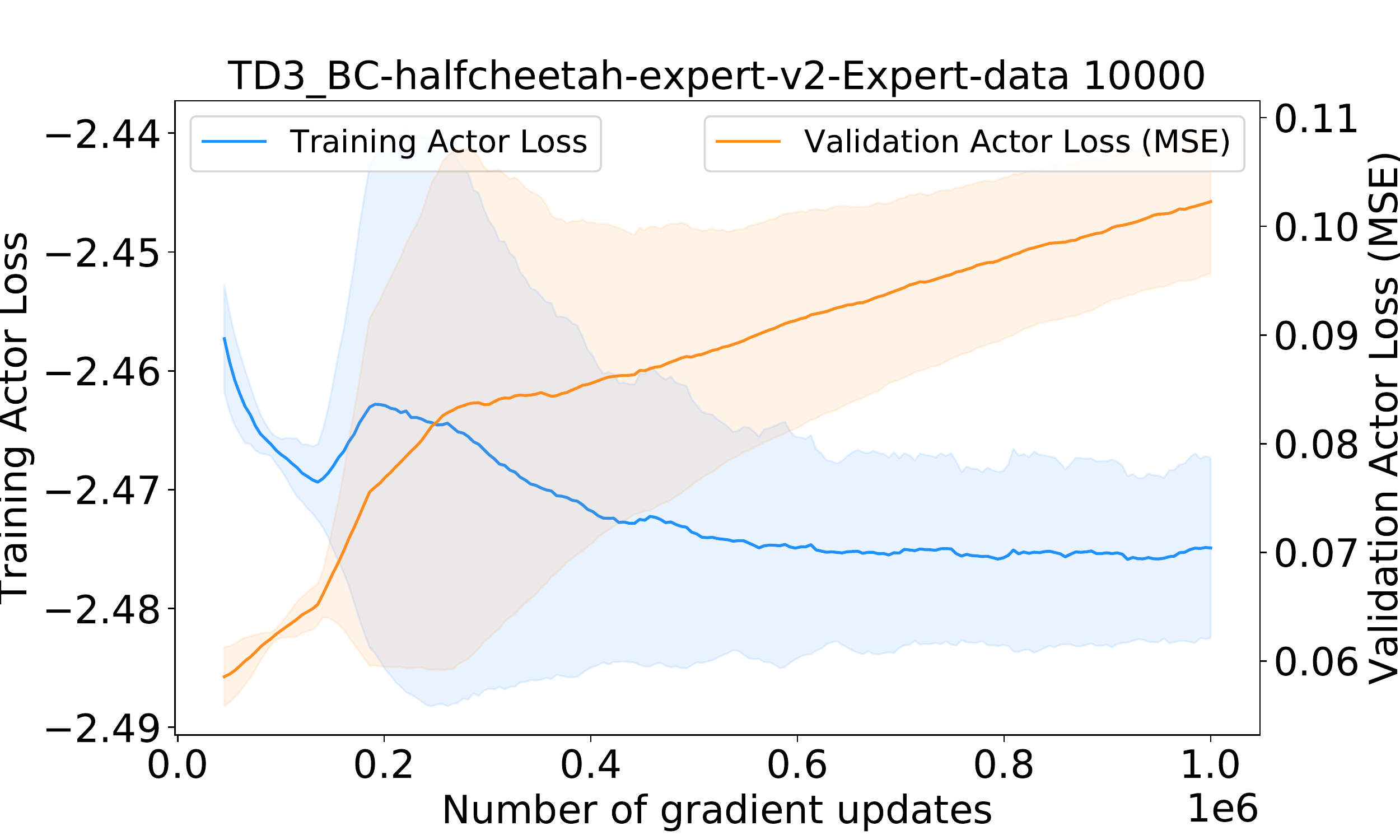}}
  \!
\subfloat[]{\includegraphics[width=0.26\linewidth]{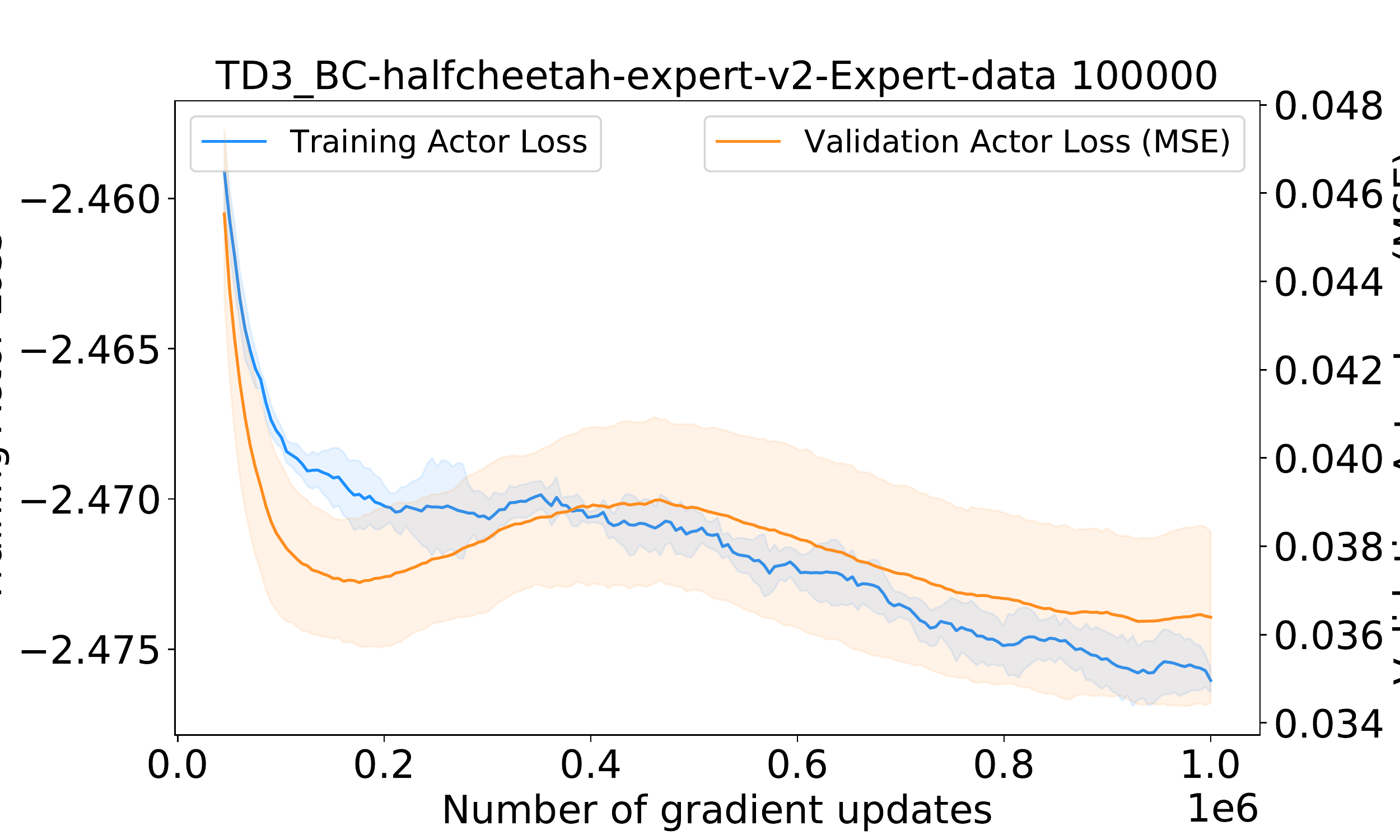}}
     \!
  \subfloat[]{\includegraphics[width=0.26\linewidth]{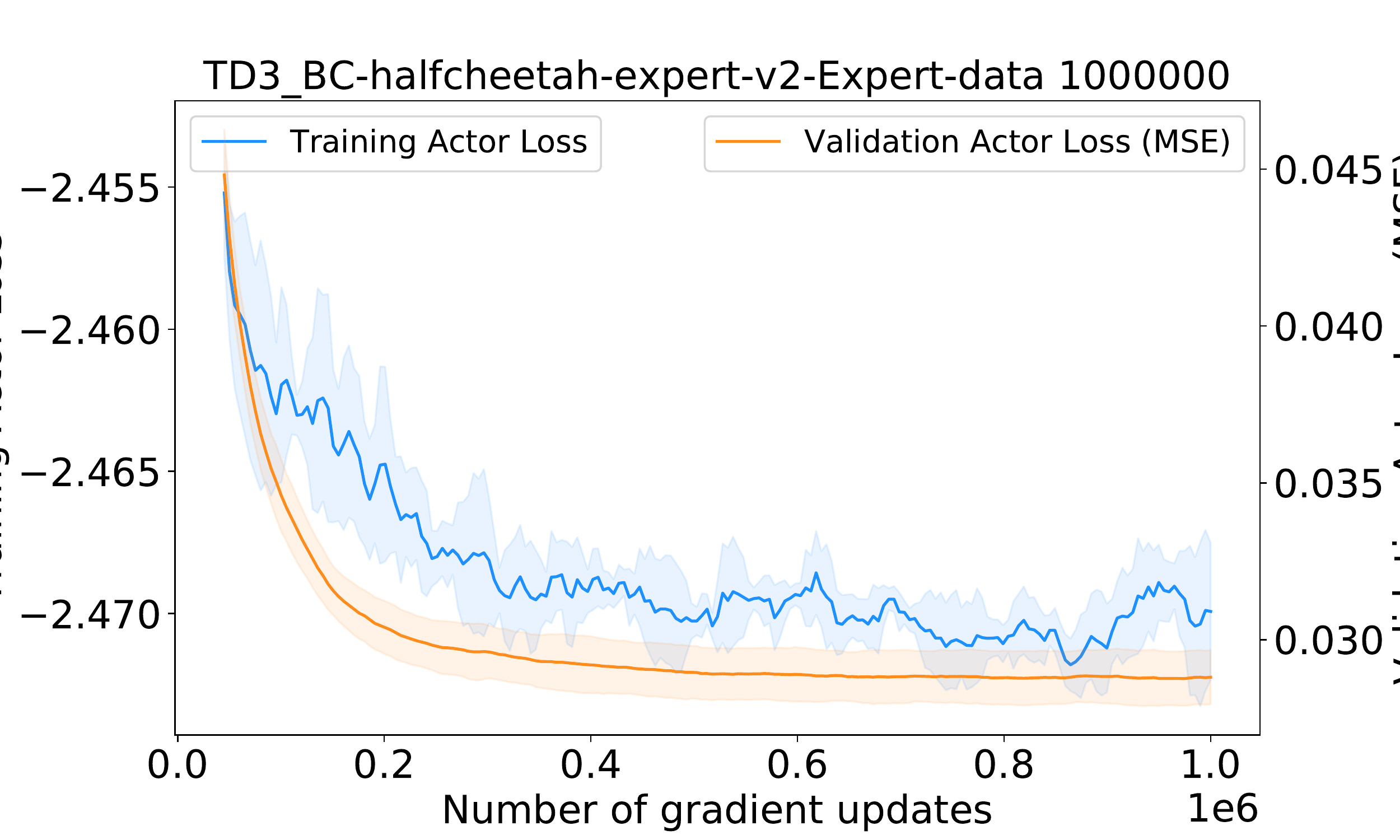}}
     \!

     \vspace{-0.45cm}
    \hspace*{-0.6in}
  \subfloat[]{\includegraphics[width=0.26\linewidth]{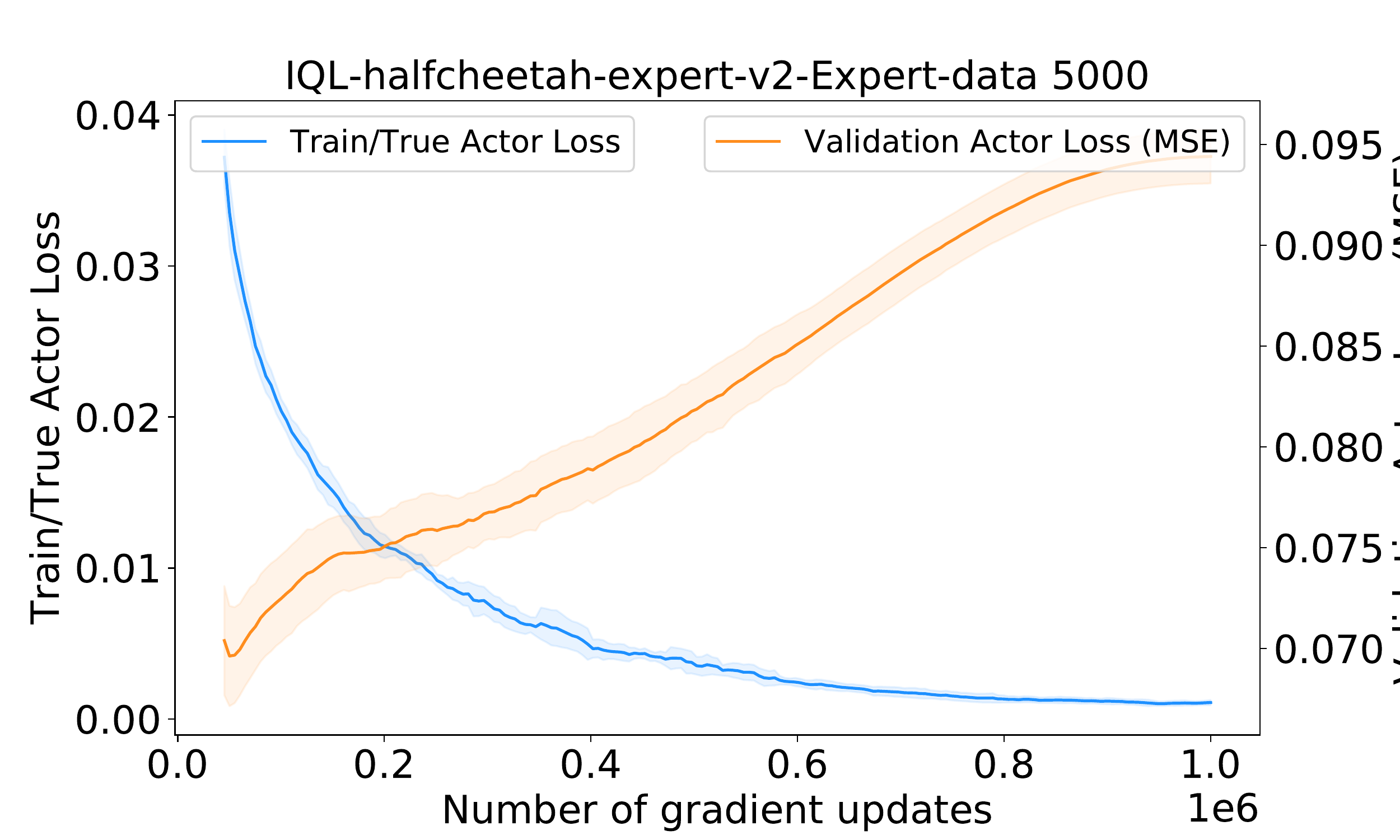}}
     \!
  \subfloat[]{\includegraphics[width=0.26\linewidth]{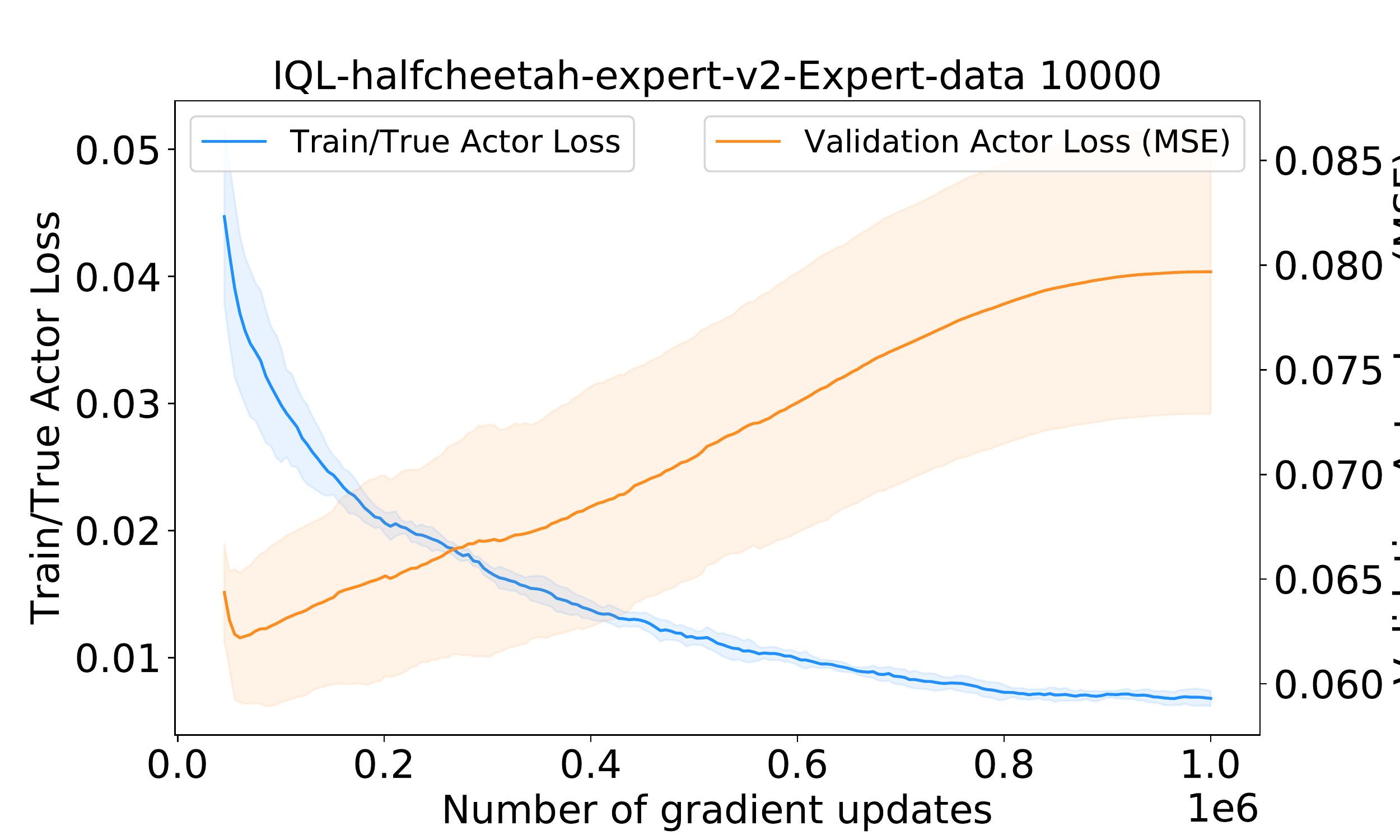}}
  \!
\subfloat[]{\includegraphics[width=0.26\linewidth]{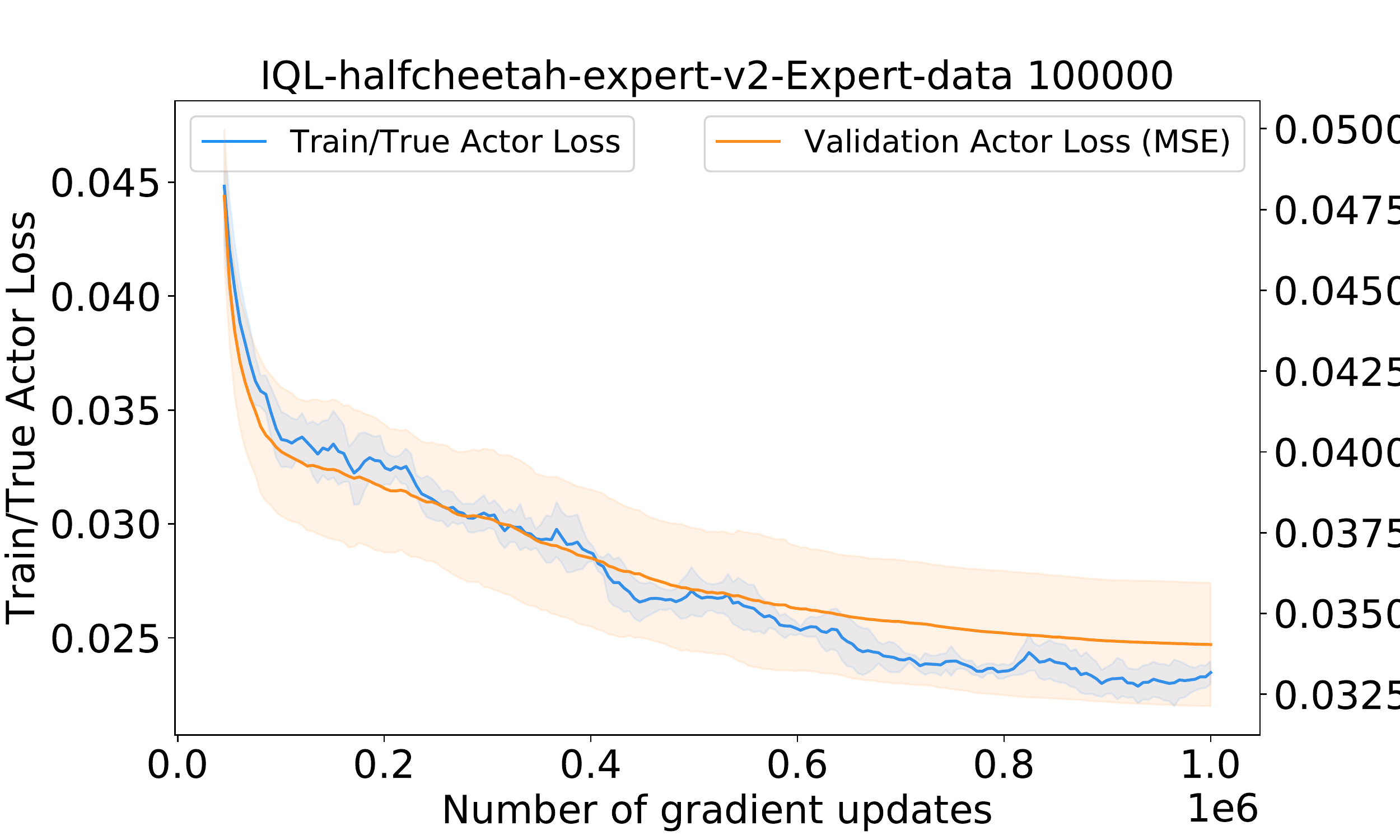}}
     \!
  \subfloat[]{\includegraphics[width=0.26\linewidth]{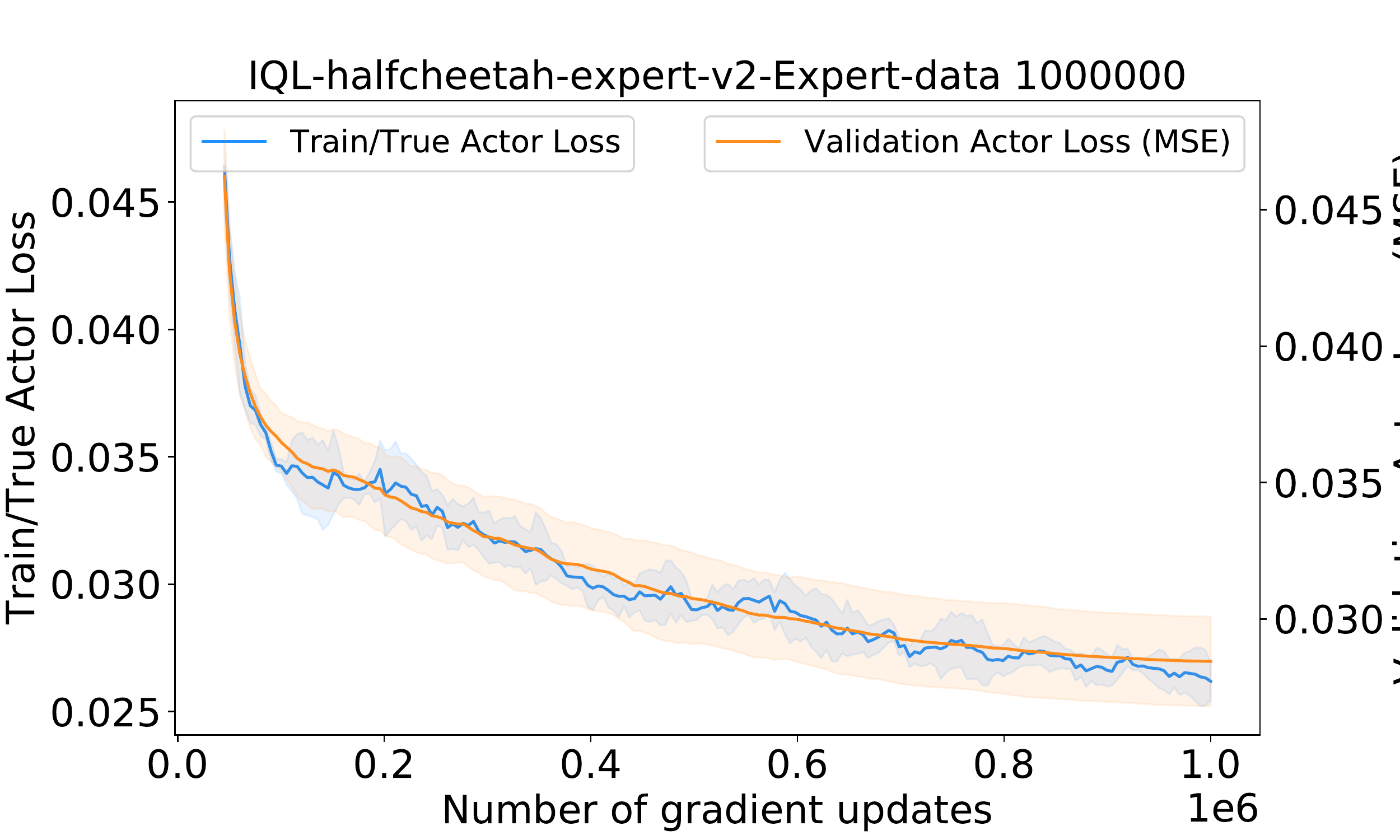}}

   \vspace{-0.45cm}
    \hspace*{-0.6in}
  \subfloat[]{\includegraphics[width=0.26\linewidth]{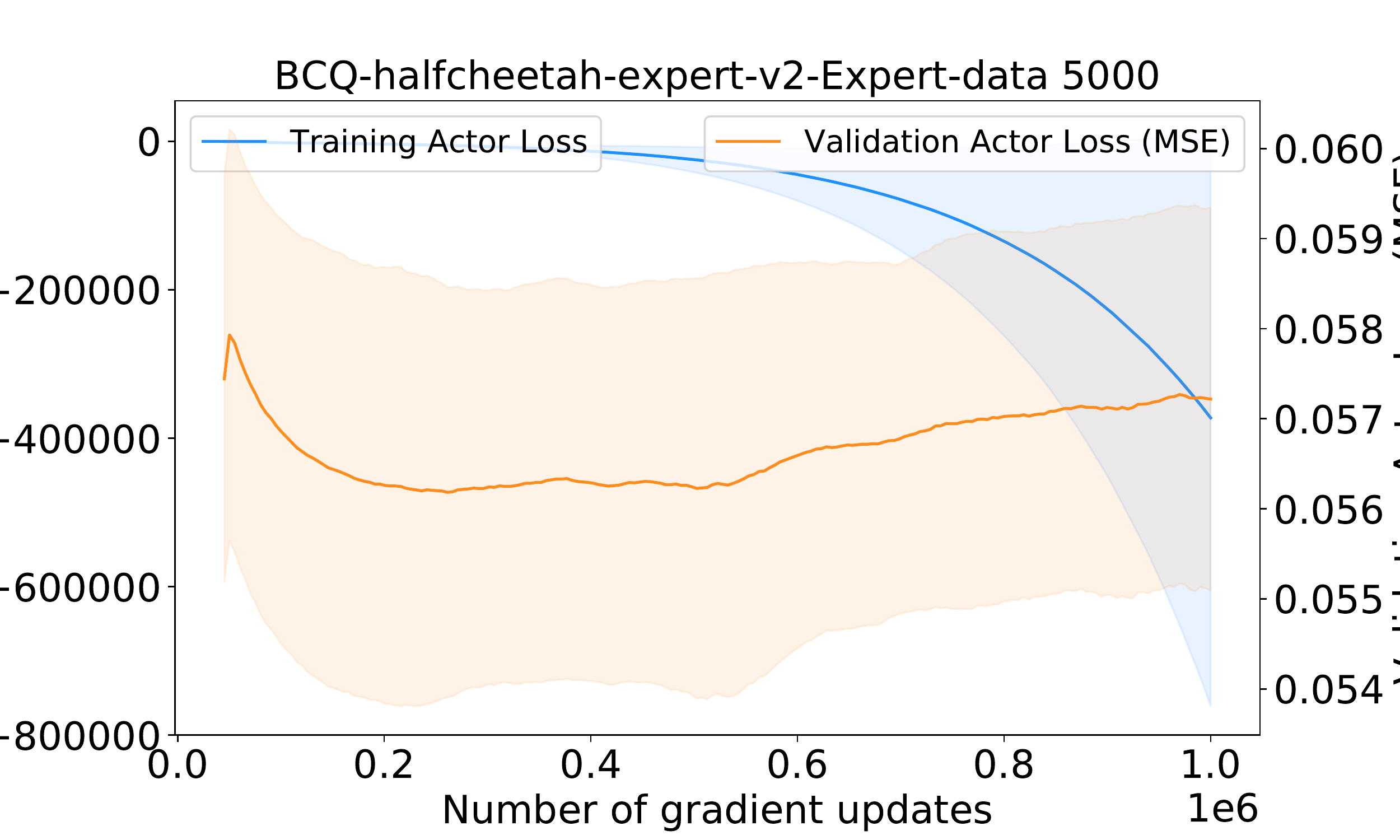}}
   \! 
    \subfloat[]{\includegraphics[width=0.26\linewidth]{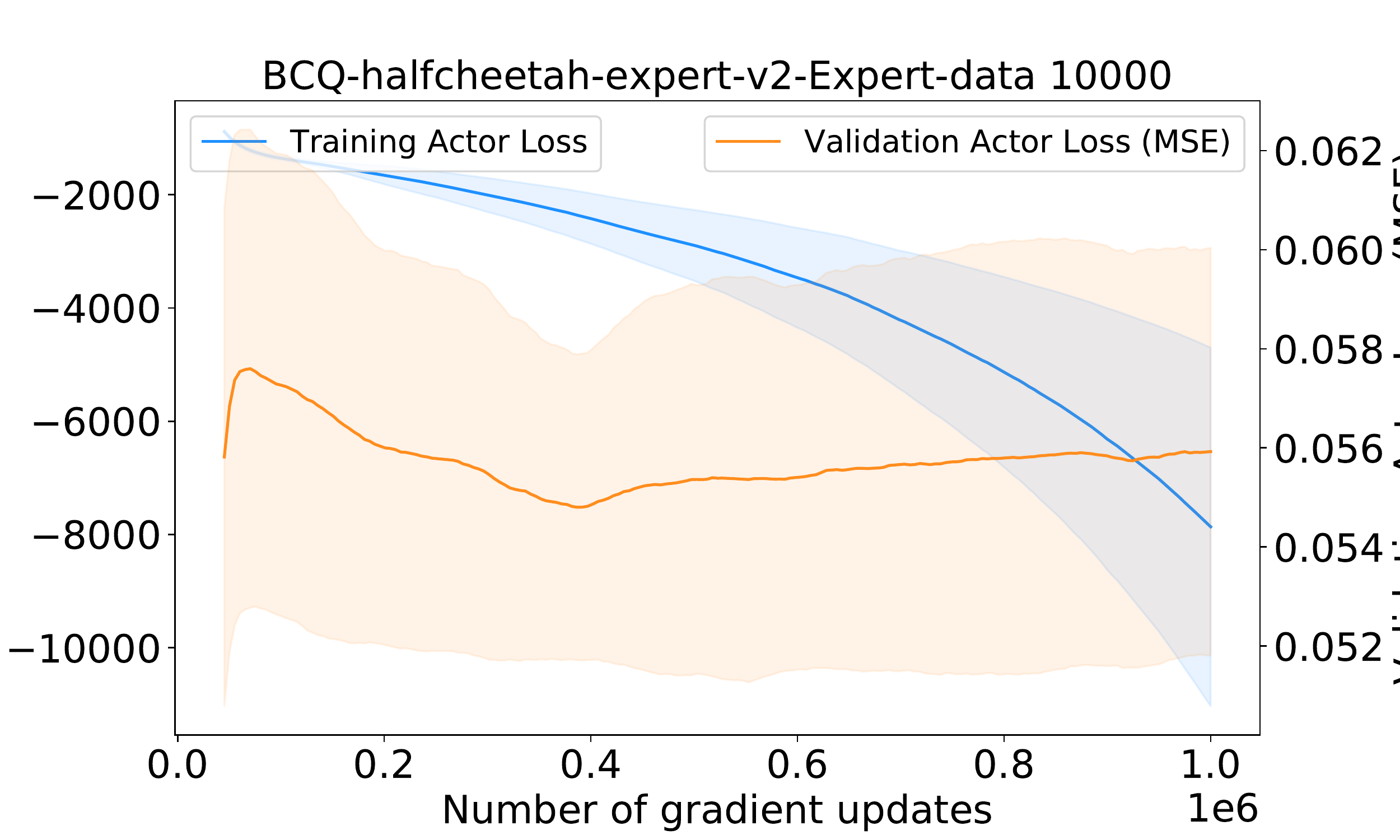}}
   \!  
    \subfloat[]{\includegraphics[width=0.26\linewidth]{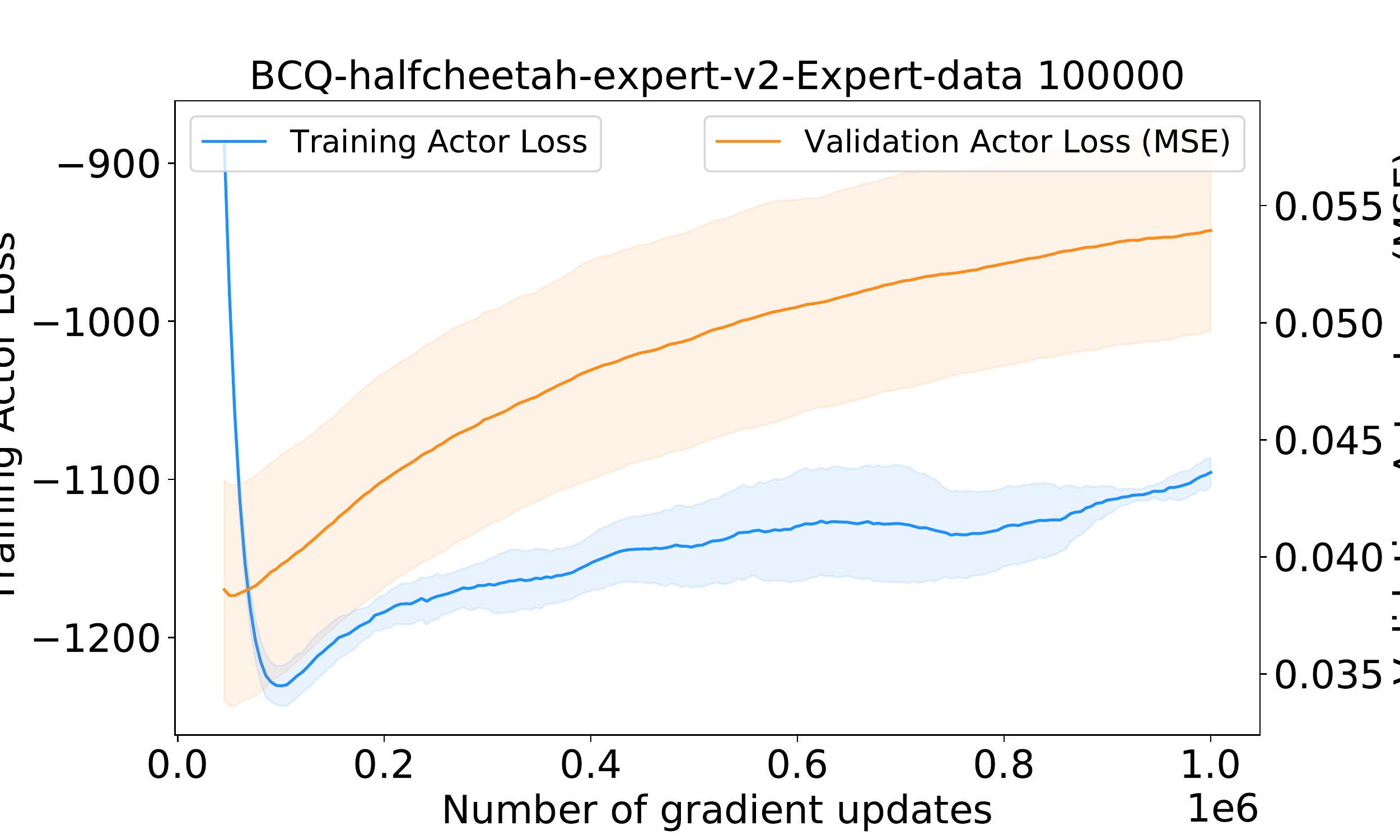}}
   \! 
    \subfloat[]{\includegraphics[width=0.26\linewidth]{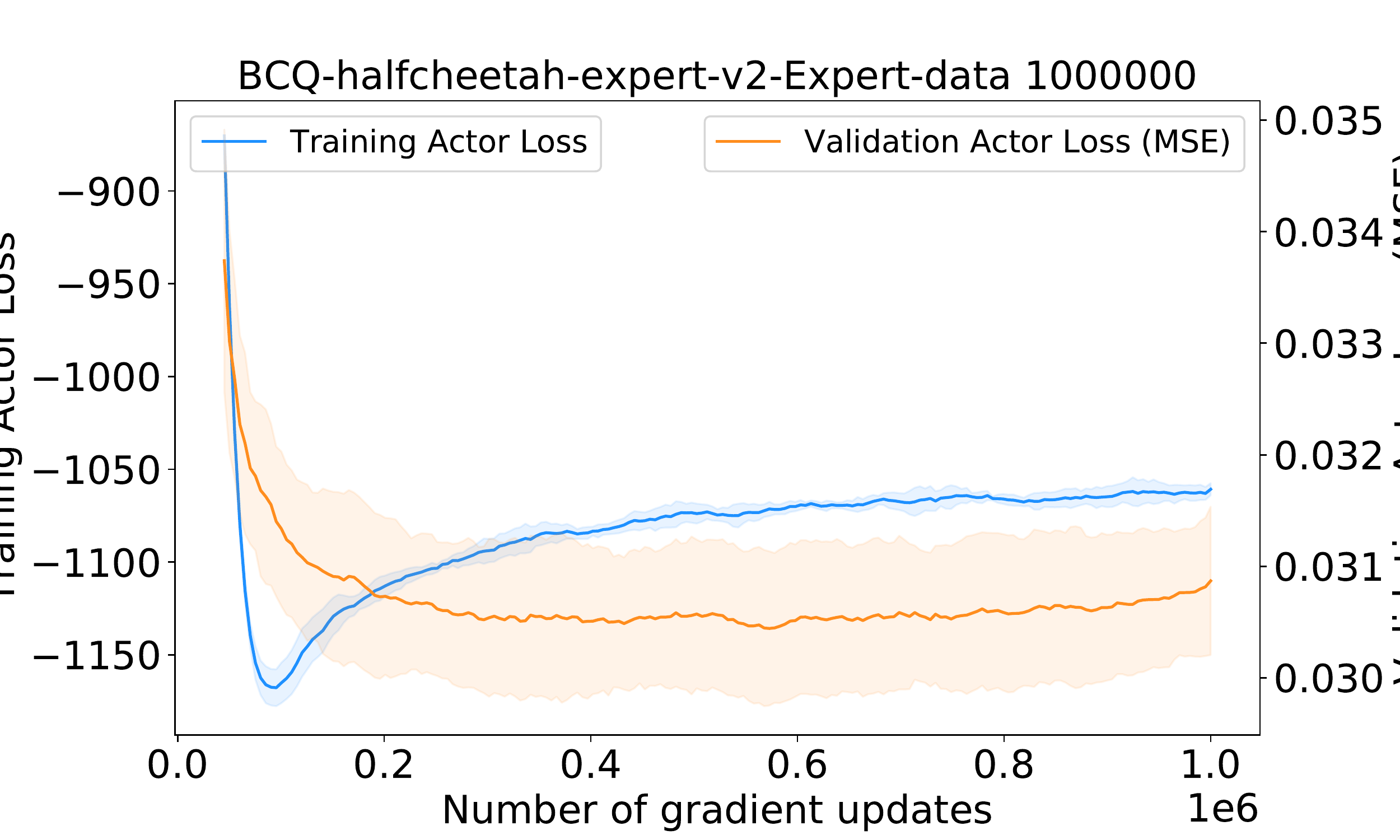}}
   \! 
   
   \vspace{-0.45cm}
    \hspace*{-0.6in}
   \subfloat[]{\includegraphics[width=0.26\linewidth]{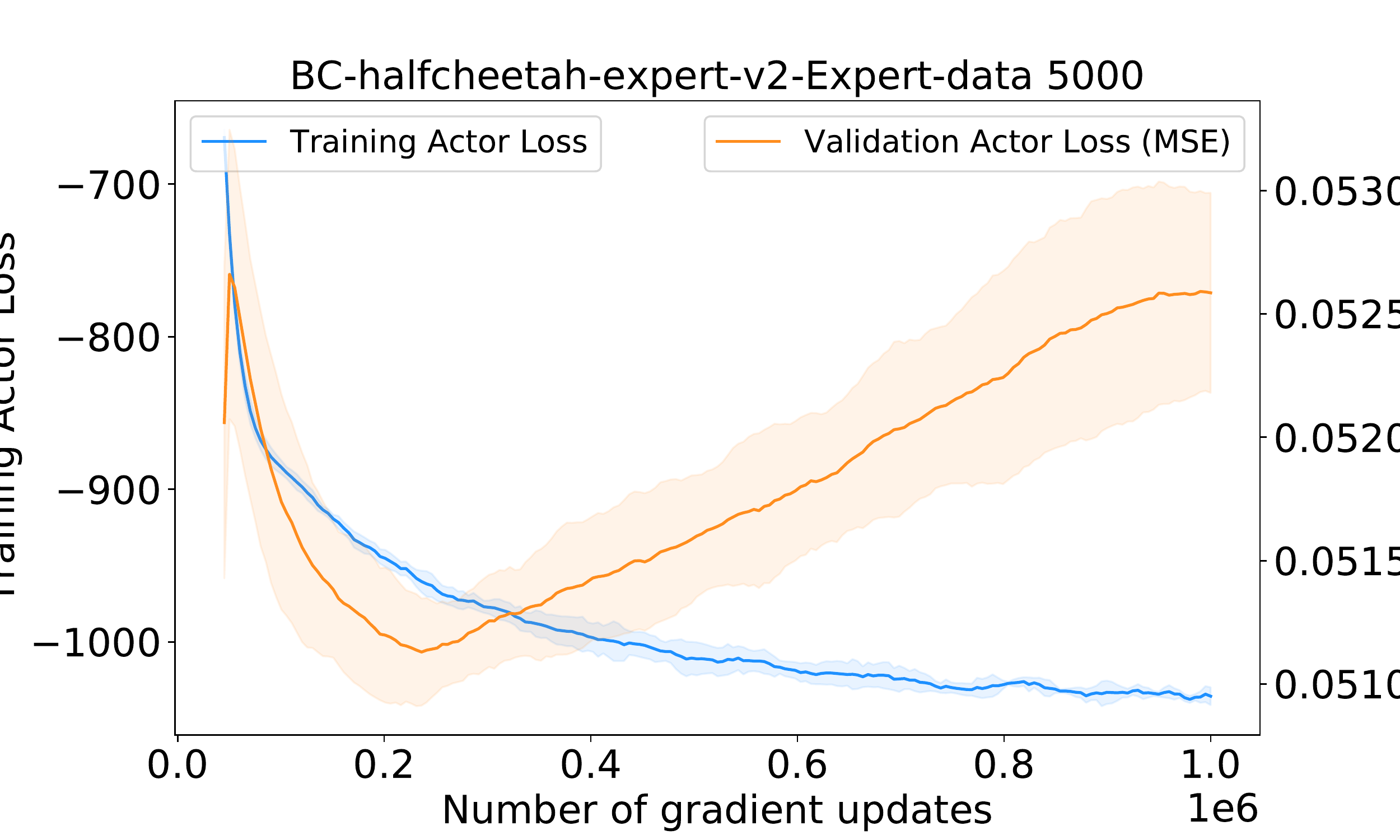}}
   \!
    \subfloat[]{\includegraphics[width=0.26\linewidth]{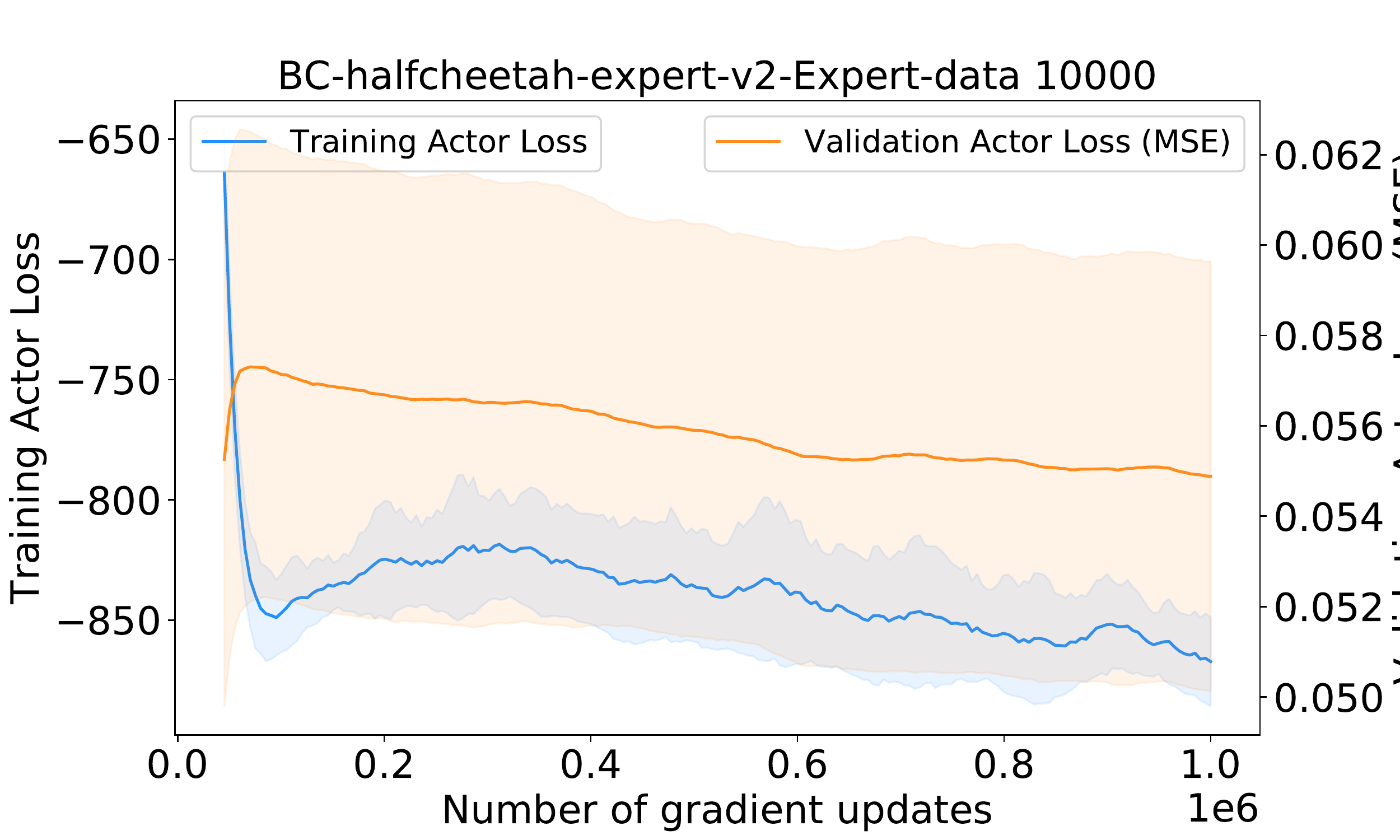}}
   \!
    \subfloat[]{\includegraphics[width=0.26\linewidth]{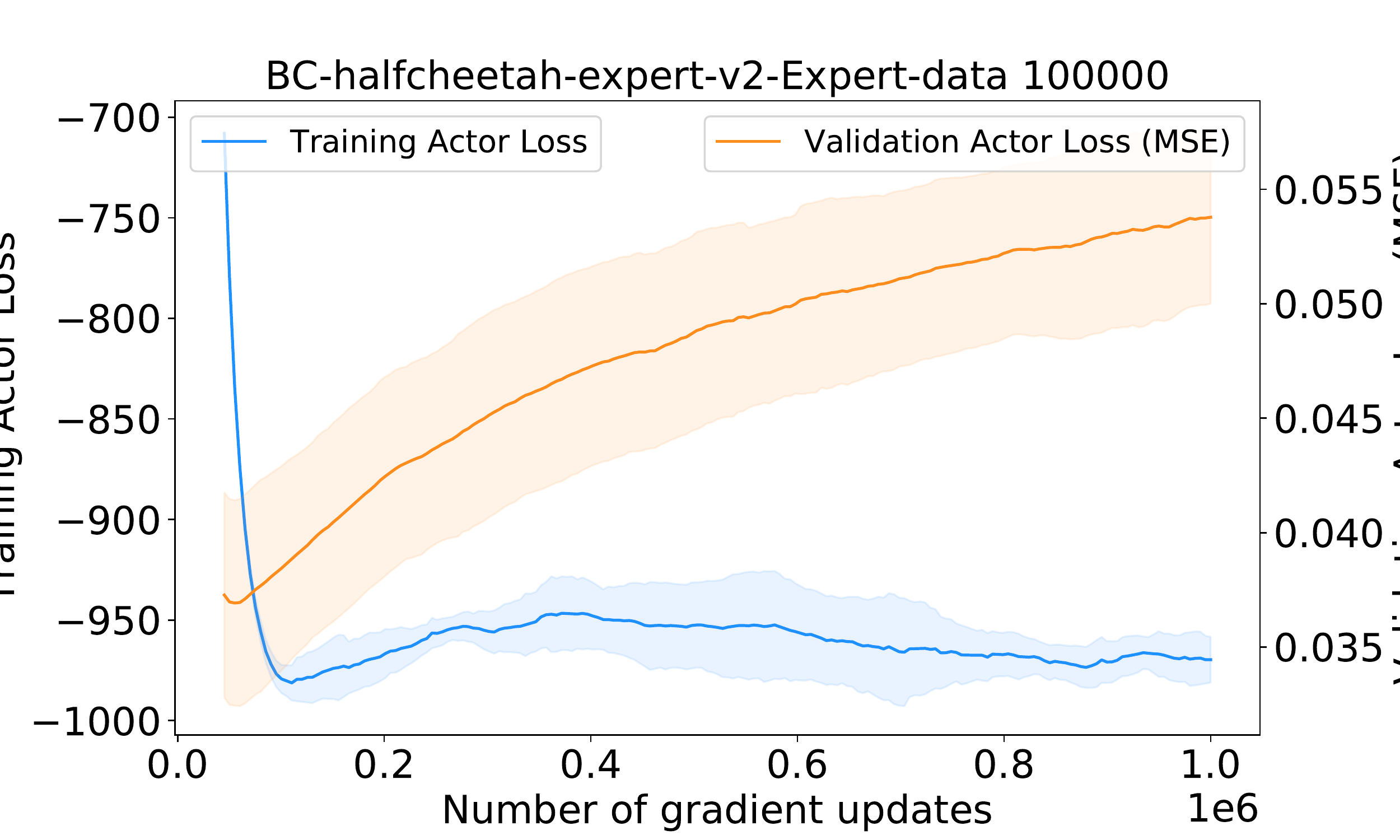}}
   \!
    \subfloat[]{\includegraphics[width=0.26\linewidth]{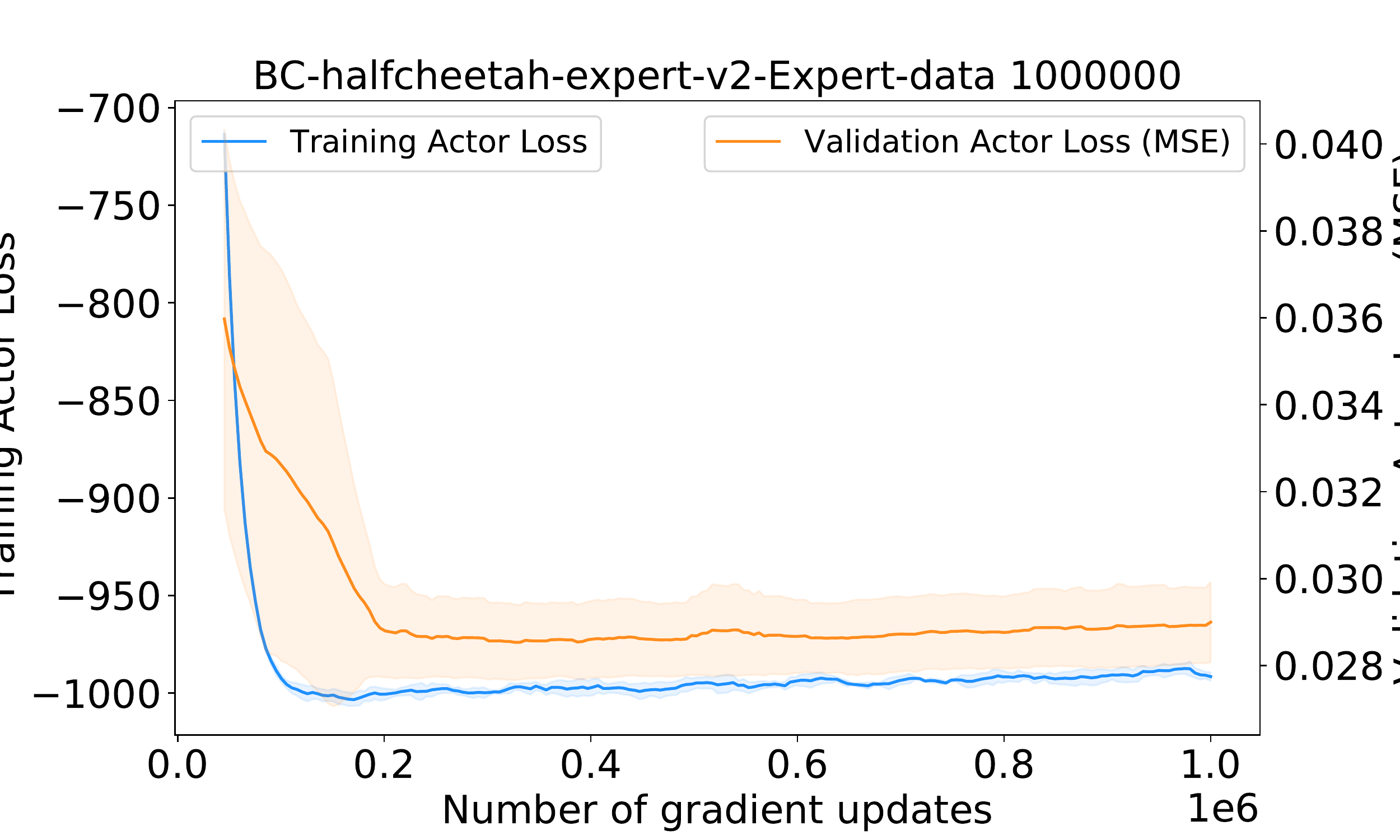}}
   \!

\caption{Training (actor/policy $\pi_theta$ training loss) and validation (MSE loss between $\pi_\theta(s_E)$ and $a_E$) performance comparison of the policy varying size of expert samples }
\label{actor_loss_train_validation}
\end{figure}

% TODO: 
% In figure \ref{actor_loss_train_validation}, we find the actor training loss (blue) to be declining as we update the actor network for all our experiment, even when we reduce the number expert training dataset (from columns left to right). In idea case, this indicates the actor's performance should be improving for all experiments.

% But the corresponding policy evaluation in online from figure \ref{DAC_vs_offline}) does not approve that. For example, 

For example, from the experiments conducted on IQL \cite{IQL} (figure \ref{actor_loss_train_validation} ($m-p$)) shows, we find the actor training loss (blue) to be declining as we update the actor network for all our experiment, even when we reduce the number expert training dataset (from columns right to left). In idea case, this indicates the actor's performance should be improving for all experiments. But the corresponding policy evaluation in online from figure \ref{DAC_vs_offline}($g$) does not approve that.

Thus we use the validation set to perform the policy-action deviation from the experts. For larger expert dataset in the training assures a declining validation loss curve but the validation loss increases for smaller dataset, and proves that smaller expert overfits the policy. We see the similar pattern in all our offline RL experiments.

For DAC and OAIRL, since the number of expert data has negligible impact (figure \ref{DAC_vs_offline}($a-f$)), the validation performance (figure \ref{actor_loss_train_validation} ($a-h$)) is always decreases with the actor's network gradient update.

\end{document}